\documentclass[review]{elsarticle}

\usepackage{lineno,hyperref}
\usepackage{caption}
\usepackage{graphicx}
\usepackage[position=t,singlelinecheck=off]{subcaption}
\usepackage[]{overpic}
\usepackage{pict2e}
\usepackage{amsmath,amssymb}    
\usepackage{times}              
\usepackage{lscape}             
\usepackage{multirow}           
\usepackage{geometry}
\usepackage{float}
\usepackage{enumitem}
\usepackage[title]{appendix}
\usepackage[dvipsnames]{xcolor}

\usepackage{hyperref}
\hypersetup{
    hidelinks,
    colorlinks=true,
    allcolors=black,
    pdfstartview=Fit,
    breaklinks=true
}

\modulolinenumbers[1]
\let\oldequation\equation
\let\oldendequation\endequation
\renewenvironment{equation}{\linenomathNonumbers\oldequation}{\oldendequation\endlinenomath}

\journal{null}

\biboptions{authoryear}

\begin{document}
\captionsetup[figure]{
    labelfont={bf},
    labelformat={default},
    labelsep=period,
    name={Fig.}
}
\captionsetup[table]{
    labelfont={bf},
    labelformat={default},
    labelsep=period,
    name={Table}
}

\begin{frontmatter}

    \title{A knowledge-based data-driven (KBDD) framework for all-day identification of cloud types using satellite remote sensing}

    \author[myaddress01,myaddress02]{Longfeng Nie}
    \cortext[mycorrespondingauthor]{Corresponding author}
    \author[myaddress03]{Yuntian Chen\corref{mycorrespondingauthor}}
    \ead{ychen@eitech.edu.cn}
    \author[myaddress06]{Mengge Du}
    \author[myaddress01,myaddress02]{Changqi Sun}
    \author[myaddress03,myaddress02,myaddress01]{Dongxiao Zhang\corref{mycorrespondingauthor}}
    \ead{dzhang@eitech.edu.cn}

    \address[myaddress01]{School of Environmental Science and Engineering, Southern University of Science and Technology, Shenzhen 518055, P. R. China}
    \address[myaddress02]{Peng Cheng Laboratory, Shenzhen, 518000, P. R. China}
    \address[myaddress03]{Ningbo Institute of Digital Twin, Eastern Institute of Technology, Ningbo, 315200, P. R. China}
    \address[myaddress06]{College of Engineering, Peking University, Beijing, 100000, P. R. China}

    \begin{abstract}
        Cloud types, as a type of meteorological data, are of particular significance for evaluating changes in rainfall, heatwaves, water resources, floods and droughts, food security and vegetation cover, as well as land use.
        In order to effectively utilize high-resolution geostationary observations, a knowledge-based data-driven (KBDD) framework for all-day identification of cloud types based on spectral information from Himawari-8/9 satellite sensors is designed.
        And a novel, simple and efficient network, named CldNet, is proposed.
        Compared with widely used semantic segmentation networks, including SegNet, PSPNet, DeepLabV3+, UNet, and ResUnet, our proposed model CldNet with an accuracy of 80.89±2.18\% is state-of-the-art in identifying cloud types and has increased by 32\%, 46\%, 22\%, 2\%, and 39\%, respectively.
        With the assistance of auxiliary information (e.g., satellite zenith/azimuth angle, solar zenith/azimuth angle), the accuracy of CldNet-W using visible and near-infrared bands and CldNet-O not using visible and near-infrared bands on the test dataset is 82.23±2.14\% and 73.21±2.02\%, respectively.
        Meanwhile, the total parameters of CldNet are only 0.46M, making it easy for edge deployment.
        More importantly, the trained CldNet without any fine-tuning can predict cloud types with higher spatial resolution using satellite spectral data with spatial resolution $\mathrm{0.02^{\circ}\times0.02^{\circ}}$, which indicates that CldNet possesses a strong generalization ability.
        In aggregate, the KBDD framework using CldNet is a highly effective cloud-type identification system capable of providing a high-fidelity, all-day, spatiotemporal cloud-type database for many climate assessment fields.
    \end{abstract}


\end{frontmatter}



\section{Introduction}
\label{section:introduction}
Clouds are important substances in the Earth's ecosystem, affecting hydrological cycles, energy balance, terrestrial ecosystems, air quality, and even food security~\citep{Narenpitak2016MS000872, Watanabe2018, Buhl2019, Eytan2020, Goldblatt2021, Cesana2021, Hieronymus2021MS002849}.
Different level clouds will have different impacts on the underlying surface.
For example, deep convection clouds bring flood-inducing extremes of precipitation to endanger life and cause economic losses~\citep{Furtado2017JD028192}; long periods of no precipitation cause drought~\citep{Hartick2022GL100924}, leading to reduced, or even no, harvest of crops.
A process-oriented climate model assessment was carried out by \citet{Kaps202361} based on cloud types derived from satellites.
Especially in the context of global climate change~\citep{Jrgensen2022, Zhang2023}, in-depth research on clouds is urgently needed.

Many studies have been conducted on cloud detection, especially the application of satellite remote sensing data in cloud mask identification~\citep{LI2007311, Shang2016JD025659, SKAKUN2022112990, QIU2019111205, SUN201770, LI2019196, SEDANO2011588, POULSEN2020111999, JOSHI2019101898}.
\citet{FOGA2017379} applied CFMask, the C language version of the function of mask (Fmask) algorithm, to detect clouds with Landsat products, and found that this algorithm performed best overall compared to other algorithms.
In order to reduce the possibility of the mismatch of cloud and cloud shadows, and improve the accuracy of cloud shadow detection in the areas with large gradients, mountainous Fmask (MFmask) was developed by \citet{QIU2017107}.
Fmask cannot distinguish whether clouds are thick or thin, therefore \citet{GHASEMIAN2018288} proposed the random forest algorithm with feature level fusion or decision level fusion to achieve this function in combination with the visible, infrared spectrum, and texture features provided by Landsat-8.
Random forest, as a machine learning method, can identify clouds without setting and debugging thresholds.
Drawing on machine learning that can simplify tedious procedures, \citet{LI201534} used subblock cloud images with brightness characteristics as learning samples for the support vector machine classifier to recognize clouds.
XGBoost-based retrieval was proposed to improve the accuracy of cloud detection over different underlying surfaces in the East Asian region, which was compared with Japan Aerospace Exploration Agency (JAXA) AHI cloud product~\citep{YANG2022112971}.

With advancements in computer hardware, many data-driven technologies have been widely used in the field of cloud detection~\citep{SEGALROZENHAIMER2020111446, LI2020112045, WU2022218, KANU2020100417, MATEOGARCIA20201}.
\citet{LI2022113197} combined the generative adversarial network (GAN) and physics-based cloud distortion model (CDM) to construct a hybrid model, GAN-CDM, to detect cloud over different underlying surfaces, including ice/snow, barren, water, urban, wetland, and forest.
The GAN-CDM model not only requires very few patch-level labels during training, but also has good transferability for different optical satellite sensors.
RS-Net based on the UNet structure was applied to Landsat-8 by \citet{Jeppesen2019247}, which is suitable for production environments due to its ability to execute quickly.
Similarly, \citet{Wieland2019111203} used the modified UNet to segment clouds in remote sensing images obtained from multiple sensors (Landsat TM, OLI, ETM+, and Sentinel-2).
In order to make the model training more accurate, cloud mask labels of the training data were obtained from images of ground-based sky cameras in the researches of \citet{Dev20191814} and \citet{SKAKUN2021102253}. However, the characteristics of clouds have not been fully explored in cloud detection algorithms.

In order to better understand the characteristics of clouds~\citep{Wang2016JD025239, Wang2019JD030457, Teng2020GL088941, Ding10020783, Khatri2017JD028165}, they are categorized into distinct cloud types based on different standards.
The most well-known standard is the International Satellite Cloud Climatology Project (ISCCP).
According to cloud top pressure and cloud optical thickness, clouds are divided into cirrus (Ci), cirrostratus (Cs), deep convection (Dc), altocumulus (Ac), altostratus (As), nimbostratus (Ns), cumulus (Cu), stratocumulus (Sc), and stratus (St)~\citep{RossowISCCP, Wang200140}.
Among them: Ci, Cs, and Dc belong to high-level clouds; Ac, As, and Ns belong to mid-level clouds; and Cu, Sc, and St belong to low-level clouds.
In early work, \citet{Jun1990} clustered satellite spectral signals to obtain different cloud types.
\citet{Segal2020111446} has developed a cloud detection algorithm based on convolutional neural networks, which simplifies the tedious process of previous threshold methods.
\citet{YuZhuofu2021} used the random forest method to divide clouds into multi-layer clouds and eight types of single-layer clouds based on the satellite FengYun-4A.
\citet{Zhang20196464} found that the use of visible channels significantly improves the ability of random forest to identify cloud types.
\citet{Purbantoro8898451} used the split window method to classify clouds based on the brightness temperature (BT) of channel 13 and brightness temperature difference (BTD) between channel 13 and channel 15 derived from the satellite Himawari-8.
\citet{Wang9554737} developed VectNet, and conducted pixel-level cloud-type classification work in the region (117°E - 129.8°E, 29.2°N - 42°N) using remote sensing data from 16 channels of the satellite Himawari-8.

Although the above researches have achieved good results in identifying cloud types depending on satellite remote sensing data, substantial work remains for further exploration.
For example, methods based on threshold judgment rely largely on professional knowledge and experience~\citep{YANG2022112971};
Many studies have focused on the recognition of cloud masks~\citep{TANA2023113548, WANG2022113079, CARABALLOVEGA2023113332}, but there is relatively little in-depth research on cloud types using deep learning~\citep{Larosa15071798, Zhao10058897};
Cloud-type recognition is mostly carried out in daytime areas~\citep{Huang9883178}, while cloud-type recognition in nighttime areas needs to be investigated.
Considering the aforementioned problems, the following efforts are mainly made:
\begin{itemize}
    \item A knowledge-based data-driven (KBDD) framework for identifying cloud types based on spectral information from Himawari-8/9 satellite sensors is designed.
    \item In order to simplify the tedious process of threshold setting, a novel, simple and efficient network, named CldNet, is proposed. Meanwhile, the widely used networks SegNet~\citep{SegNet}, PSPNet~\citep{PSPNet}, DeepLabV3+~\citep{DeepLabV3plus}, UNet~\citep{UNet}, and ResUnet~\citep{ResUnet} for pixel-level classification are used to compare with CldNet.
    \item Our proposed KBDD framework is capable of achieving all-day identification of cloud types over the entire satellite observation region, regardless of whether some areas of the region are daytime or nighttime.
    \item The trained model CldNet is applied directly to higher resolution satellite spectral input data without any fine-tuning, resulting in higher resolution cloud-type distributions ($\mathrm{0.05^{\circ}\times0.05^{\circ}}$ to $\mathrm{0.02^{\circ}\times0.02^{\circ}}$).
\end{itemize}

The main purpose of this paper is to develop a knowledge-based data-driven (KBDD) framework for all-day identification of cloud types based on spectral information from Himawari-8/9 satellite sensors.
Meanwhile, a novel, simple and efficient network, named CldNet, is proposed.
More importantly, a highly effective cloud-type identification system capable of providing a high-fidelity, all-day, spatiotemporal cloud-type database for many climate assessment fields is established.
In section~\ref{section:Data}, the study area and data processing are introduced in detail. Section~\ref{section:Methodology} describes the KBDD framework, the specific architecture of CldNet and quantitative evaluation metrics of cloud-type classification.
The performance of CldNet, SegNet, PSPNet, DeepLabV3+, UNet, ResUnet, and UNetS is shown in section~\ref{section:Results}.
And the generalization ability of CldNet, limitations of current study and prospects for future research will be discussed in depth in section~\ref{section:Discussion}.
Finally, some important conclusions and directions for future work are presented.

\section{Data}
\label{section:Data}
\subsection{Remote sensing data of the satellites Himawari-8/9}
\label{section:RS}
\begin{figure}[!htp]
    \vspace{5mm}
    \centering
    \includegraphics[width=0.6\textwidth,trim=0 0 0 0,clip]{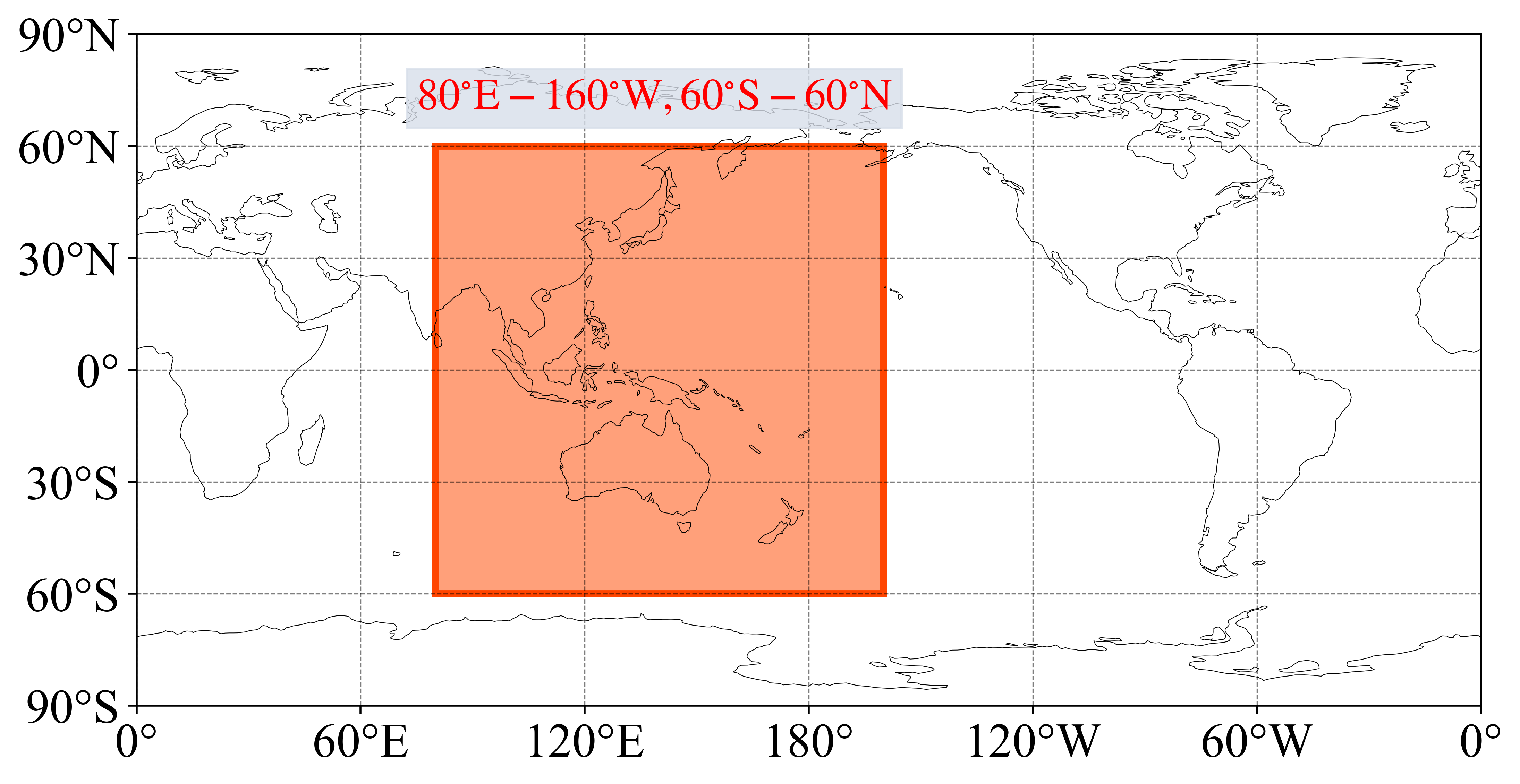}
    \caption{The scope of the satellites Himawari-8/9.}
    \label{fig:H08_scope}
\end{figure}

The satellites Himawari-8/9 belong to the Himawari 3rd generation programme~\citep{BESSHO2016, Kurihara2016, YANG2020117068}, whose main mission is operational meteorology and additional mission is environmental applications.
The Himawari-8/9 satellites were launched on October 7, 2014 and November 2, 2016, respectively.
The satellite Himawari-8 replaced the satellite MTSAT-2 as an operational satellite on July 7, 2015, and the satellite Himawari-9 replaced the satellite Himawari-8 as the primary satellite on December 13, 2022.
The scope of the satellites Himawari-8/9 is shown in Fig.~\ref{fig:H08_scope}, whose coverage area is from 80°E to 160°W, and from 60°N to 60°S.
Satellite spectral channels provide albedo from B01 to B06 and brightness temperature (BT) from B07 to B16, whose specific information is shown in Table~\ref{table:Himawari_infos}~\citep{HuangYipeng122, Taniguchi2022}.
In this study, Himawari L1 gridded data within the study area were downloaded through the Japan Aerospace Exploration Agency (JAXA) Himawari Monitor P-Tree System (available at \href{https://www.eorc.jaxa.jp/ptree/index.html}{https://www.eorc.jaxa.jp/ptree/index.html}).

\begin{table}[!htp]
    \caption{Specific information for each spectral channel of the satellites Himawari-8/9 imagery.}
    \centering
    \resizebox{0.9\linewidth}{!}{
        \begin{tabular}{ccccccc}
            \hline
            Channel  & Central wavelength & Bandwidth & SNR or NE$\Delta$T $@$ specified input & Resolution & Primary application                     & Value range    \\ \hline
            B01 & 455 nm             & 50 nm     & $\ge$ 300 $@$ albedo (\%)           & 1.0 km     & Aerosol                                 & [0, 100]   \\
            B02 & 510 nm             & 20 nm     & $\ge$ 300 $@$ albedo (\%)           & 1.0 km     & Aerosol                                 & [0, 100]   \\
            B03 & 645 nm             & 30 nm     & $\ge$ 300 $@$ albedo (\%)           & 0.5 km     & Low cloud, Fog                          & [0, 100]   \\
            B04 & 860 nm             & 20 nm     & $\ge$ 300 $@$ albedo (\%)           & 1.0 km     & Vegetation, Aerosol                     & [0, 100]   \\
            B05 & 1610 nm            & 20 nm     & $\ge$ 300 $@$ albedo (\%)           & 2.0 km     & Cloud phase, SO2                        & [0, 100]   \\
            B06 & 2260 nm            & 20 nm     & $\ge$ 300 $@$ albedo (\%)           & 2.0 km     & Particle size                           & [0, 100]   \\
            B07 & 3.85 µm            & 0.22 µm   & $\le$ 0.16 $@$ BT (K)               & 2.0 km     & Low cloud, Fog, Forest fire             & [220, 335] \\
            B08 & 6.25 µm            & 0.37 µm   & $\le$ 0.40 $@$ BT (K)               & 2.0 km     & Upper level moisture                    & [200, 260] \\
            B09 & 6.95 µm            & 0.12 µm   & $\le$ 0.10 $@$ BT (K)               & 2.0 km     & Mid-upper level moisture                & [200, 270] \\
            B10 & 7.35 µm            & 0.17 µm   & $\le$ 0.32 $@$ BT (K)               & 2.0 km     & Mid-level moisture                      & [200, 275] \\
            B11 & 8.60 µm            & 0.32 µm   & $\le$ 0.10 $@$ BT (K)               & 2.0 km     & Cloud phase, SO2                        & [200, 320] \\
            B12 & 9.63 µm            & 0.18 µm   & $\le$ 0.10 $@$ BT (K)               & 2.0 km     & Ozone content                           & [210, 295] \\
            B13 & 10.45 µm           & 0.30 µm   & $\le$ 0.10 $@$ BT (K)               & 2.0 km     & Cloud imagery, Information of cloud top & [200, 330] \\
            B14 & 11.20 µm           & 0.20 µm   & $\le$ 0.10 $@$ BT (K)               & 2.0 km     & Cloud imagery, Sea surface temperature  & [200, 330] \\
            B15 & 12.35 µm           & 0.30 µm   & $\le$ 0.10 $@$ BT (K)               & 2.0 km     & Cloud imagery, Sea surface temperature  & [200, 320] \\
            B16 & 13.30 µm           & 0.20 µm   & $\le$ 0.30 $@$ BT (K)               & 2.0 km     & Cloud top height                        & [200, 295] \\ \hline
            \end{tabular}
    }
    \label{table:Himawari_infos}
\end{table}

\begin{figure}[!htp]
    \centering
    \subcaptionbox{\vspace{-2mm}\label{fig:RS_BT:frequency}}{%
        \includegraphics[width=0.75\textwidth,trim=0 0 0 0,clip]{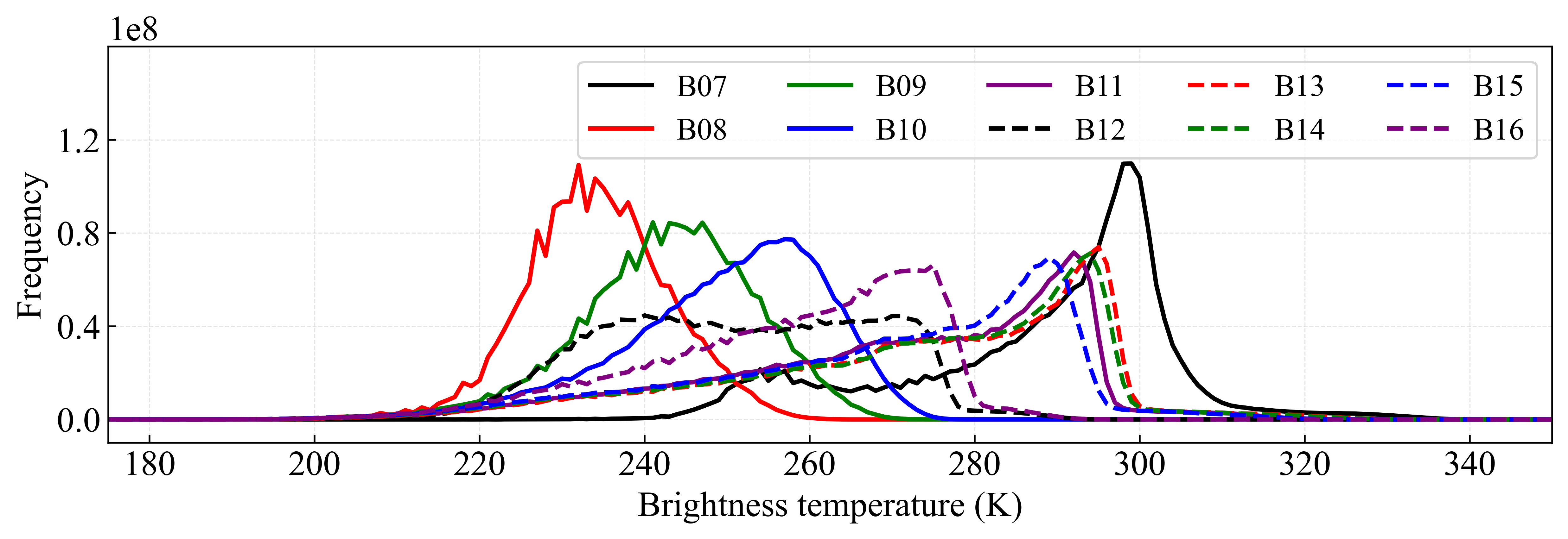}
    }
    \subcaptionbox{\vspace{-2mm}\label{fig:RS_BT:cumulative_percentage}}{%
        \includegraphics[width=0.75\textwidth,trim=0 0 0 0,clip]{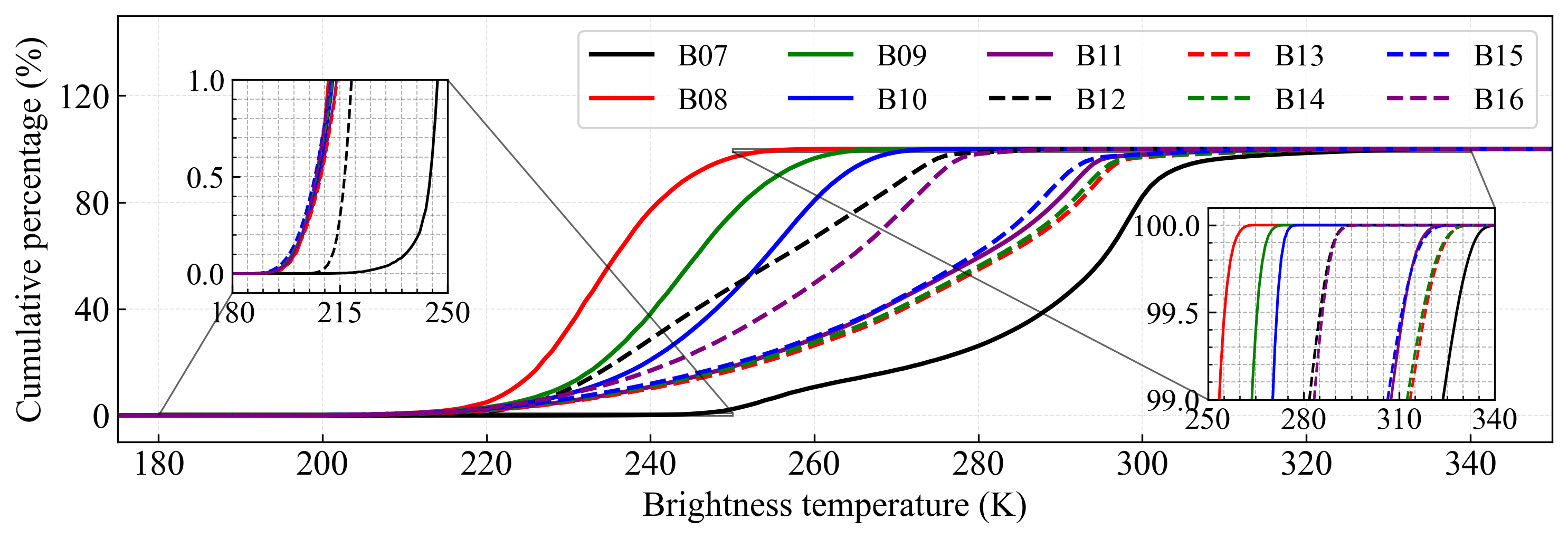}
    }
    \caption{The (a) frequency and (b) cumulative percentage curves of brightness temperature (BT) from B07 to B16.}
    \label{fig:RS_BT}
\end{figure}

The remote sensing data of the spatial resolution $0.05^{\circ}\times0.05^{\circ}$ at 03:00 (UTC+0) were downloaded every 3 d from January 1, 2020 to December 31, 2022.
The value range of albedo from B01 to B06 is [0, 100]\%, but the value range of BT from B07 to B16 needs to be determined statistically.
The BT values corresponding to different channels of each pixel in the downloaded remote sensing images are counted.
The frequency and cumulative percentage curves of BT from B07 to B16 are plotted in Fig.~\ref{fig:RS_BT:frequency}~and~\ref{fig:RS_BT:cumulative_percentage}, respectively.
In Fig.~\ref{fig:RS_BT:frequency}, the BT value of the peak frequency of B08 is the smallest, while the BT value of the peak frequency of B07 is the largest.
The frequency curves for B11, B13, B14, and B15 have the same trend; those for B08, B09, and B10 are relatively consistent with a normal distribution; and the BT values for B12 are evenly distributed between 235K and 275K.
In order to accurately determine the range of BT values for each channel, the cumulative percentage of BT for each channel is calculated in Fig.~\ref{fig:RS_BT:cumulative_percentage}.
The images at the 0\% and 100\% positions of the cumulative percentage curve are magnified and observed to determine the range of BT values for each channel.
The range of BT values for each channel is determined and recorded in Table~\ref{table:Himawari_infos}, which covers 99.9\% of its own data.
Based on the range of spectral channel values, the model inputs can be reasonably normalized.

\subsection{Cloud types}
\label{section:Cloud}
\begin{figure}[!htp]
    \vspace{10mm}
    \centering
    \includegraphics[width=0.6\textwidth,trim=0 0 0 27,clip]{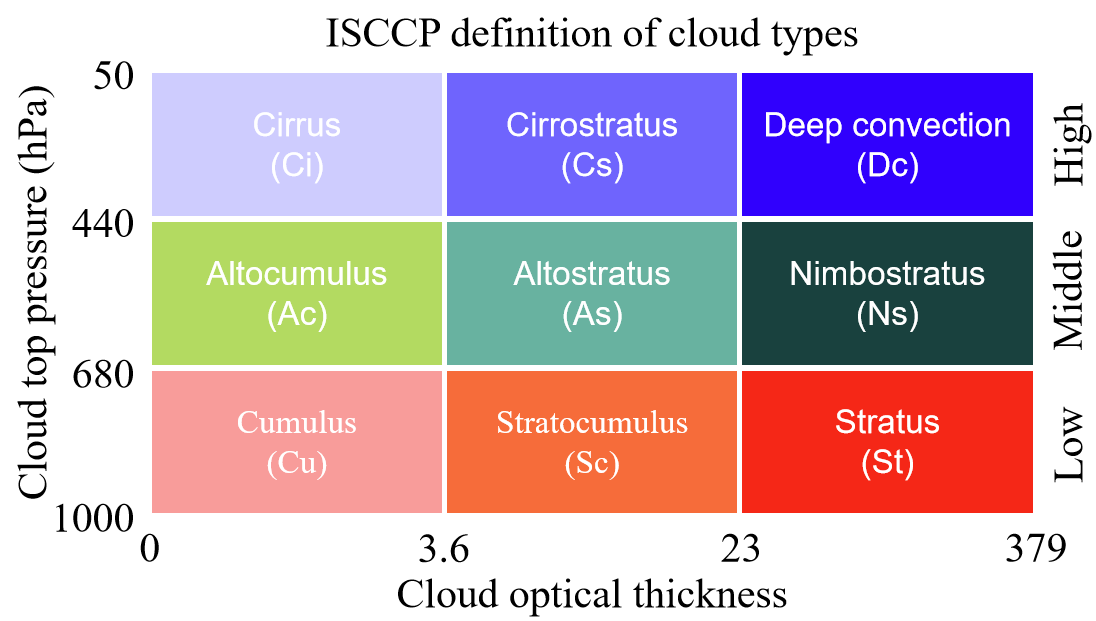}
    \caption{The ISCCP definition of cloud types.}
    \label{fig:Cld_definition}
\end{figure}

The International Satellite Cloud Climatology Project (ISCCP) definition of cloud types is presented in Fig.~\ref{fig:Cld_definition}, which is detailed at \href{https://isccp.giss.nasa.gov}{https://isccp.giss.nasa.gov}.
Under cloudy skies, cloud types are classified as cirrus (Ci), cirrostratus (Cs), deep convection (Dc), altocumulus (Ac), altostratus (As), nimbostratus (Ns), cumulus (Cu), stratocumulus (Sc), and stratus (St), depending on cloud optical thickness and cloud top pressure.
Otherwise, the sky is clear (Cl).
The JAXA Himawari Monitor P-Tree System provides Himawari L2 gridded data cloud types with spatial resolution $0.05^{\circ}\times0.05^{\circ}$ and temporal resolution 10 min~\citep{Letu8579544, Lai11141703, LETU2020111583}.

\begin{figure}[!htp]
    \centering
    \includegraphics[width=0.98\textwidth,trim=70 0 70 0,clip]{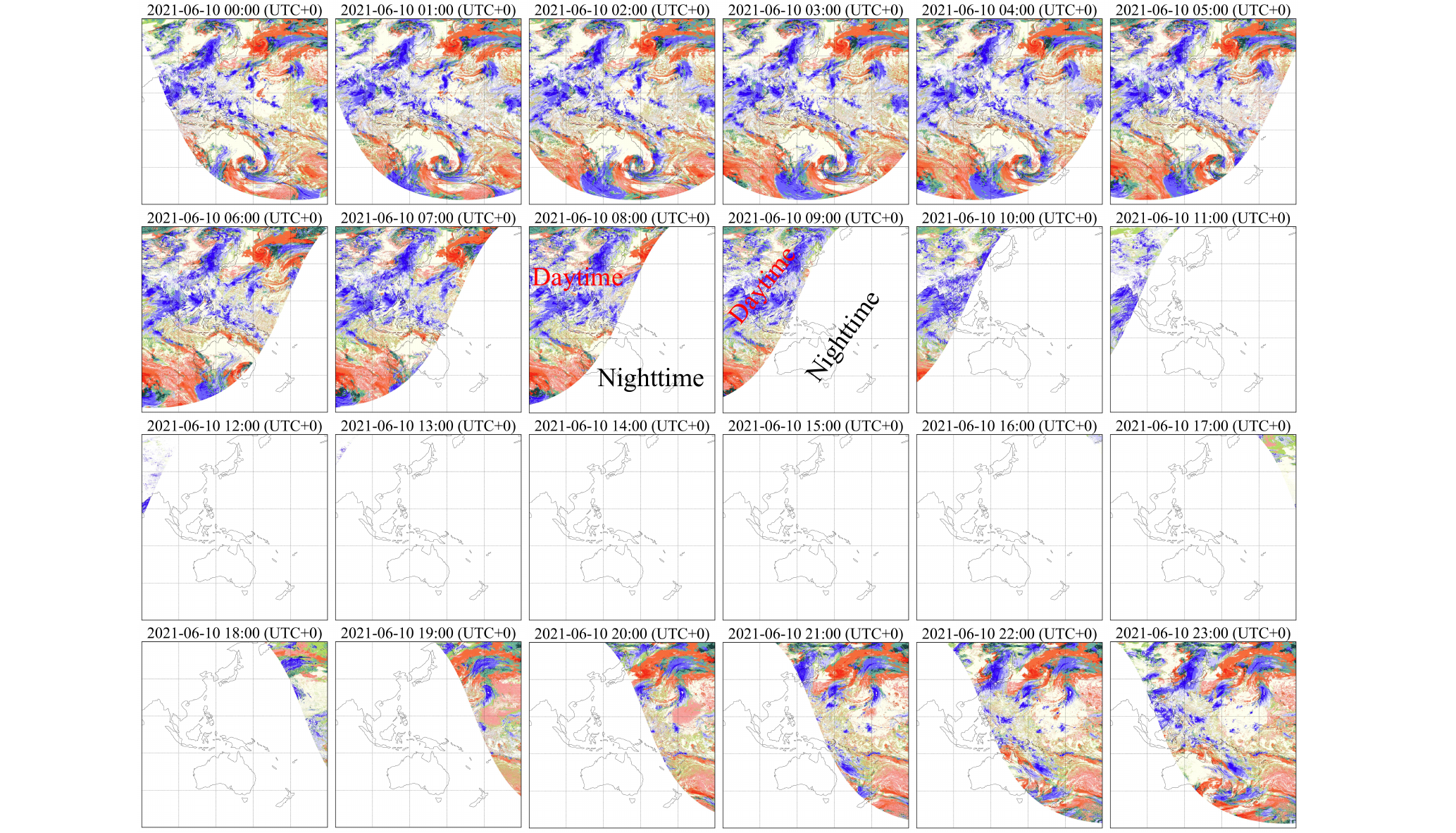}
    \caption{The overall cloud-type distributions every hour during June 10, 2021, whose color scheme is from Fig.~\ref{fig:Cld_definition}.}
    \label{fig:1day_Cld}
\end{figure}

In Fig.~\ref{fig:1day_Cld}, the change process of the overall cloud-type distributions every hour on June 10, 2021 is shown.
Cloud types are found to be available only in the daytime region, while those in the nighttime region are unknown.
\citet{Wang2022113026} pointed out that the physical algorithms~\citep{Nakajima1995,Ishida2008JD010710} used by JAXA's cloud-type product involve visible (VIS) B01 $\sim$ B03, near-infrared (NIR) B04 $\sim$ B06 and infrared (IR) B07 $\sim$ B16 bands, and thus cloud-type retrieval is limited to the daytime region only.
Here, the cloud-type distributions at 03:00 (UTC+0) every 3 d are chosen as classification labels for this research from January 1, 2020 to December 31, 2022.
Therefore, Himawari L1 gridded data with the same resolution at the corresponding time are downloaded in section~\ref{section:RS}.

\begin{figure}[!htp]
    \centering
    \includegraphics[width=0.8\textwidth,trim=0 0 0 0,clip]{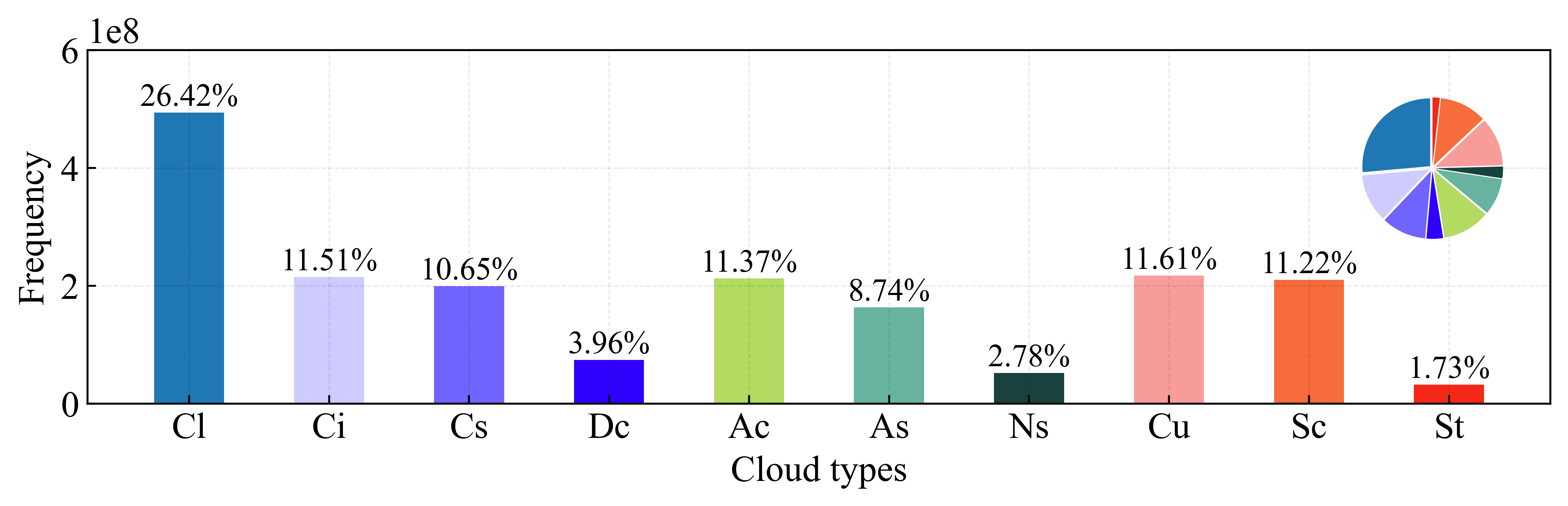}
    \caption{The frequency histogram of cloud types at 03:00 (UTC+0) every 3 d from January 2020 to December 2022.}
    \label{fig:Cld_cls_freq}
\end{figure}

The frequency of each cloud type is calculated, as shown in Fig.~\ref{fig:Cld_cls_freq}.
The frequency of clear sky (Cl) in the entire satellite observation area is the highest, while the frequency of St is the lowest.
Cl accounts for 26.42\%, St accounts for 1.73\%, and their ratio is approximately 15:1.
The proportion of all types of clouds is 73.58\%.

\section{Methodology}
\label{section:Methodology}
\subsection{Knowledge-based data-driven framework}
\begin{figure}[!htp]
    \centering
    \includegraphics[width=0.83\textwidth,trim=0 0 0 -20,clip]{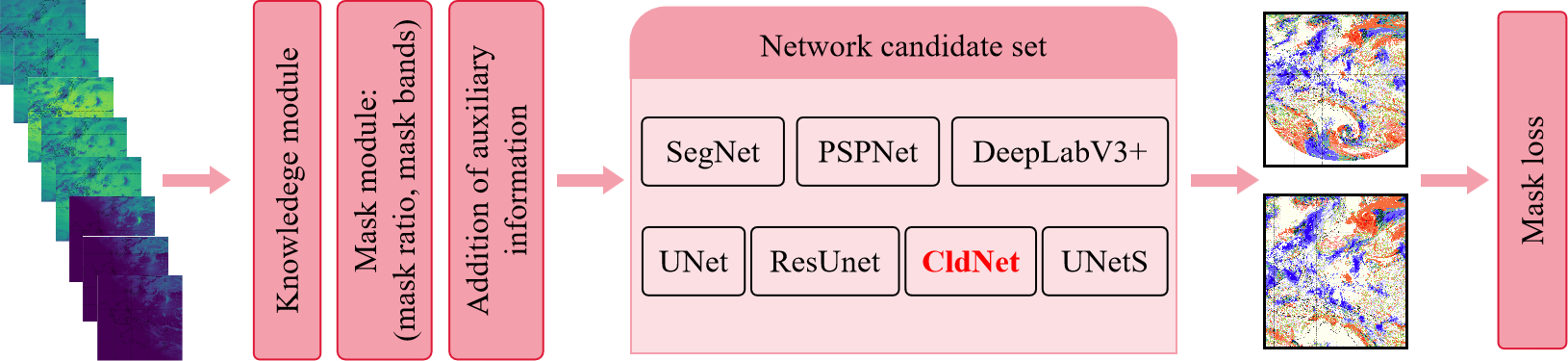}
    \caption{The architecture of the knowledge-based data-driven (KBDD) framework is composed of knowledge module, mask module, addition of auxiliary information, network candidate set, and mask loss. And the network candidate set includes SegNet, PSPNet, DeepLabV3+, UNet, ResUnet, CldNet, and UNetS.}
    \label{fig:KBDD}
\end{figure}

In order to derive all-day identification of cloud types through spectral information from Himawari-8/9 satellite sensors, a knowledge-based data-driven (KBDD) framework is proposed, whose specific architecture is depicted in Fig.~\ref{fig:KBDD}.
The KBDD framework mainly consists of knowledge module, mask module, addition of auxiliary information, network candidate set, and mask loss.

The main function of the knowledge module is to reorganize spectral information from Himawari-8/9 satellite sensors based on existing research knowledge.
From previous research knowledge, it has been found that cloud types are not only related to single channel remote sensing, but also to the differences between two different channels.
Here, it should be noted that satellite spectral channels B01 to B06 are characterized by albedo, while satellite spectral channels B07 to B16 are characterized by brightness temperature.
Since albedo and brightness temperature are distinct variables and their units are different, the pairwise combinations must have the same dimension in order to make a difference.
The knowledge module completed the reorganization of the satellite spectral information of B01 $\sim$ B16, the differences between the pairwise combinations of B01 $\sim$ B06, and the differences between the pairwise combinations of B07 $\sim$ B16.

The mask module is a convenient way to mask certain spectral channels and increase the flexibility of the framework, which has the ability to train with/without VIS and NIR data but does not require changing the model structure.
When not using VIS and NIR channels, the mask module can set the satellite spectral information of B01 $\sim$ B06 and the differences between the pairwise combinations of B01 $\sim$ B06 to zero through the parameters mask ratio and mask bands.

The addition of auxiliary information is to improve the performance of the framework.
The auxiliary information includes satellite zenith angle (SAZ), satellite azimuth angle (SAA), solar zenith angle (SOZ), solar azimuth angle (SOA), longitude, latitude, altitude, and underlying surface attributes (land or water).
The auxiliary information and the reorganized satellite spectral information are merged through concatenation.
The merged data is input into the network candidate set to obtain the probability result of cloud types.
Due to the fact that the cloud-type references in the nighttime area do not exist, the mask loss mainly calculates the loss of labeled pixels in the target image, excluding unlabeled pixels.

\subsection{SegNet, PSPNet, DeepLabV3+, UNet and ResUnet}
\label{section:SegNet}
The networks SegNet, PSPNet, DeepLabV3+, UNet and ResUnet are adopted as the most commonly used semantic segmentation networks in this study.
SegNet is proposed by~\citet{SegNet}, and its network structure is a convolutional encoder-decoder.
It mainly uses 2D convolution, 2D max pooling, and 2D max upsampling pooling.
The input pooling indices of 2D max upsampling pooling come from the corresponding output of 2D max pooling.
The input and output features of 2D convolution depend on the input and output data, respectively.
The output feature number of the last convolution is the number of cloud-type categories.
Under the function of softmax and argmax, the cloud type corresponding to the pixel can be obtained.

PSPNet is proposed by~\citet{PSPNet} and ResNet50~\citep{ResNet} is used as a feature map extractor for PSPNet in this study.
The size of the feature map is reduced to 1/4 of the original input size to conserve memory.
The feature maps in different levels are generated by the pyramid pooling module, and all feature maps are concatenated to predict cloud types.

DeepLabV3+ is proposed by~\citet{DeepLabV3plus} and the feature map extractor of DeepLabV3+ is the same as that of PSPNet.
In encoder processing, the encoder mainly uses four convolutional kernels with different dilation parameters and one global average pooling.
In decoder processing, the low-level features obtained by the feature map extractor and the features obtained by the encoder are concatenated to predict cloud types.

\citet{UNet} designed UNet.
UNet is widely used in the field of imaging~\citep{WALDNER2020111741, rs15092246, YOO2022102678, AMINIAMIRKOLAEE2022113014, rs15061706}, and it has significant similarities with SegNet.
The main difference is that UNet concatenates the feature maps corresponding to the downsampling process during the upsampling process, and ultimately predicts cloud types.
\citet{ResUnet} developed ResUnet, which is built with residual units and has similar architecture to that of UNet.

\subsection{CldNet}
\label{section:CldNet}
\begin{figure}[!htp]
    \centering
    \includegraphics[width=0.86\textwidth,trim=0 0 0 0,clip]{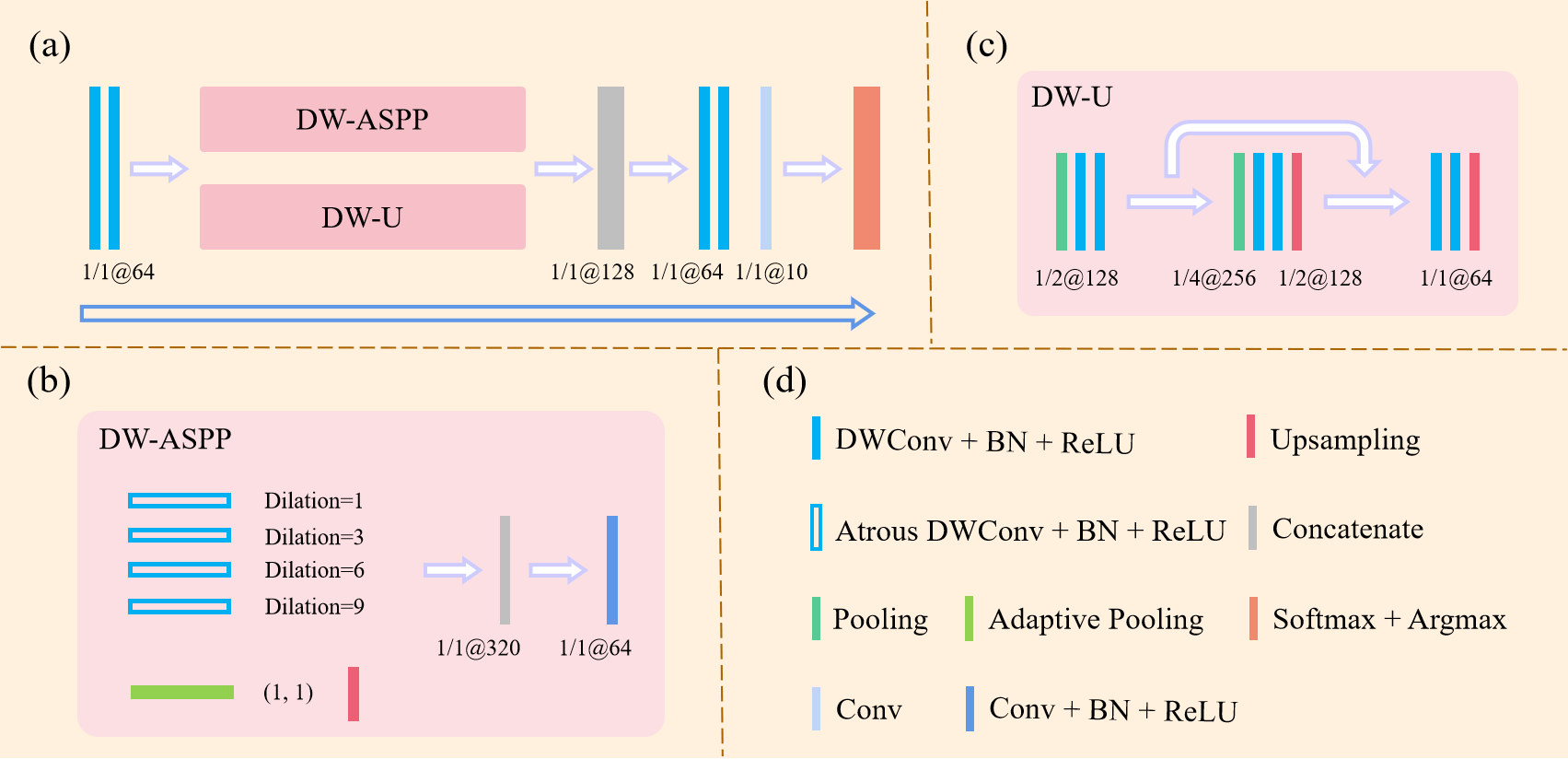}
    \caption{The specific network architecture of CldNet.}
    \label{fig:CldNet}
\end{figure}

In this study, a novel simple and effective deep learning-based network for cloud-type classification, named CldNet, is proposed and illustrated in Fig.~\ref{fig:CldNet}.
Depthwise separable convolution (DWConv) is applied into CldNet, which factorizes a standard convolution into a depthwise convolution~\citep{Xception8099678} followed by a pointwise convolution~\citep{Pointwise8578207}.
The difference between atrous DWConv and DWConv is the dilation of depthwise convolution.
CldNet mainly consists of a DW-ASPP module and a DW-U module in Fig.~\ref{fig:CldNet}.
The DW-ASPP module and DW-U module are inspired by the atrous spatial pyramid pooling (ASPP) module of DeepLabV3+ and UNet, respectively.
The DW-ASPP module uses four atrous depthwise separable convolutions with different dilations to capture useful multi-scale spatial context, and this can enhance the receptive field of the feature map.
The DW-U module can extract feature maps at different levels.
The fusion of the two extracted feature maps is beneficial for mining the inherent relationships between model input data and cloud types and achieving cloud-type recognition.
More importantly, the two modules of CldNet have fewer parameters, which saves memory and makes them easier to deploy on edge devices.

\subsection{Loss function}
The probability distribution of cloud-type prediction for each pixel is expected to be consistent with that of the cloud-type reference.
Cross entropy can measure the difference information between two probability distributions.
In the multi classification problem of deep learning, cross entropy can be replaced by negative logarithmic likelihood to calculate model loss.
The loss update model parameters through backpropagation to achieve optimal performance of the model.
Negative logarithmic likelihood loss (NLLLoss) is computed by Eq.~\ref{eq:NLLLoss} and Eq.~\ref{eq:NLLLoss_prob}, and the label of cloud type is calculated by Eq.~\ref{eq:NLLLoss_label}.

\begin{equation}
    \label{eq:NLLLoss}
    NLLLoss = \frac{1}{N}\sum_{i=1}^{N}\sum_{j=0}^{C} -Cld_{ref,ij}^{prob}\ln\left(Cld_{pre,ij}^{prob}\right)
    =-\frac{1}{N}\sum_{i}^{N}\ln\left(Cld_{pre,i[label]}^{prob}\right)
\end{equation}

\begin{equation}
    \label{eq:NLLLoss_prob}
    Cld_{pre, i}^{prob}=softmax\left(x\right)_{i}=\frac{e^{x_{ij}}}{\sum_{j=0}^{C}e^{x_{ij}}}
\end{equation}

\begin{equation}
    \label{eq:NLLLoss_label}
    Cld_{pre, i}^{label}=argmax\left(Cld_{pre, i}^{prob}\right)
\end{equation}
where $N$ is the number of samples per batch;
$C$ is the number of cloud-type labels;
$Cld_{ref}^{prob}$ and $[label]$ is the probability and label of the reference cloud type, respectively;
$Cld_{pre}^{prob}$ and $Cld_{pre}^{label}$ are the probability and label of the model's predicted results, respectively;
$x$ is the output result of the network structure;
and the function $argmax$ returns the indices of the maximum value of all elements in the input vector.

\subsection{Experimental setup}
This section will provide a detailed introduction about experimental setup.
Satellite spectral data and cloud-type data from 2020 and 2021 are used as the training dataset, and those from 2022 are used as the test dataset.
During the training process, the training dataset is randomly divided into five parts.
Four of these parts are used for training, whose main purpose is to update the parameters of the model.
The remaining part is used for validation, whose purpose is to prevent overfitting by early stopping of the model training~\citep{Dietterich1995} when the loss on the validation dataset no longer decreases after 10 consecutive epochs.
The test dataset is used to evaluate the performance of the trained model.

All models in this study are trained on single NVIDIA GPU A100.
In each training batch, single remote sensing image is segmented into $5 \times 5$ slices and input into the model, and its dimension is $25 \times 76 \times 480 \times 480$.
The batch size is 25.
Optimization algorithm Adam is used and its learning rate (lr) is controlled through a multi-step learning rate approach.
If epoch $<$ 5, lr = 0.01; If 5 $\le$ epoch $<$ 10, lr = 0.001; If 10 $\le$ epoch $<$ 20, lr = 0.0001; If 20 $\le$ epoch $<$ 30, lr = 0.00001; And if epoch $\ge$ 30, lr = 0.000001.

\subsection{Quantitative evaluation metrics}
The recognition of nine cloud types and clear sky is a multi classification problem. In order to better evaluate the predictive classification performance of the model, accuracy is a commonly used indicator. $\mathrm{Accuracy}$ can measure the ratio of correctly classified predictions to the total number, as follows:

\begin{equation}
    \label{eq:Accuracy}
    \mathrm{Accuracy} = \frac{\sum_{j=0}^{C} \mathrm{TP}_{j}}{N}
\end{equation}
In order to understand the model's ability to distinguish between clouds and clear skies, the indicator accuracy (N/Y for cloud) is defined as follows:

\begin{equation}
    \label{eq:Accuracy_NY}
    \mathrm{Accuracy\;(N/Y\;for\;cloud)} = \frac{\sum_{k=0}^{1} \mathrm{TP}_{k}}{N}
\end{equation}
$\mathrm{Precision}$ focuses on evaluating the proportion of ture positive data among all predicted positive data.
$\mathrm{Recall}$ focuses on evaluating how much data has been successfully predicted as positive among all positive data.
For $\mathrm{precision}$ and $\mathrm{recall}$, each class needs to calculate its precision and recall separately.
$\mathrm{F_{1}\mbox{-}score}$ is a comprehensive indicator of both $\mathrm{precision}$ and $\mathrm{recall}$, as follows:

\begin{equation}
    \label{eq:Recall_j}
    \mathrm{Recall} _{j}=\frac{\mathrm{TP} _{j}}{\mathrm{TP} _{j}+\mathrm{FN} _{j}}
\end{equation}

\begin{equation}
    \label{eq:Precision_j}
    \mathrm{Precision} _{j}=\frac{\mathrm{TP} _{j}}{\mathrm{TP} _{j}+\mathrm{FP} _{j}}
\end{equation}

\begin{equation}
    \label{eq:F1-score_j}
    \mathrm{F_{1}\mbox{-}score} _{j}=2\frac{\mathrm{Recall} _{j}\times \mathrm{Precision} _{j}}{\mathrm{Recall} _{j}+\mathrm{Precision} _{j}}
\end{equation}
$\mathrm{F_{1}\mbox{-}score_{macro}}$ directly adds up the $\mathrm{F_{1}\mbox{-}score}$ of different classes to calculate the average, which can treat each class equally.

\begin{equation}
    \label{eq:F1-score_macro}
    \mathrm{F_{1}\mbox{-}score_{macro}} =\frac{\sum_{j}^{C} \mathrm{F_{1}\mbox{-}score} _{j}}{C}
\end{equation}
$\mathrm{F_{1}\mbox{-}score_{weight}}$ is the sum of the $\mathrm{F_{1}\mbox{-}score}$ of each class multiplied by its weight, and this method considers class imbalance issues.

\begin{equation}
    \label{eq:F1-score_weight}
    \mathrm{F_{1}\mbox{-}score_{weight}}  = \mathrm{F_{1}\mbox{-}score} _{j}\times \mathrm{R} _{j}
\end{equation}
$\mathrm{F_{1}\mbox{-}score_{micro}}$ adds the TP, FP, and FN of each class first, and then calculates them based on the binary classification.

\begin{equation}
    \label{eq:Recall_m}
    \mathrm{Recall_{m}}  = \frac{\sum_{j=0}^{C} \mathrm{TP}_{j}}{\sum_{j=0}^{C} \left(\mathrm{TP} _{j}+\mathrm{FN} _{j} \right)}
\end{equation}

\begin{equation}
    \label{eq:Precision_m}
    \mathrm{Precision_{m}} = \frac{\sum_{j=0}^{C} \mathrm{TP} _{j}}{\sum_{j=0}^{C} \left(\mathrm{TP} _{j}+\mathrm{FP} _{j} \right)}
\end{equation}

\begin{equation}
    \label{eq:F1-score_micro}
    \mathrm{F_{1}\mbox{-}score_{micro}} =2\frac{\mathrm{Recall_{m}} \times \mathrm{Precision_{m}} }{\mathrm{Recall_{m}} +\mathrm{Precision_{m}} }
\end{equation}
where TP is true positive; FN is false negative; FP is false positive; and TN is true negative.
True and false refer to the correctness of the test result.
True means that the test result is correct, and false means that the test result is incorrect.
Positive and negative refer to the test results of the sample.
Positive means that the intended target is detected, and negative means that the intended target is not detected.
$N$ is the total number of samples in sigle image.
$C$ is the number of cloud-type labels.
$\mathrm{R}_{j}$ is the true distribution proportion of the class $j$.

\section{Results}
\label{section:Results}
\subsection{Performance of different models} 
\label{section:models_performance}
\subsubsection{Comparison of all models} 
\label{section:models_comparison}

\begin{figure}[!htp]
    \vspace{5mm}
    \centering
    \includegraphics[width=0.95\textwidth,trim=0 0 0 0,clip]{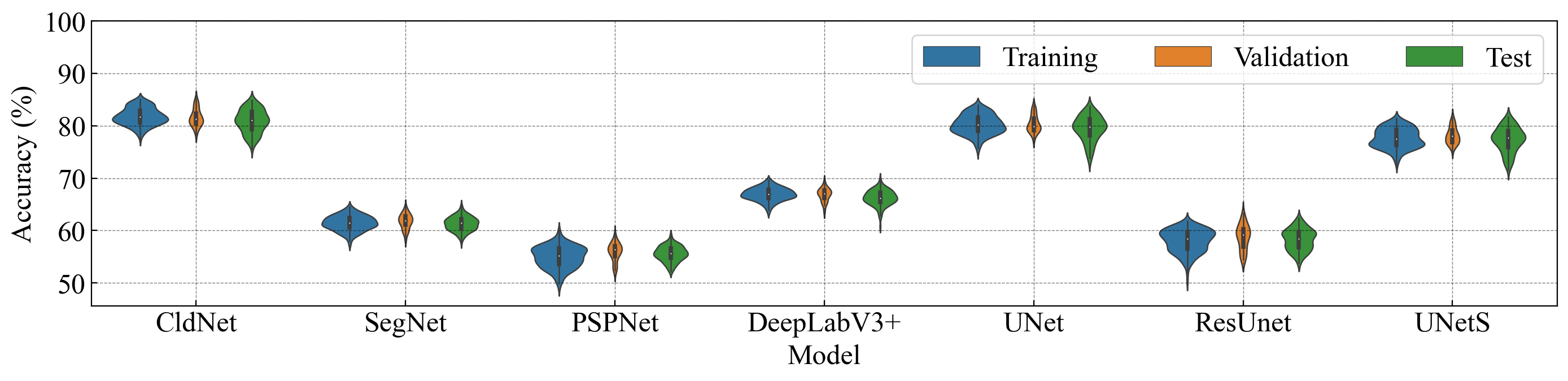}
    \caption{The accuracy of all models for the converged epoch at different stages.}
    \label{fig:ModelAcc_All_violinplot}
\end{figure}

The networks with different structures, including CldNet, SegNet, PSPNet, DeepLabV3+, UNet and ResUnet, are used to train network parameters through the training dataset.
The accuracy of each batch under each epoch is recorded, and Fig.~\ref{fig:ModelAcc_All_violinplot} compares the accuracy of different models for the converged epoch.
The accuracy performance of each model in the three stages of training, validation, and test is very close, indicating that the model possesses good stability and robustness.
Our proposed model CldNet is state-of-the-art in identifying cloud types, which achieves average accuracies of 81.76±1.64\%, 81.58±1.63\%, and 80.89±2.18\% during the training, validation, and test stages, respectively.
PSPNet is the worst, which achieves average accuracies of 54.95±2.06\%, 55.92±1.70\%, and 55.58±1.43\% during the training, validation, and test stages, respectively.
The accuracy of CldNet is only approximately 1.4\% higher than that of UNet, but the total parameters (0.46M) of CldNet are much smaller than the total parameters (31.09M) of UNet.
The number of features in the UNet backbone structure is reduced to a quarter of its original number, and the network is marked as UNetS.
The total parameters of UNetS are reduced to 1.96M, and it has an accuracy of 77.37±2.34\% on the test set.
The accuracy of CldNet is approximately 3.5\% higher than that of UNetS.
The mean, standard deviation, minimum, median, and maximum of accuracy for all models during the training, validation, and test stages are recorded in Table~\ref{table:model_acc_summary}.

\begin{figure}[!htp]
    \vspace{5mm}
    \centering
    \begin{overpic}[abs,unit=1mm,width=0.95\textwidth,trim=0 0 0 0,clip]{
            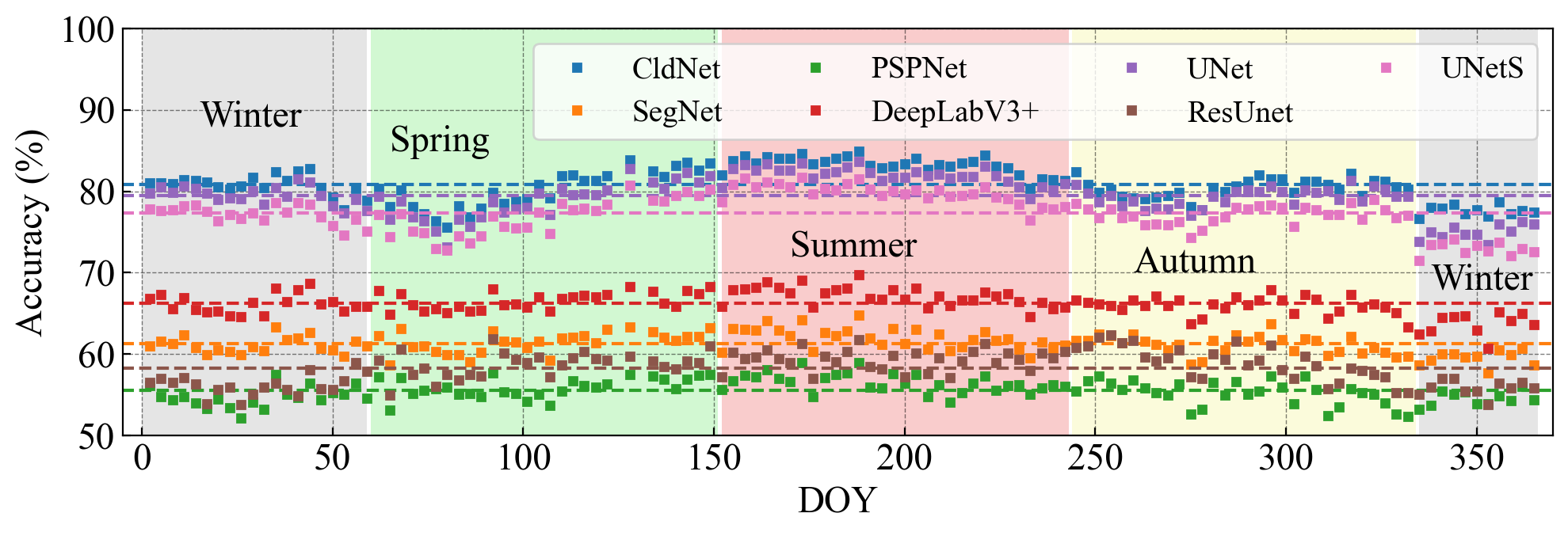
        }
    \end{overpic}
    \caption{The overall accuracy trend of all models on the test dataset. DOY means the day of the year.}
    \label{fig:ModelAcc_All_scatter}
\end{figure}

The overall accuracy trend of all models on the test dataset is shown in Fig.~\ref{fig:ModelAcc_All_scatter}, and DOY refers to the day of the year.
Basically, all models maintain the same pattern (CldNet $>$ UNet $>$ UNetS $>$ DeepLabV3+ $>$ SegNet $>$ ResUnet $>$ PSPNet) of testing accuracy on the same day.
The average accuracies of the cloud-type distributions for CldNet, SegNet, PSPNet, DeepLabV3+, UNet, ResUnet, and UNetS are 80.89±2.18\%, 61.30±1.32\%, 55.58±1.43\%, 66.25±1.45\%, 79.50±2.37\%, 58.27±2.00\%, and 77.37±2.34\%, respectively.
Compared with SegNet, PSPNet, DeepLabV3+, UNet, ResUnet and UNetS, our proposed model CldNet has increased by 32\%, 46\%, 22\%, 2\%, 39\% and 5\%, respectively.

Starting from December 2022, the spectral information released by JAXA comes from the Himawari-9 sensors, while the spectral information before December 2022 comes from the Himawari-8 sensors.
The trained parameters of the model are based on the spectral data from the Himawari-8 sensor. Although both the sensors of Himawari-8 and Himawari-9 are the Advanced Himawari Imager, the accuracy of cloud-type prediction using the spectral data from Himawari-9 sensor in the trained model still decreases in Fig.~\ref{fig:ModelAcc_All_scatter}.
In 2022, March, April, and May belong to spring; June, July, and August belong to summer; September, October, and November belong to autumn; December, January, and February belong to winter.
For CldNet, the accuracy for spring, summer, autumn, January and February of winter, and December of winter is 80.17±2.36\%, 83.26±1.08\%, 80.36±1.13\%, 80.81±1.20\%, and 77.62±0.61\%.
This result indicates that the model has an advantage in identifying cloud types in summer.

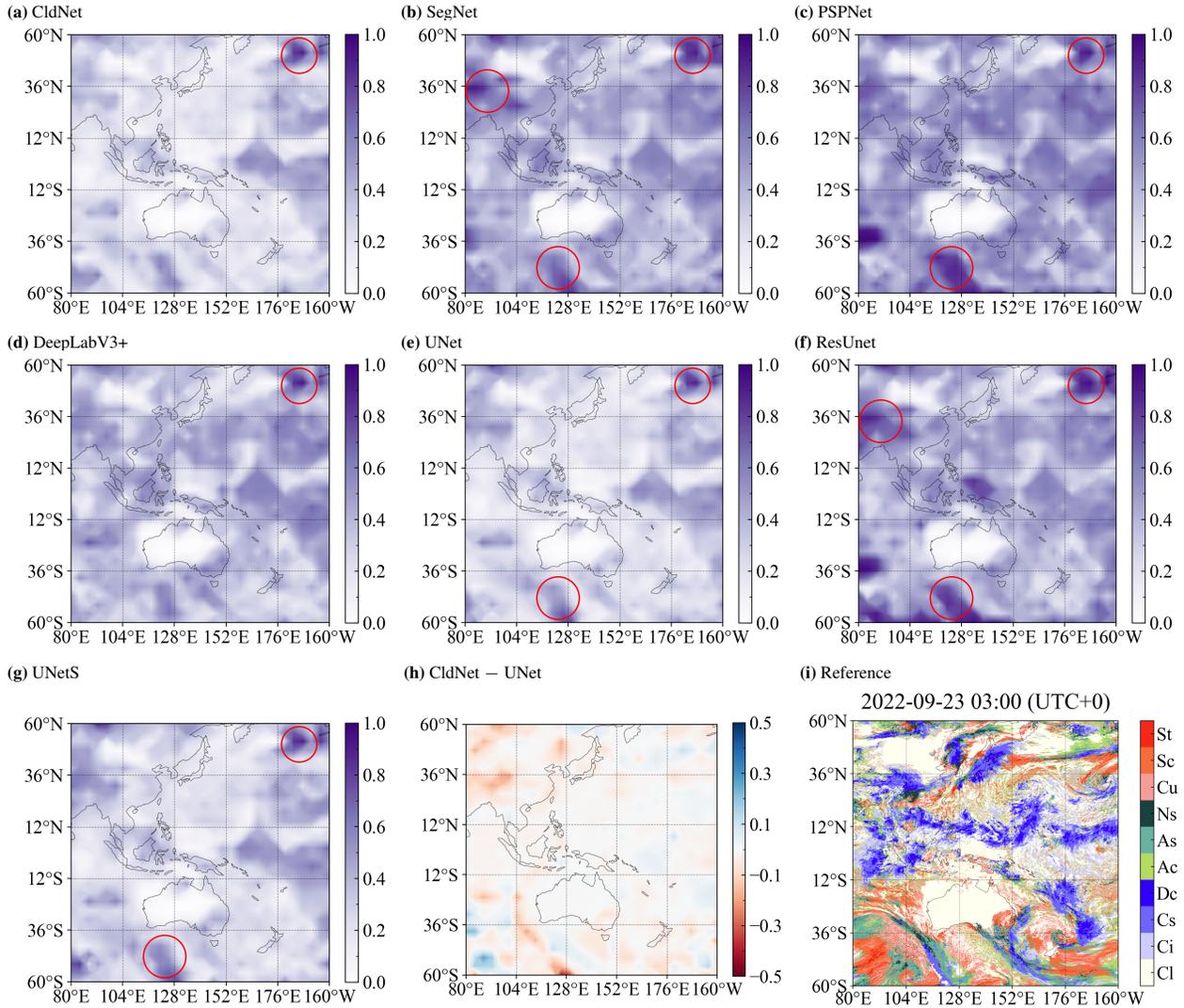
\begin{figure}[!htp]
    \centering
    \subcaptionbox{CldNet\vspace{-2mm}\label{fig:cldDD_all:CldNet}}{
        \begin{overpic}[abs,unit=1mm,width=0.32\textwidth,trim=0 0 0 0,clip]{
                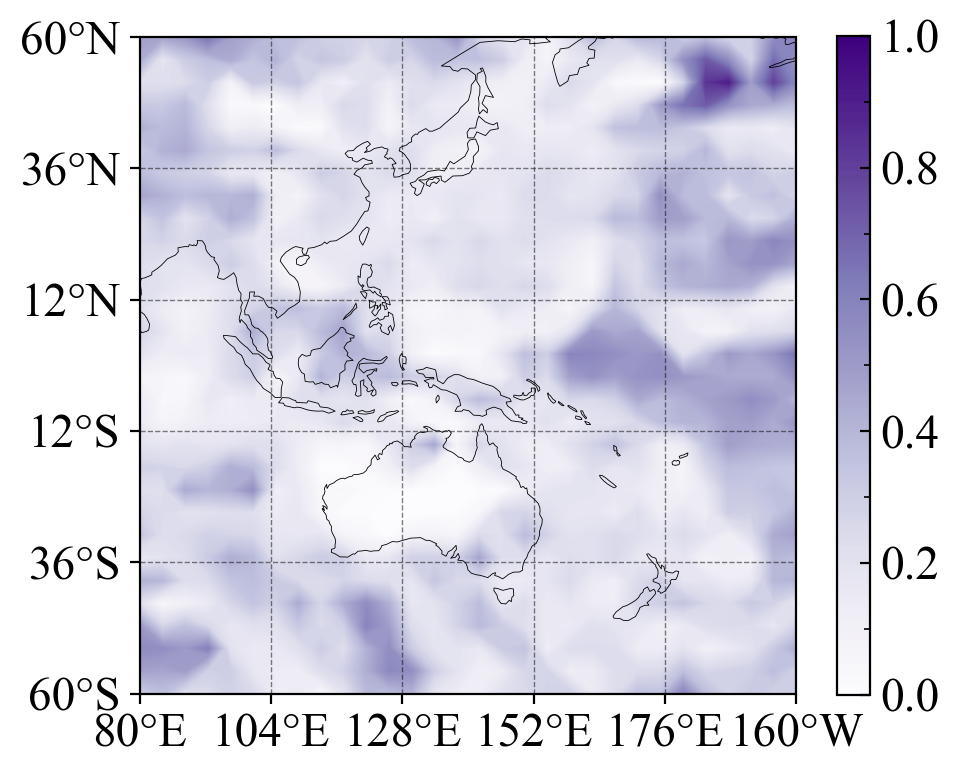
            }
            \color{red}
            \linethickness{0.2mm}
            \put(39,38){
                \circle{5}
            }
        \end{overpic}
    }
    \hfill
    \subcaptionbox{SegNet\vspace{-2mm}\label{fig:cldDD_all:SegNet}}{
        \begin{overpic}[abs,unit=1mm,width=0.32\textwidth,trim=0 0 0 0,clip]{
                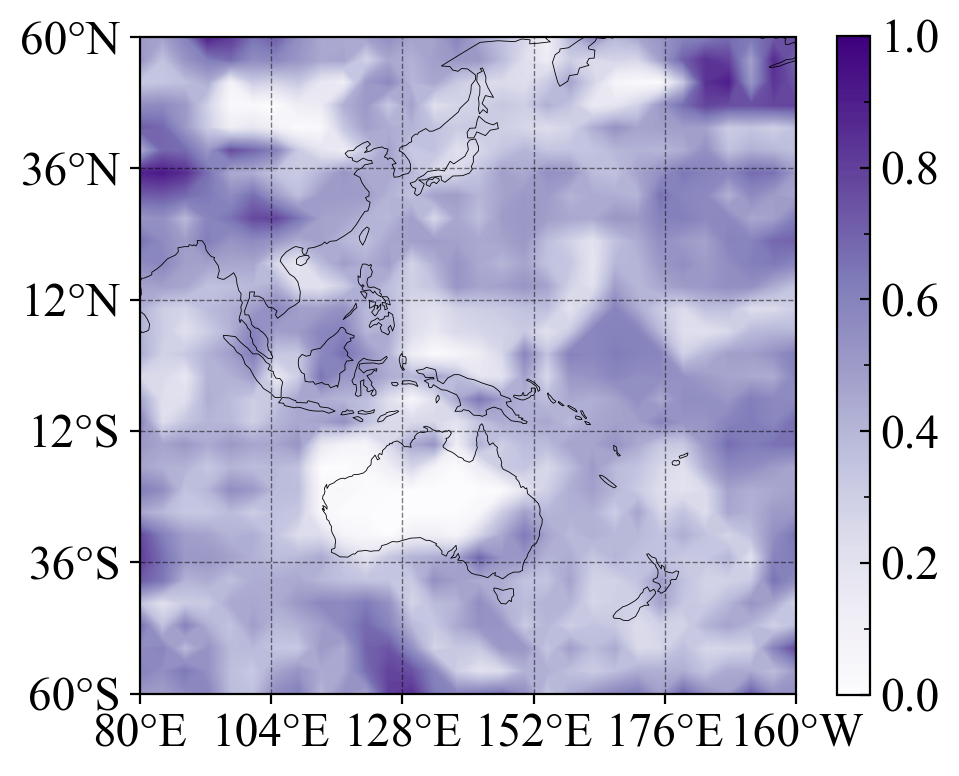
            }
            \color{red}
            \linethickness{0.2mm}
            \put(39,38){
                \circle{5}
            }
            \put(20,8){
                \circle{6}
            }
            \put(10,33){
                \circle{6}
            }
        \end{overpic}
    }
    \hfill
    \subcaptionbox{PSPNet\vspace{-2mm}\label{fig:cldDD_all:PSPNet}}{
        \begin{overpic}[abs,unit=1mm,width=0.32\textwidth,trim=0 0 0 0,clip]{
                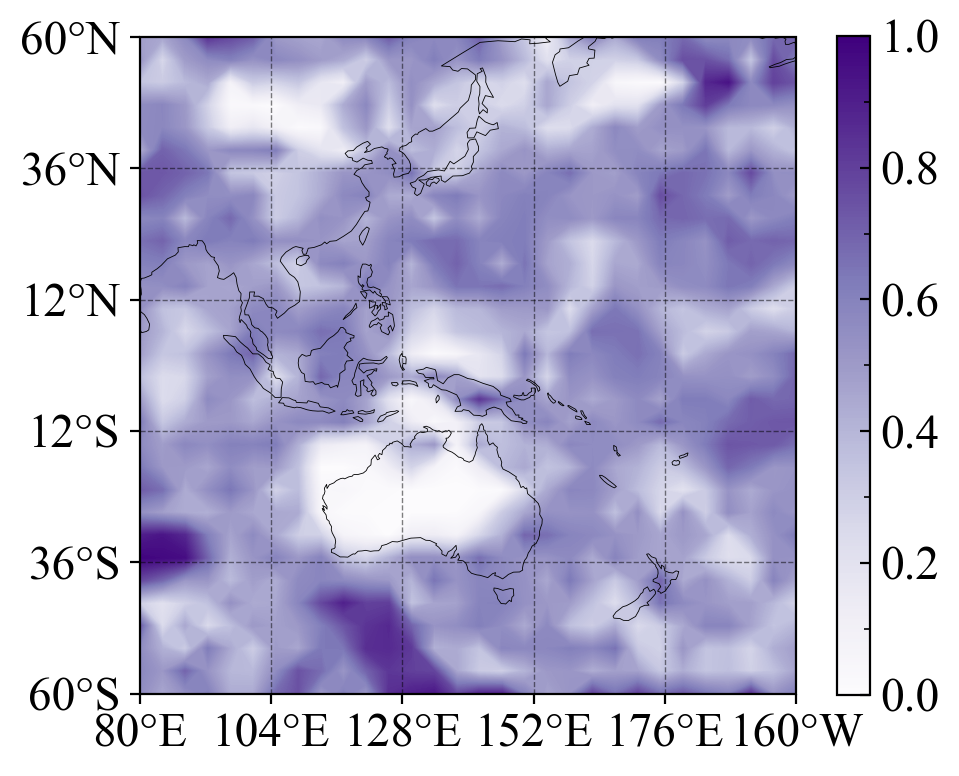
            }
            \color{red}
            \linethickness{0.2mm}
            \put(39,38){
                \circle{5}
            }
            \put(20,8){
                \circle{6}
            }
        \end{overpic}
    }
    \\
    \subcaptionbox{DeepLabV3+\vspace{-2mm}\label{fig:cldDD_all:DeepLabV3plus}}{
        \begin{overpic}[abs,unit=1mm,width=0.32\textwidth,trim=0 0 0 0,clip]{
                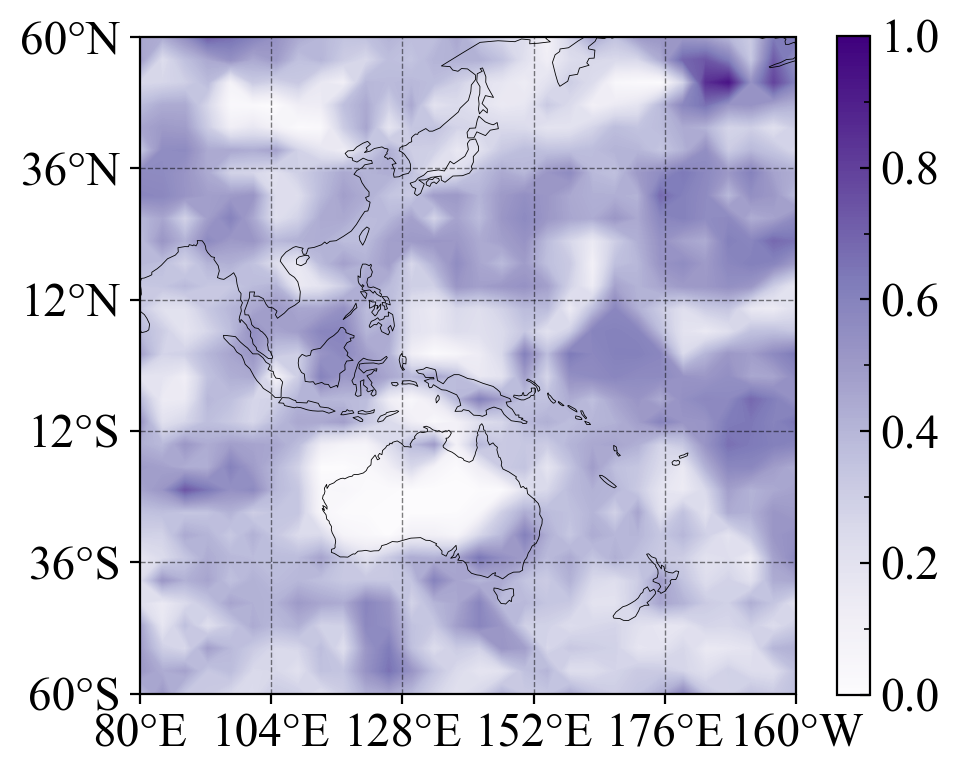
            }
            \color{red}
            \linethickness{0.2mm}
            \put(39,38){
                \circle{5}
            }
        \end{overpic}
    }
    \hfill
    \subcaptionbox{UNet\vspace{-2mm}\label{fig:cldDD_all:UNet}}{
        \begin{overpic}[abs,unit=1mm,width=0.32\textwidth,trim=0 0 0 0,clip]{
                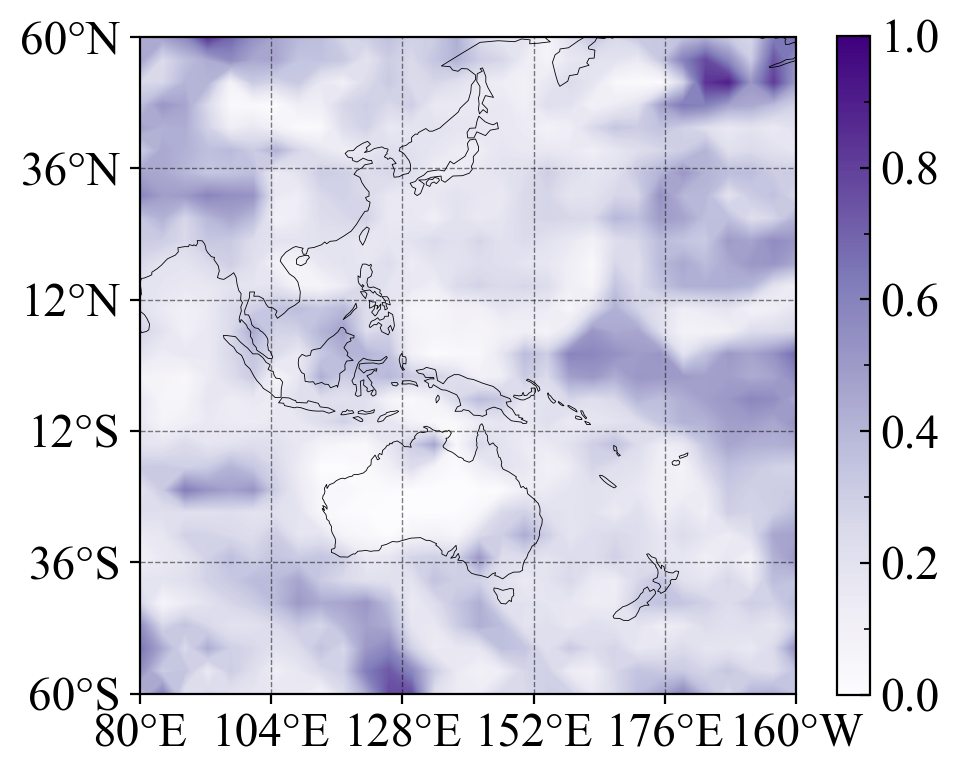
            }
            \color{red}
            \linethickness{0.2mm}
            \put(39,38){
                \circle{5}
            }
            \put(20,8){
                \circle{6}
            }
        \end{overpic}
    }
    \hfill
    \subcaptionbox{ResUnet\vspace{-2mm}\label{fig:cldDD_all:ResUnet}}{
        \begin{overpic}[abs,unit=1mm,width=0.32\textwidth,trim=0 0 0 0,clip]{
                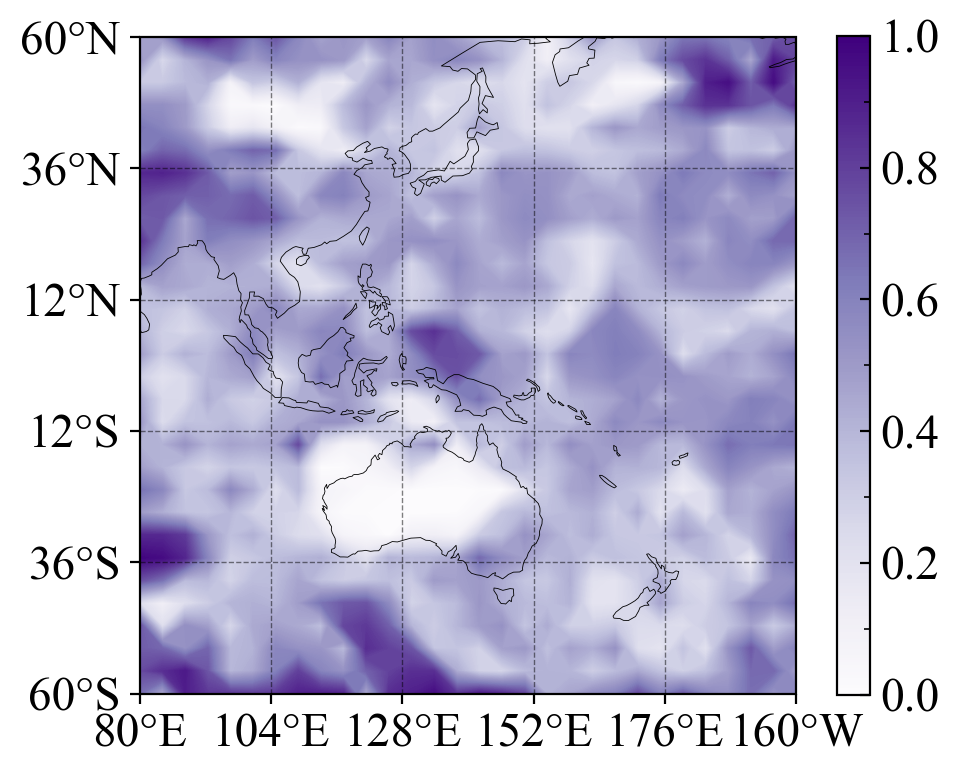
            }
            \color{red}
            \linethickness{0.2mm}
            \put(39,38){
                \circle{5}
            }
            \put(20,8){
                \circle{6}
            }
            \put(10,33){
                \circle{6}
            }
        \end{overpic}
    }
    \\
    \subcaptionbox{UNetS\vspace{-2mm}\label{fig:cldDD_all:UNetS}}{
        \begin{overpic}[abs,unit=1mm,width=0.32\textwidth,trim=0 0 0 -20,clip]{
                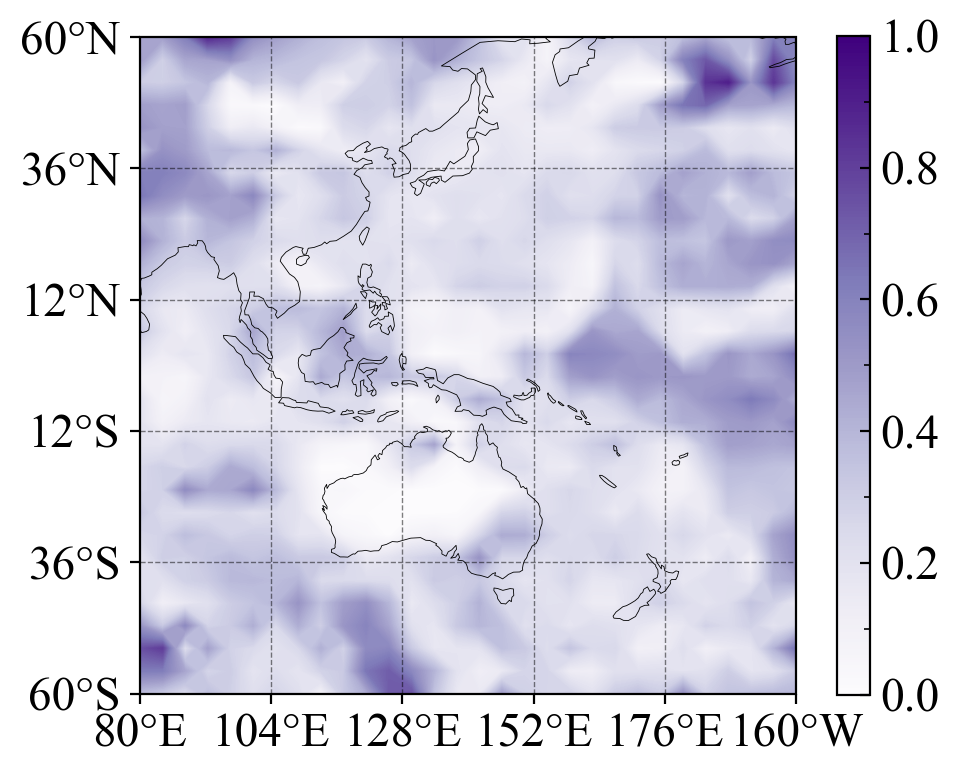
            }
            \color{red}
            \linethickness{0.2mm}
            \put(39,38){
                \circle{5}
            }
            \put(20,8){
                \circle{6}
            }
        \end{overpic}
    }
    \hfill
    \subcaptionbox{CldNet $\mathrm{-}$ UNet\vspace{-2mm}\label{fig:cldDD_all:cldDDA}}{
        \begin{overpic}[abs,unit=1mm,width=0.32\textwidth,trim=0 0 0 -25,clip]{
                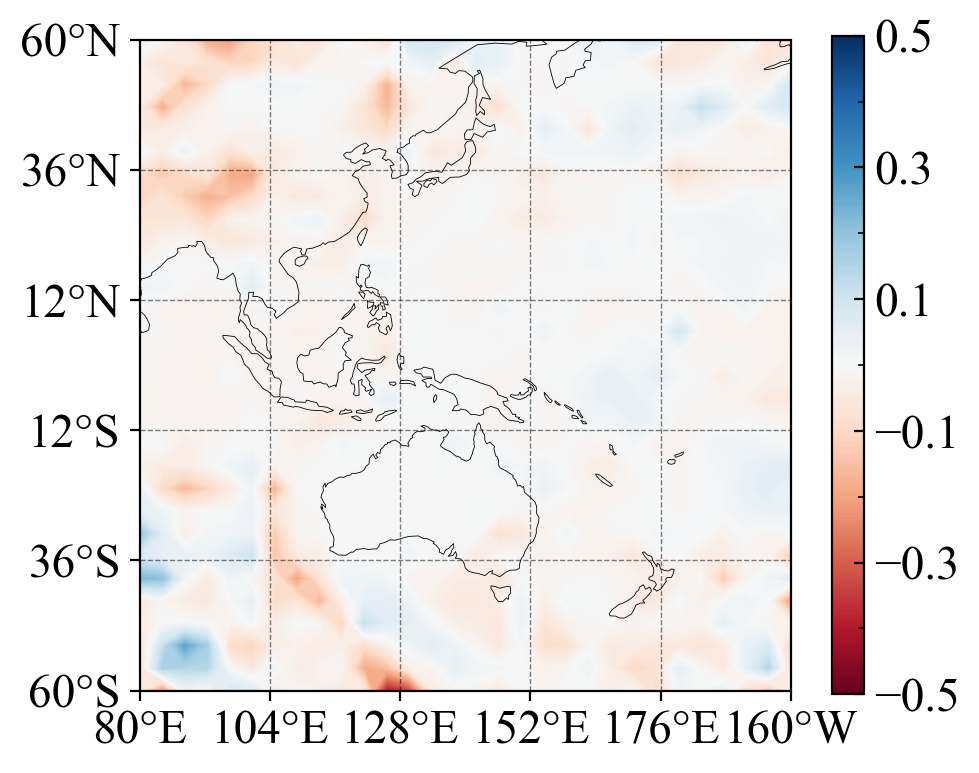
            }
        \end{overpic}
    }
    \hfill
    \subcaptionbox{Reference\vspace{-2mm}}{%
        \includegraphics[width=0.32\textwidth,trim=0 0 30 0,clip]{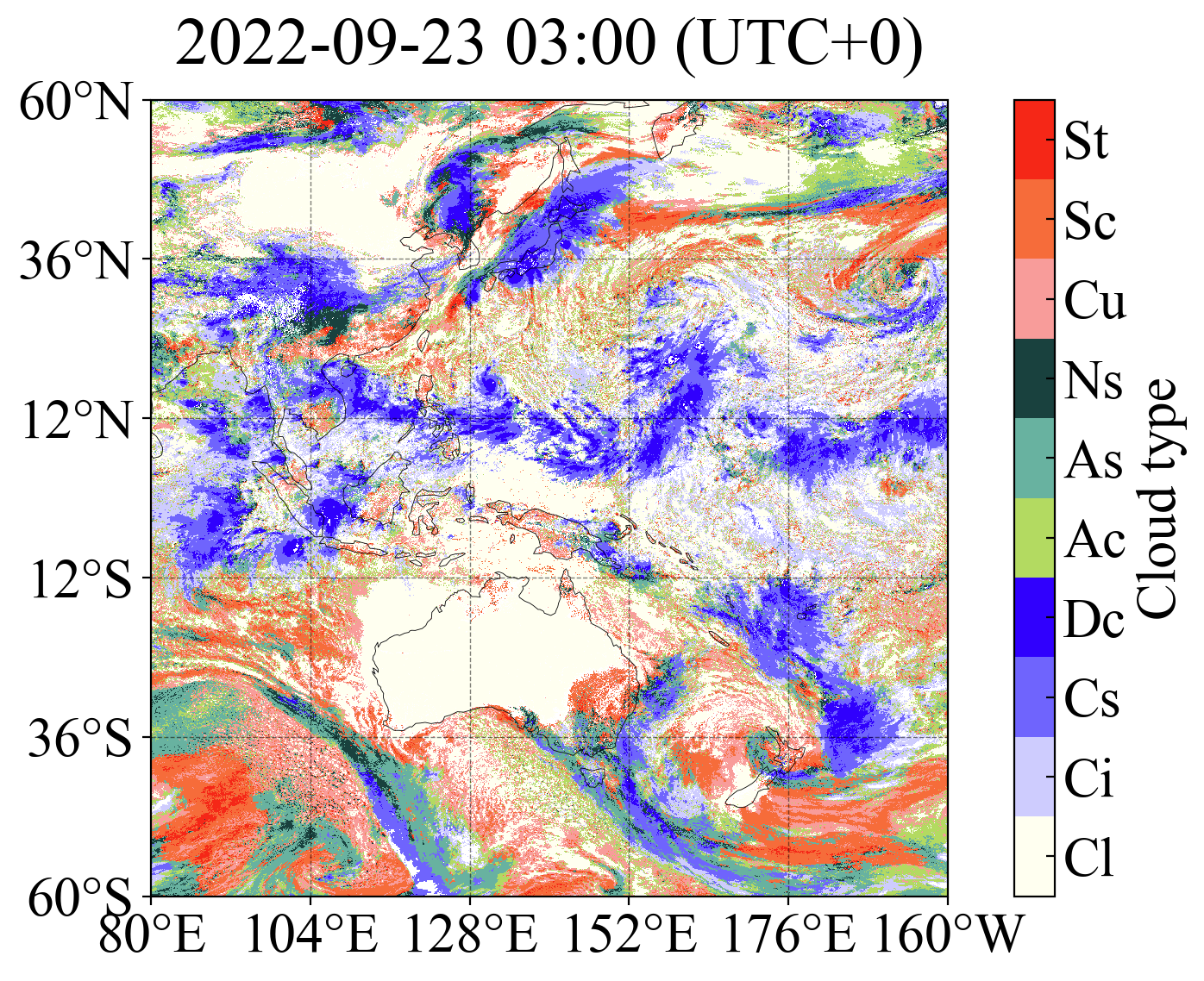}
    }
    \caption{
        The cloud-type prediction error density distributions of (a) CldNet, (b) SegNet, (c) PSPNet, (d) DeepLabV3+, (e) UNet, (f) ResUnet and (g) UNetS, (h) the differences between CldNet and UNet in terms of the cloud-type prediction error density, and (i) the cloud-type distributions from JAXA's cloud-type products as a reference at 2022-09-23 03:00 (UTC+0).
    }
    \label{fig:cldDD_all}
\end{figure}

The satellite observation area is divided into 30$\times$30 sub-areas, and the cloud-type prediction error density of each sub-area is defined as the proportion of pixels with cloud-type prediction errors in that sub-area to the total pixels.
The cloud-type prediction error density distributions of CldNet, SegNet, PSPNet, DeepLabV3+, UNet, ResUnet and UNetS at 2022-09-23 03:00 (UTC+0) are shown in Fig.~\ref{fig:cldDD_all:CldNet} - \ref{fig:cldDD_all:UNetS}.
The lighter is the color in the figure, the higher is the prediction accuracy of the model.
The areas marked with circles in the figure are the areas where the model's cloud-type prediction ability is poor.
From the perspective of overall cloud-type prediction error density distributions, our proposed model CldNet is significantly better than SegNet, PSPNet, DeepLabV3+, and ResUnet.
The differences between CldNet and UNet in terms of the cloud-type prediction error density are computed in Fig.~\ref{fig:cldDD_all:cldDDA}, where a negative value indicates that CldNet is better than UNet, while a positive value indicates that UNet is better than CldNet.
Most areas are positive, indicating that CldNet is better than UNet.
Among all the models, our proposed model CldNet is state-of-the-art (SOTA) in cloud-type recognition.

\subsubsection{Performance of CldNet across cloud types and clear sky} 
\label{section:Performance_of_CldNet_cloudtypes}
\begin{figure}[!htp]
    \vspace{5mm}
    \centering
    \subcaptionbox{Precision\vspace{-2mm}\label{fig:CldNet_Stat_Radar:Precision}}{%
        \includegraphics[width=0.25\textwidth,trim=0 0 0 0,clip]{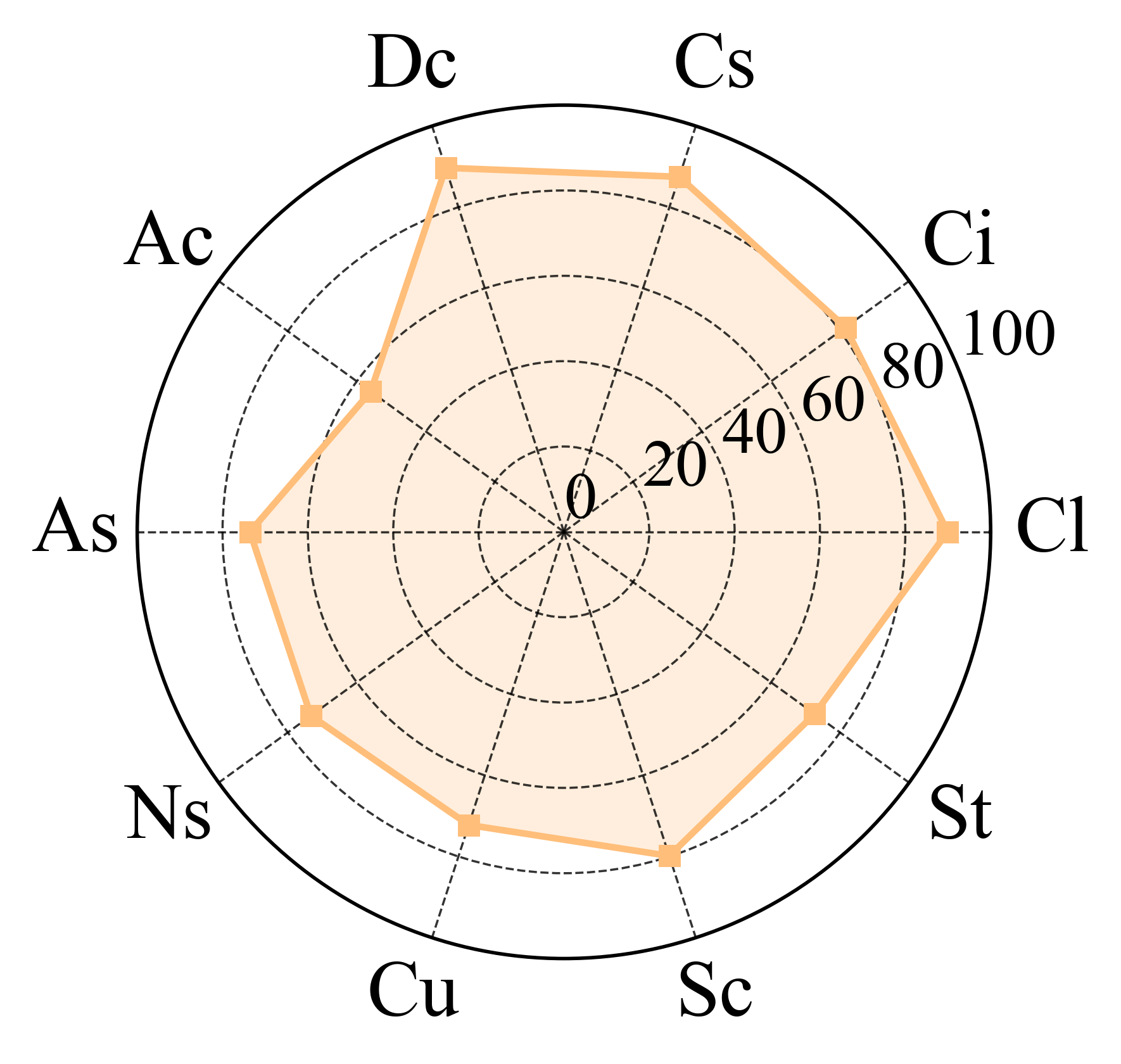}
    }
    \hspace{3.5mm}
    \subcaptionbox{Recall\vspace{-2mm}\label{fig:CldNet_Stat_Radar:Recall}}{%
        \includegraphics[width=0.25\textwidth,trim=0 0 0 0,clip]{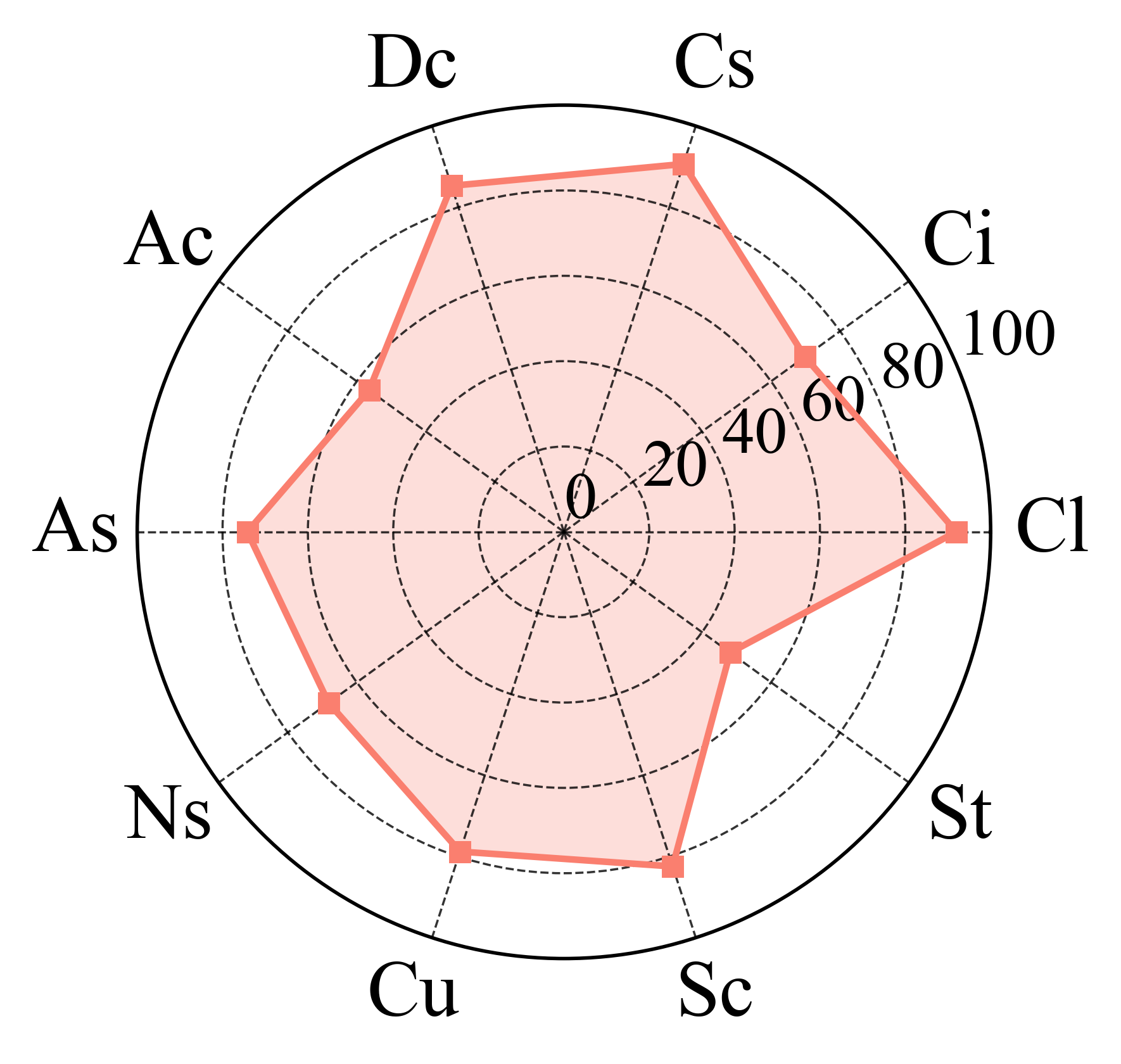}
    }
    \hspace{3.5mm}
    \subcaptionbox{$\mathrm{F_{1}}$-score\vspace{-2mm}\label{fig:CldNet_Stat_Radar:F1-score}}{%
        \includegraphics[width=0.25\textwidth,trim=0 0 0 0,clip]{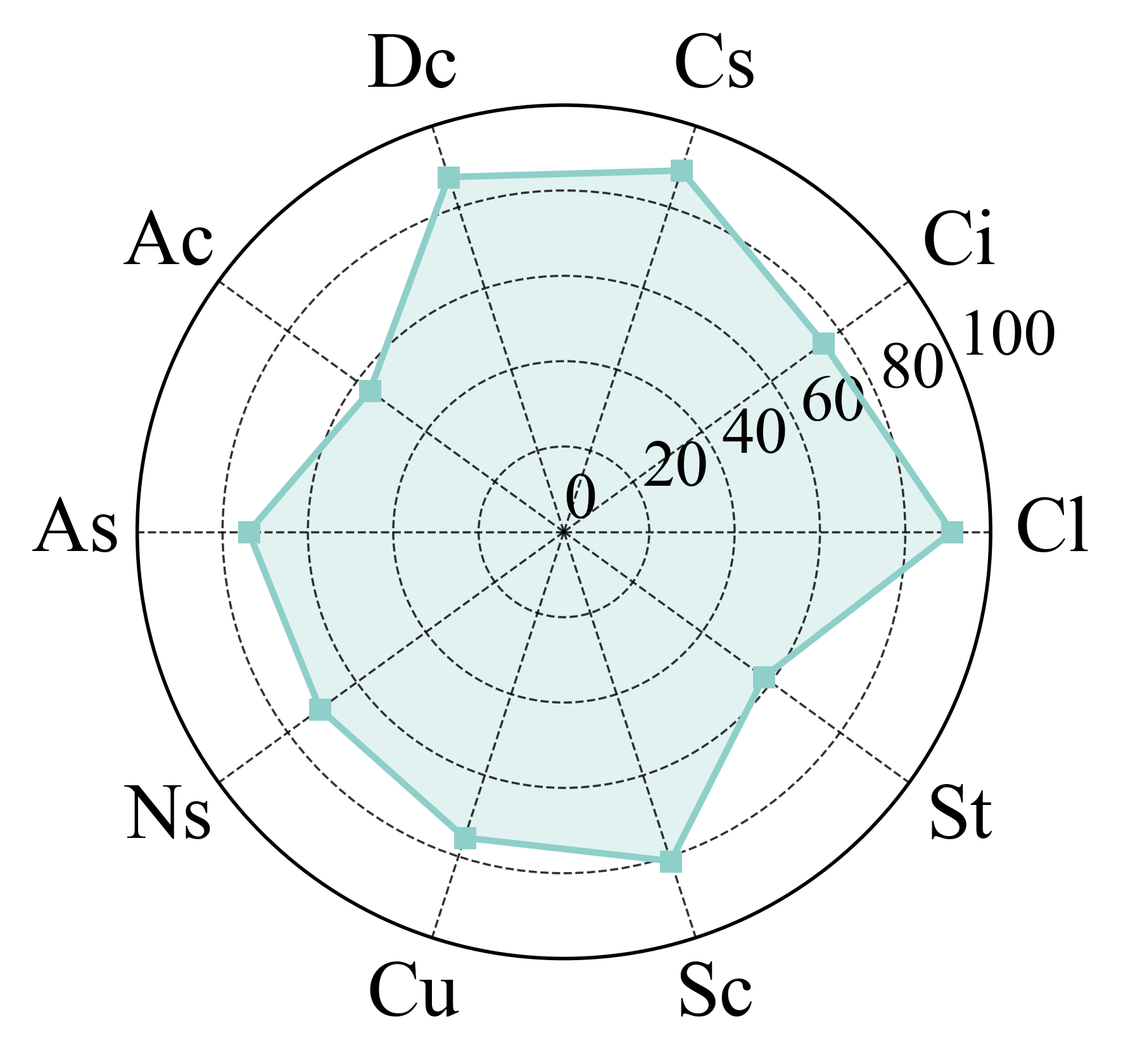}
    }
    \caption{The comparison of each cloud type predicted by CldNet at 2022-09-23 03:00 (UTC+0) in term of classification indicators, including (a) precision, (b) recall, and (c) $\mathrm{F_{1}}$-score.}
    \label{fig:CldNet_Stat_Radar}
\end{figure}

For a better assessment, detailed statistics of each cloud type prediction at 2022-09-23 03:00 (UTC+0) are provided.
Classification indicators of all models, including precision, recall, and $\mathrm{F_{1}}$-score are computed and recorded in Table~\ref{table:model_statA}.
The results show that CldNet performs best, and PSPNet performs worst.
The classification indicators obtained by CldNet for each cloud type are presented in Fig.~\ref{fig:CldNet_Stat_Radar}.
Meanwhile, the number of pixels for each cloud type obtained by CldNet at 2022-09-23 03:00 (UTC+0) is summarized in Table~\ref{table:CldNet_stat}.
From Table~\ref{table:CldNet_stat}, cloud types Ci and Cu are mistakenly identified as Ac due to their cloud optical thickness belonging to the same category and cloud top height of Ac being between Ci and Cu.
Therefore, the precision of Ac is relatively low in Fig.~\ref{fig:CldNet_Stat_Radar:Precision}.
In Fig.~\ref{fig:CldNet_Stat_Radar:Recall}, recall of St is lowest because a large number of cloud type St pixels are mistakenly identified as As, Ns, and Sc.

$\mathrm{F_{1}}$-score is calculated based on precision and recall, and the model performs poorly in cloud types Ac and St.
For multi-classification tasks, accuracy/$\mathrm{F_{1}\mbox{-}score_{micro}}$, $\mathrm{F_{1}\mbox{-}score_{macro}}$, $\mathrm{F_{1}\mbox{-}score_{weight}}$, and accuracy (N/Y for cloud) of cloud types at 2022-09-23 03:00 (UTC+0) are computed, which are recorded in Table~\ref{table:model_statB}.
The purpose of accuracy (N/Y for cloud) is to measure the ability of the model to distinguish between clear and cloudy skies.
The accuracies (N/Y for cloud) of CldNet, SegNet, PSPNet, DeepLabV3+, UNet, ResUnet, and UNetS are 95.27\%, 89.70\%, 88.33\%, 91.23\%, 95.00\%, 87.32\%, and 94.58\%, respectively.
From all of the metrics, our proposed model CldNet still maintains the performance of SOTA among all of the models.

\subsection{The impact of auxiliary information and satellite spectral channels} 
In the previous section, our proposed model CldNet is found to perform best in all models, and thus CldNet is chosen for a more in-depth study.
In order to improve the accuracy of the model, some auxiliary information has been added.
The auxiliary information includes satellite zenith angle (SAZ), satellite azimuth angle (SAA), solar zenith angle (SOZ), solar azimuth angle (SOA), longitude, latitude, altitude, and underlying surface attributes (land or water).
The impact of the auxiliary information on CldNet is explored in Table~\ref{table:CldNet_auxiliary_acc}, and it is found that CldNet-W, which adds SAZ, SAA, SOZ and SOA to the model inputs, performs best.
Compared with CldNet, the accuracy of CldNet-W has been improved by approximately 1.35\%.
This indicates that the auxiliary information can enhance the predictive ability of the model to identify cloud types.

\begin{table}[!htp]
    \centering
    \caption{The impact of auxiliary information on CldNet.}
    \label{table:CldNet_auxiliary_acc}
    \resizebox{0.96\linewidth}{!}{

        \begin{tabular}{cccccccccccc}
            \hline
            \multicolumn{8}{c}{Addition of auxiliary information} &            & \multicolumn{3}{c}{Accuracy (\%)}                                                                                                                                               \\ \cline{1-8} \cline{10-12}
            SAZ                                                       & SAA        & SOZ                               & SOA        & Longitude  & Latitude   & Altitude   & Underlying surface &  & Training            & Validation          & Test                \\ \hline
                                                                      &            &                                   &            &            &            &            &                    &  & 81.76±1.64          & 81.58±1.63          & 80.89±2.18          \\
            \checkmark                                                & \checkmark & \checkmark                        & \checkmark &            &            &            &                    &  & \textbf{82.86±1.71} & \textbf{82.96±1.60} & \textbf{82.23±2.14} \\
            \checkmark                                                & \checkmark &                                   &            &            &            &            &                    &  & 82.10±1.69          & 81.97±1.61          & 81.01±2.70          \\
                                                                      &            & \checkmark                        & \checkmark &            &            &            &                    &  & 82.42±1.65          & 82.35±1.58          & 81.68±2.02          \\
                                                                      &            &                                   &            & \checkmark & \checkmark & \checkmark & \checkmark         &  & 82.31±1.64          & 82.20±1.57          & 81.49±2.13          \\
                                                                      &            &                                   &            & \checkmark & \checkmark & \checkmark &                    &  & 82.10±1.62          & 81.88±1.53          & 81.22±1.99          \\
                                                                      &            &                                   &            &            &            &            & \checkmark         &  & 81.73±1.61          & 81.63±1.58          & 80.96±2.09          \\
            \checkmark                                                & \checkmark & \checkmark                        & \checkmark & \checkmark & \checkmark & \checkmark & \checkmark         &  & 82.56±1.77          & 82.70±1.75          & 82.01±2.28          \\ \hline
        \end{tabular}
    }
\end{table}


An important aim of this study is to extrapolate cloud types in the nighttime areas by training model parameters using daytime label data.
VIS and NIR channels can be used during the day, but not at night.
If the model does not use VIS and NIR information, mask module will set the data involving VIS and NIR channels to zero.
CldNet-O uses mask module to set the model input data involving VIS and NIR channels to zero through mask ratio and mask bands based on CldNet-W.

\begin{figure}[!htp]
    \centering
    \subcaptionbox{CldNet-W\vspace{-2mm}\label{fig:models_stage_b:CldNet-W1}}{%
        \includegraphics[width=0.32\textwidth,trim=0 0 30 0,clip]{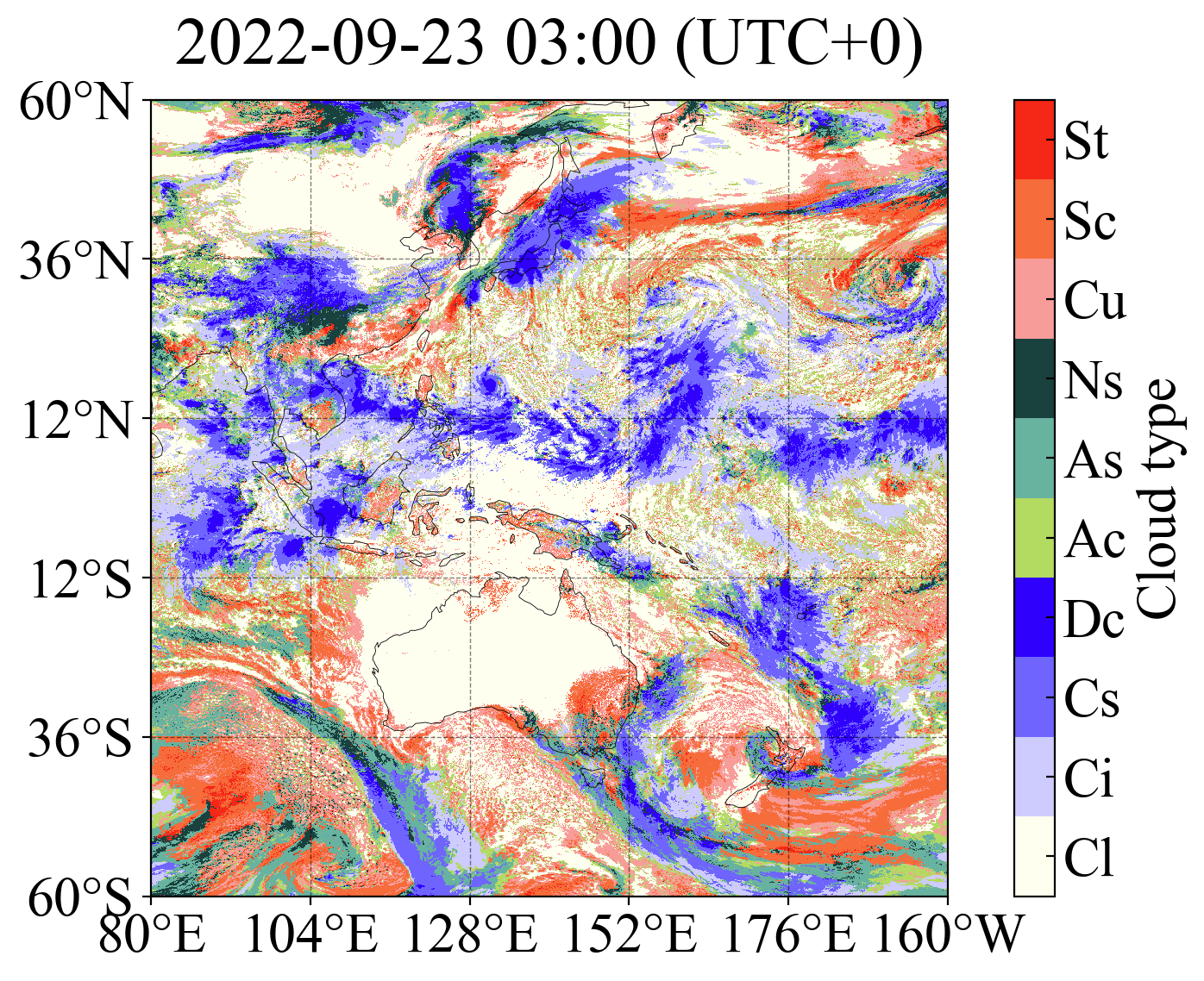}
    }
    \hfill
    \subcaptionbox{CldNet-O\vspace{-2mm}\label{fig:models_stage_b:CldNet-O1}}{%
        \includegraphics[width=0.32\textwidth,trim=0 0 30 0,clip]{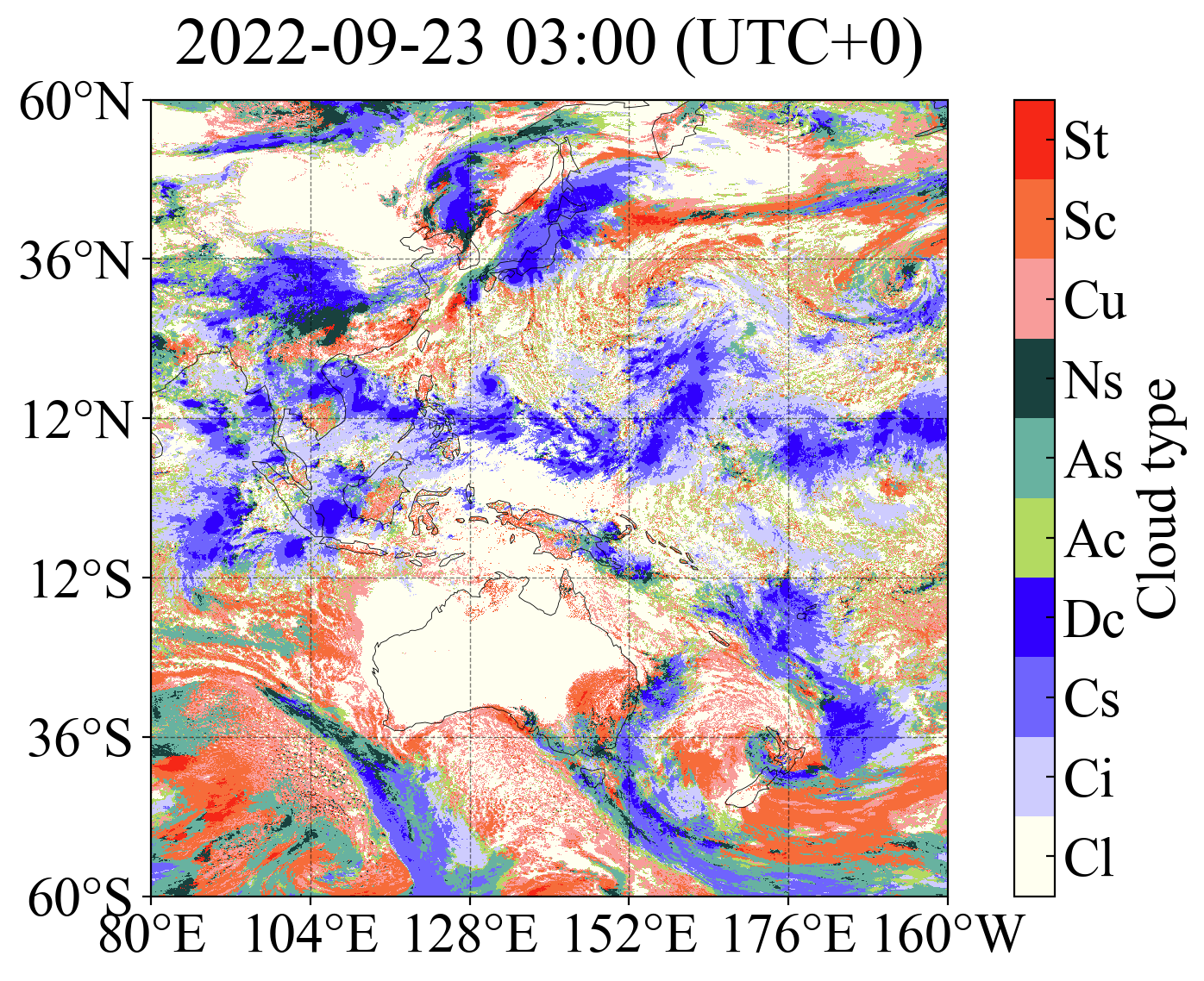}
    }
    \subcaptionbox{CldNet-W $\mathrm{-}$ CldNet-O\vspace{-2mm}\label{fig:models_stage_b:cldDD}}{%
        \begin{overpic}[abs,unit=1mm,width=0.32\textwidth,trim=0 0 0 -25,clip]{
                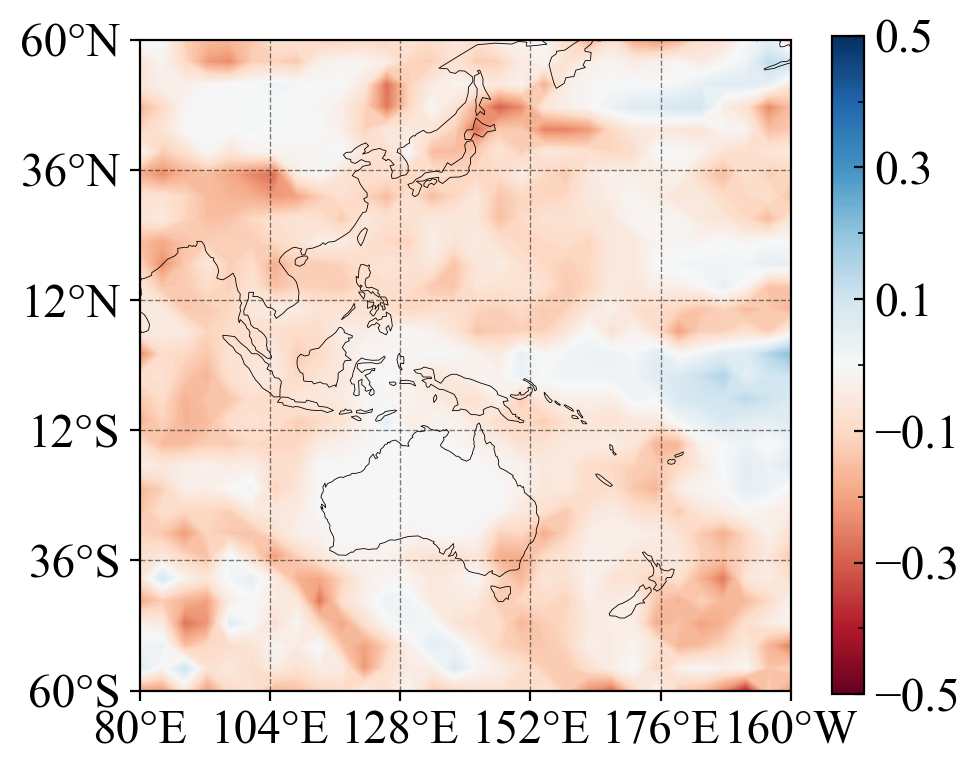
            }
        \end{overpic}
    }
    \caption{The overall cloud-type distributions at 2022-09-23 03:00 (UTC+0) are predicted by the trained (a) CldNet-W and (b) CldNet-O. And (c) the differences between CldNet-W and CldNet-O in terms of the cloud-type prediction error density.}
    \label{fig:models_stage_b}
\end{figure}

The overall cloud-type distributions predicted by the trained CldNet-W and CldNet-O at 2022-09-23 03:00 (UTC+0) are shown in Fig.~\ref{fig:models_stage_b}, whose specific indicator results are recorded in Table~\ref{table:model_statA} and Table~\ref{table:model_statB}.
In Fig.~\ref{fig:models_stage_b:cldDD}, CldNet-W performs better than CldNet-O in cloud-type identification for most regions.
However, CldNet-O extrapolates cloud types in the nighttime region due to not involving VIS and NIR channels.
Compared with CldNet-W, the accuracy/$\mathrm{F_{1}\mbox{-}score_{micro}}$ and accuracy (N/Y for cloud) of CldNet-O have decreased by 7.34\% and 0.45\%, respectively.
It can be seen that the removal of VIS and NIR has little effect on the distinction between clear and cloudy skies, but a significant impact on the identification of cloud types, and this result is also confirmed by the study of \citet{Zhang20196464}.
In order to construct a high-fidelity, all-day, spatiotemporal cloud-type database with spatial resolution $0.05^{\circ}\times0.05^{\circ}$ over the entire satellite observation area, it is vitally important that CldNet-W and CldNet-O can be deployed online for daytime and nighttime areas, respectively.

\begin{figure}[!htp]
    \centering
    \vspace{0mm}
    \subcaptionbox{CldNet-O\vspace{-2mm}\label{fig:Cld_20220922_CldNet-O1}}{
        \begin{minipage}[b]{\textwidth}
            \centering
            \includegraphics[width=0.32\textwidth,trim=0 0 30 -10,clip]{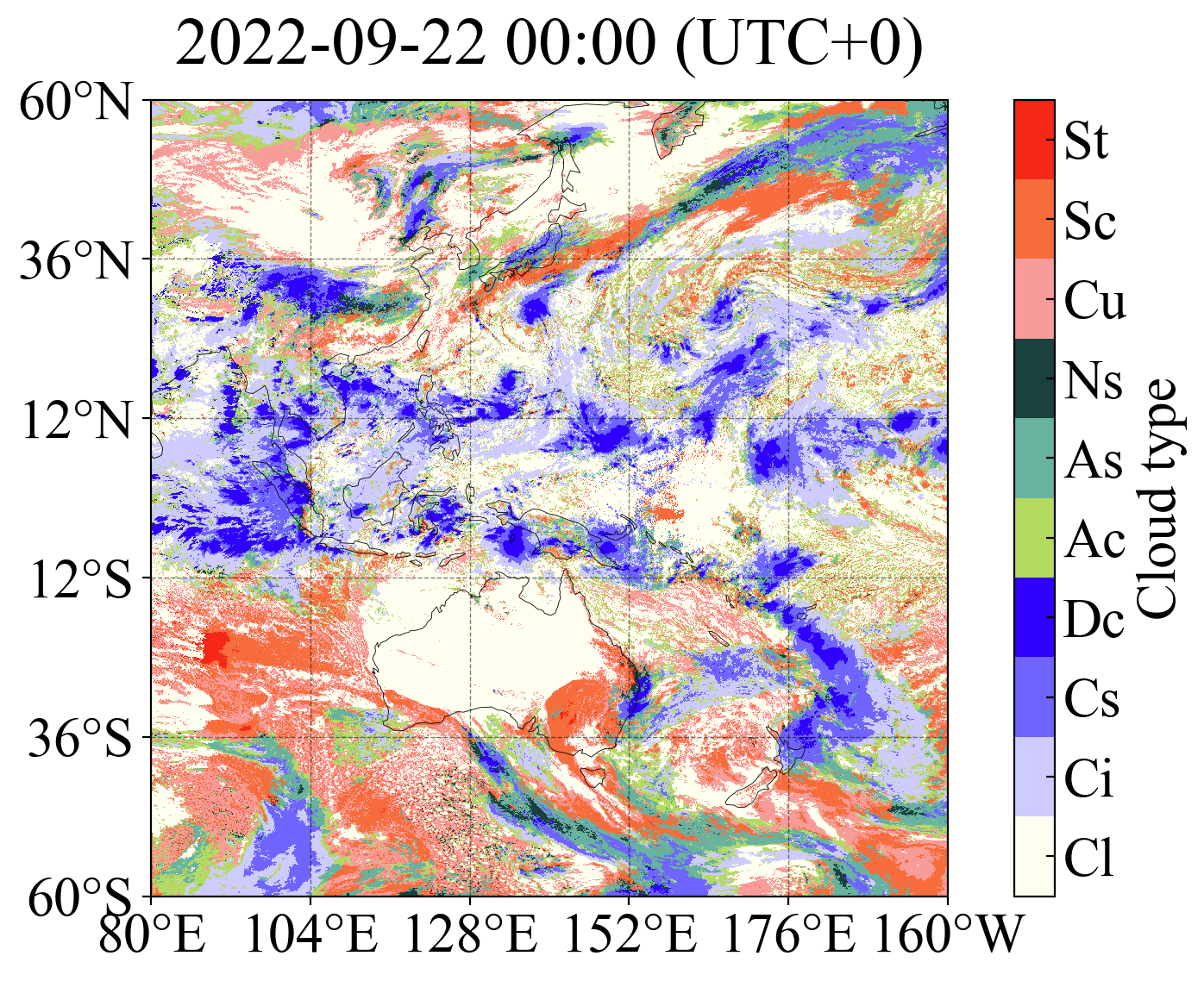}
            \hfill
            \includegraphics[width=0.32\textwidth,trim=0 0 30 -10,clip]{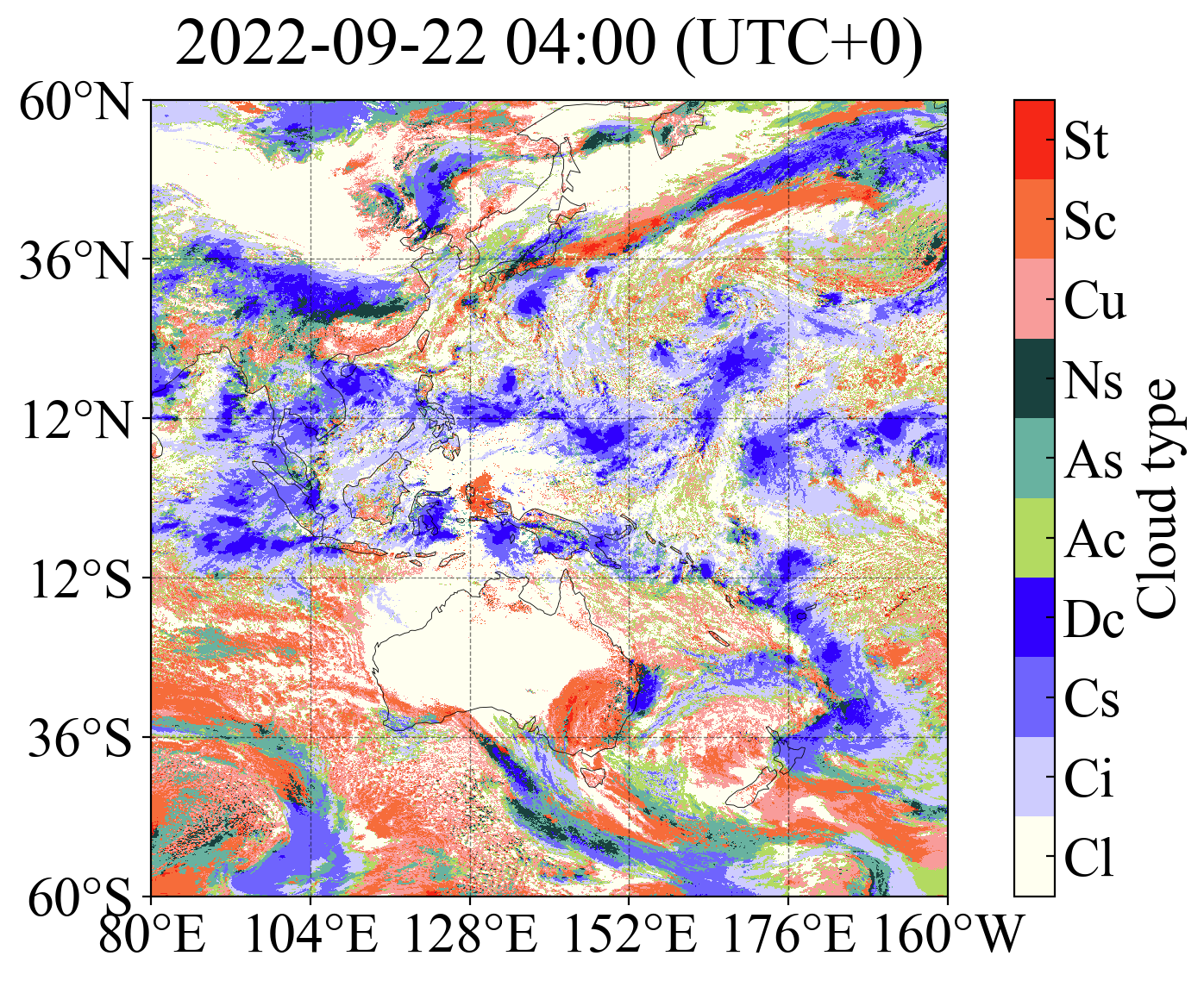}
            \hfill
            \includegraphics[width=0.32\textwidth,trim=0 0 30 -10,clip]{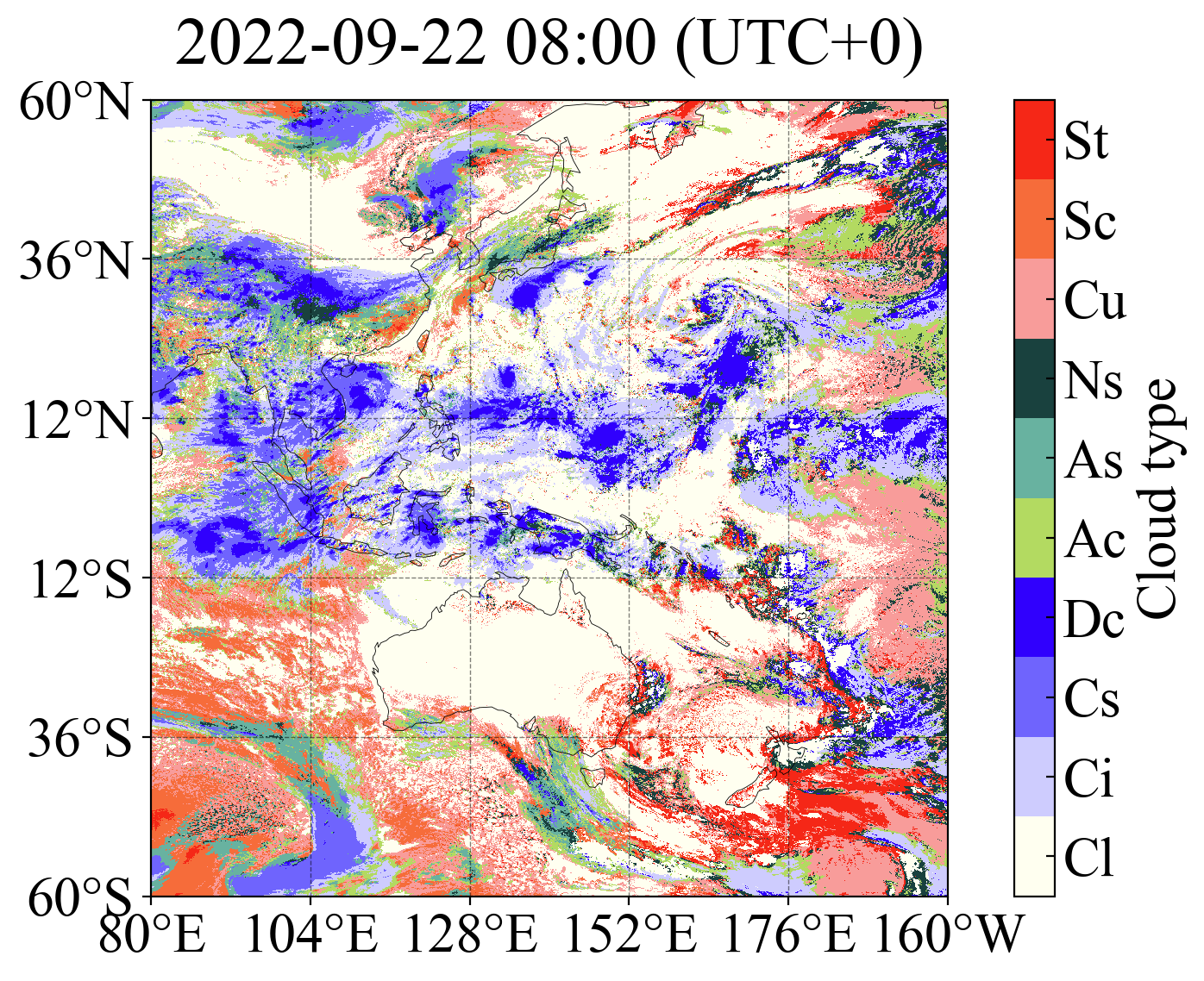}
            \\
            \includegraphics[width=0.32\textwidth,trim=0 0 30 -10,clip]{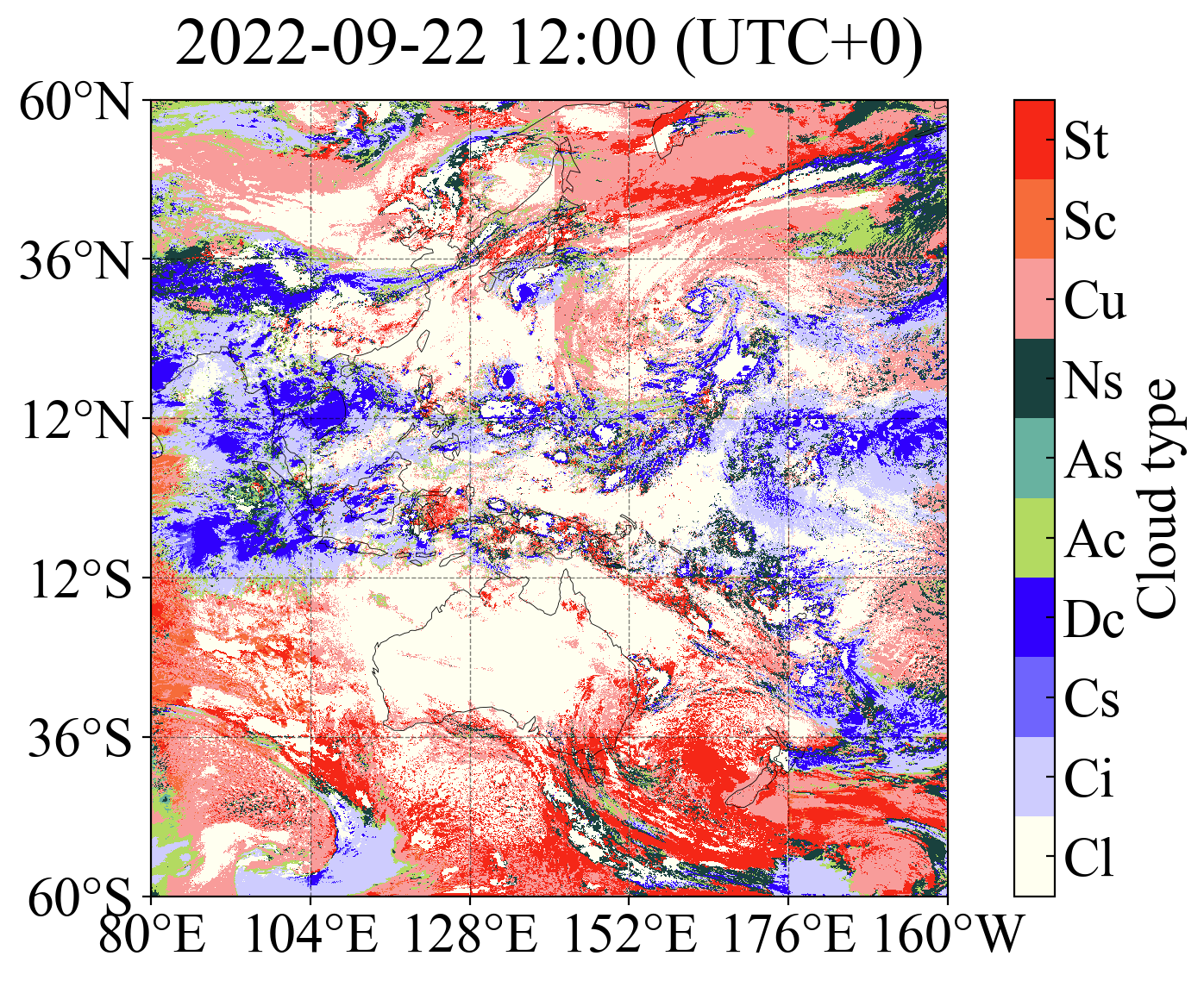}
            \hfill
            \begin{overpic}[abs,unit=1mm,width=0.32\textwidth,trim=0 0 30 -10,clip]{
                    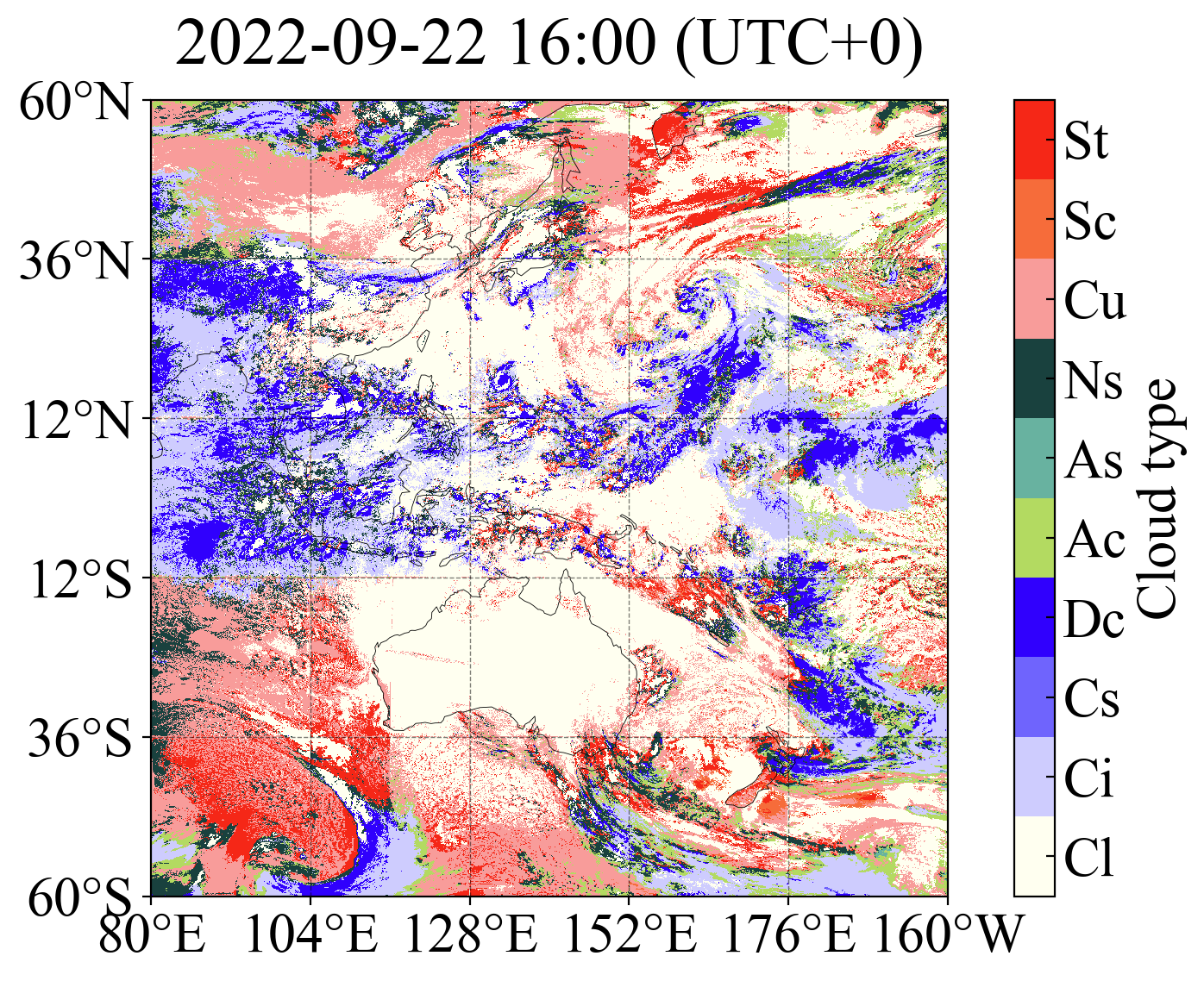
                }
                \put(18.5,15.5){\color{black}Fig.\ref{fig:models_stage_bb:cc}}
                \put(0,0){
                    \color{black}
                    \linethickness{0.35mm}
                    \polygon(17,13)(17,19)(29,19)(29,13)
                }
            \end{overpic}
            \hfill
            \includegraphics[width=0.32\textwidth,trim=0 0 30 -10,clip]{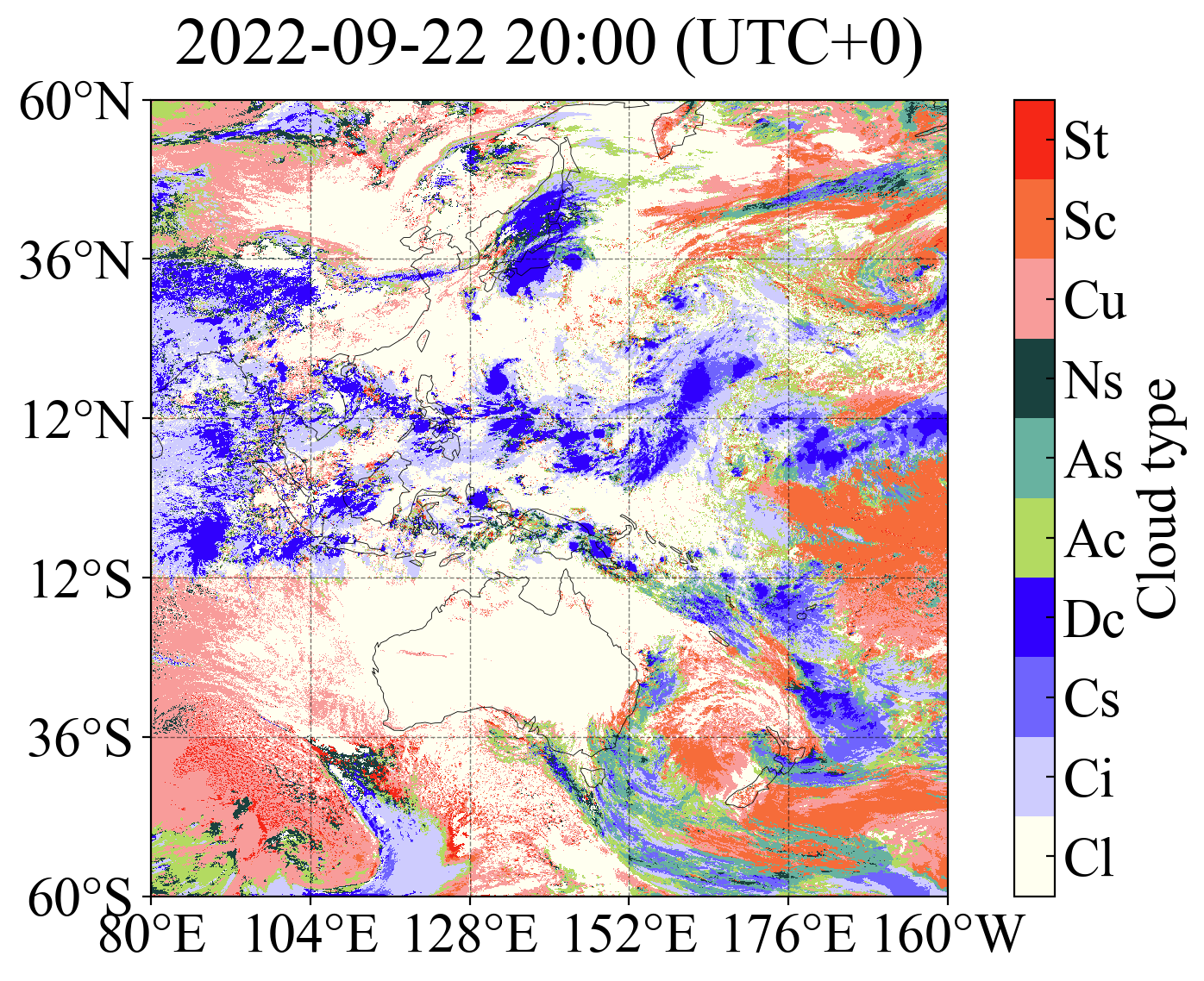}
        \end{minipage}
    }
    \newline
    \vspace{2mm}
    \subcaptionbox{Reference\vspace{-2mm}\label{fig:Cld_20220922_reference}}{
        \begin{minipage}[b]{\textwidth}
            \centering
            \includegraphics[width=0.32\textwidth,trim=0 0 30 -10,clip]{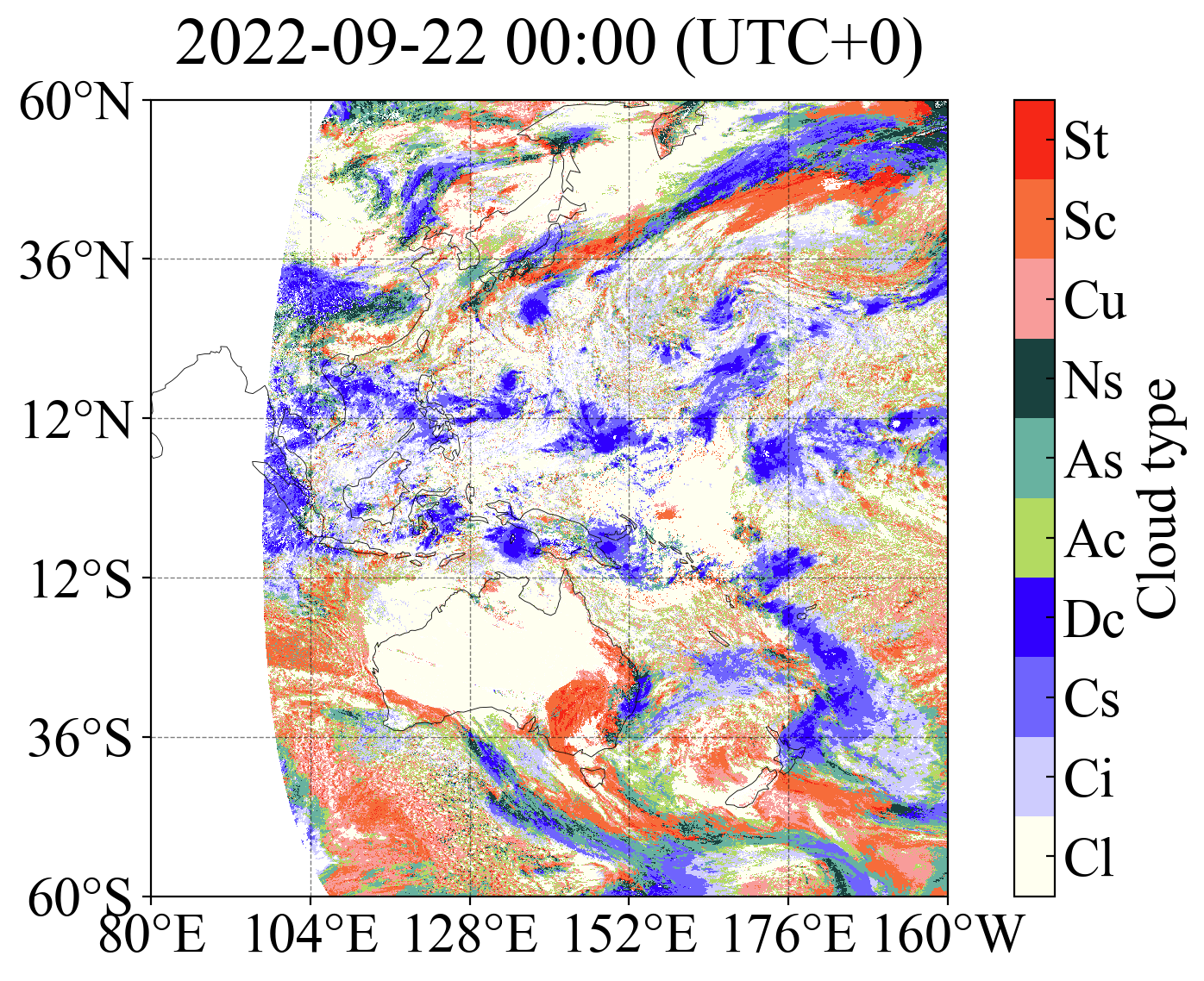}
            \hfill
            \includegraphics[width=0.32\textwidth,trim=0 0 30 -10,clip]{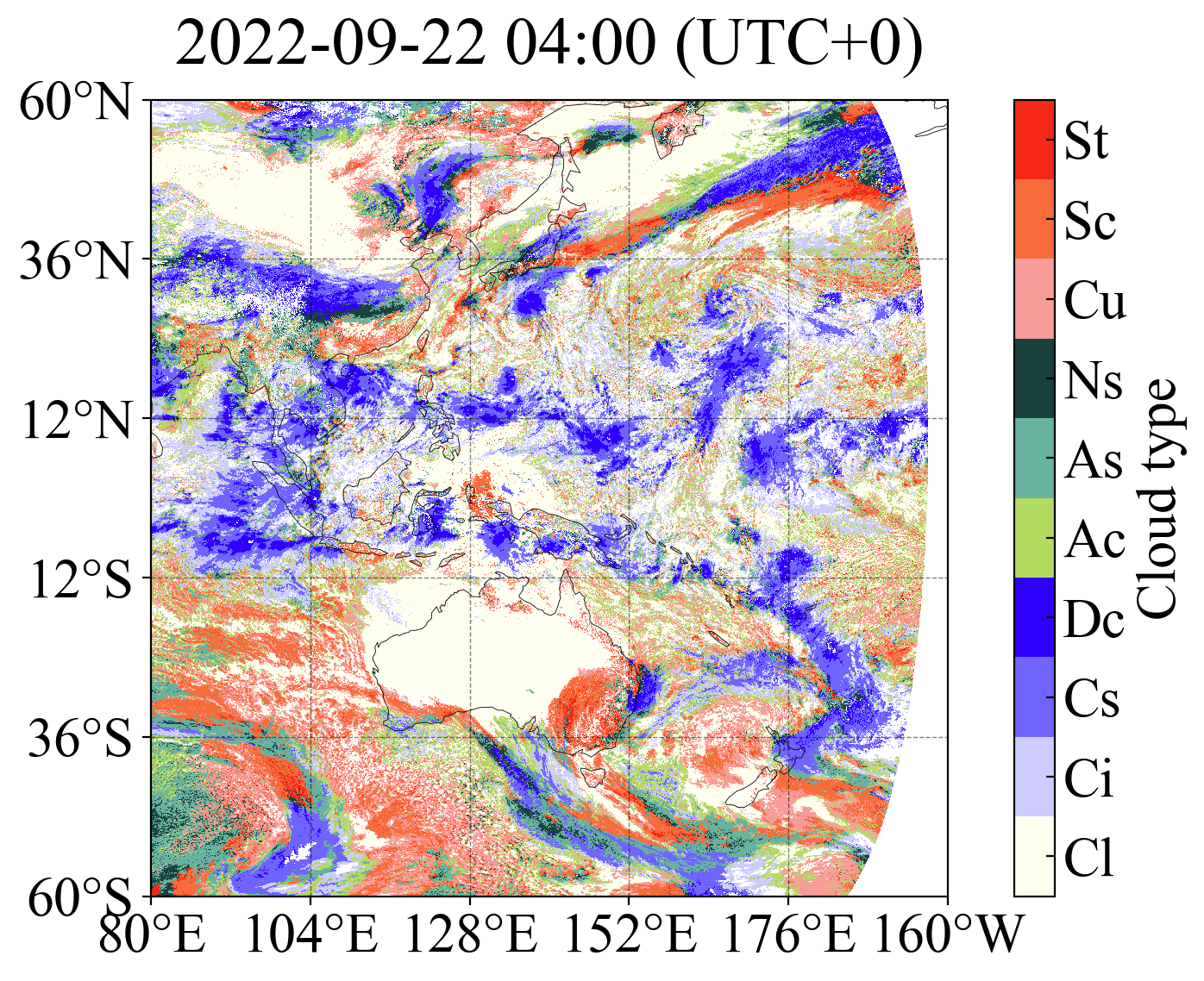}
            \hfill
            \includegraphics[width=0.32\textwidth,trim=0 0 30 -10,clip]{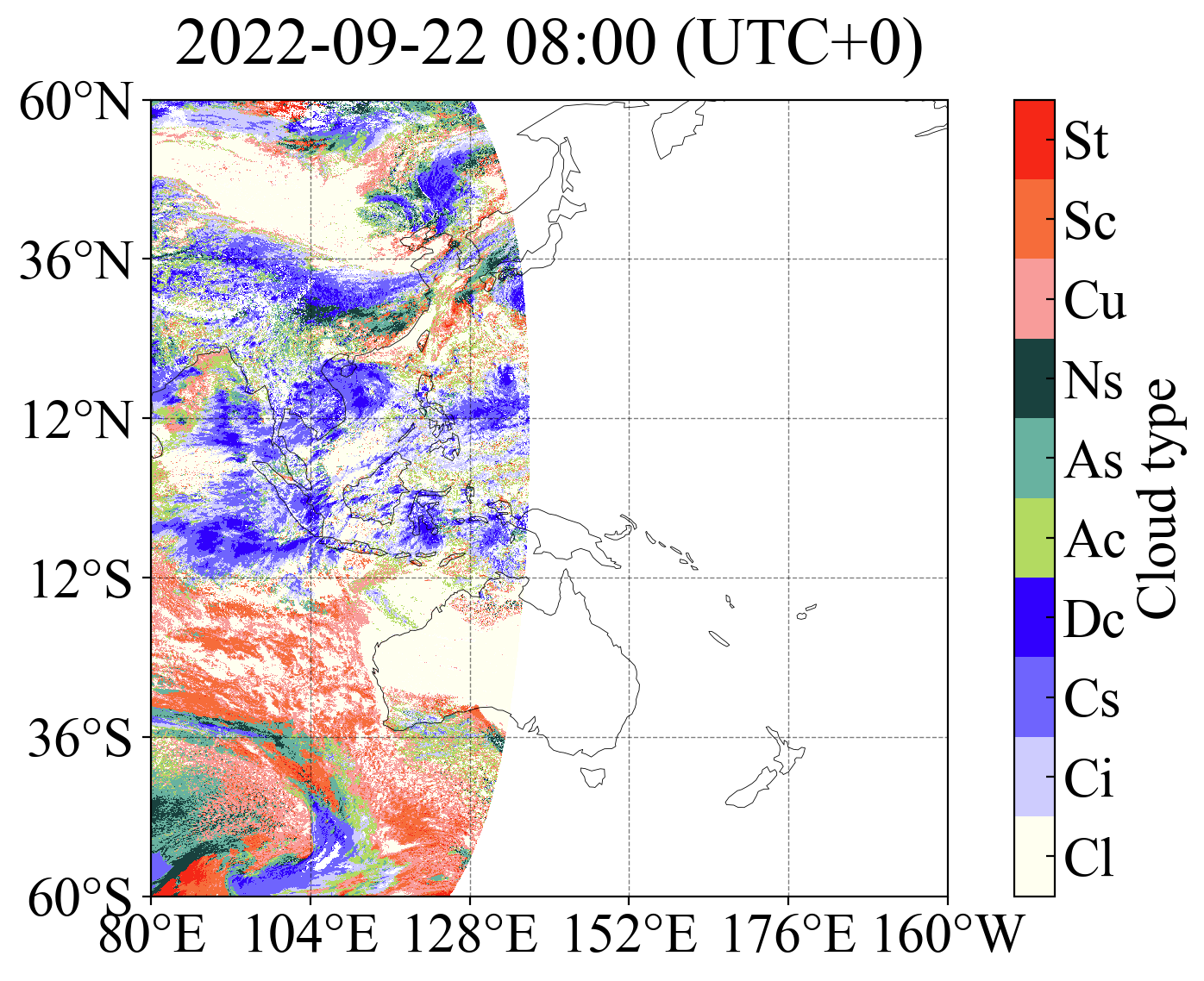}
            \\
            \includegraphics[width=0.32\textwidth,trim=0 0 30 -10,clip]{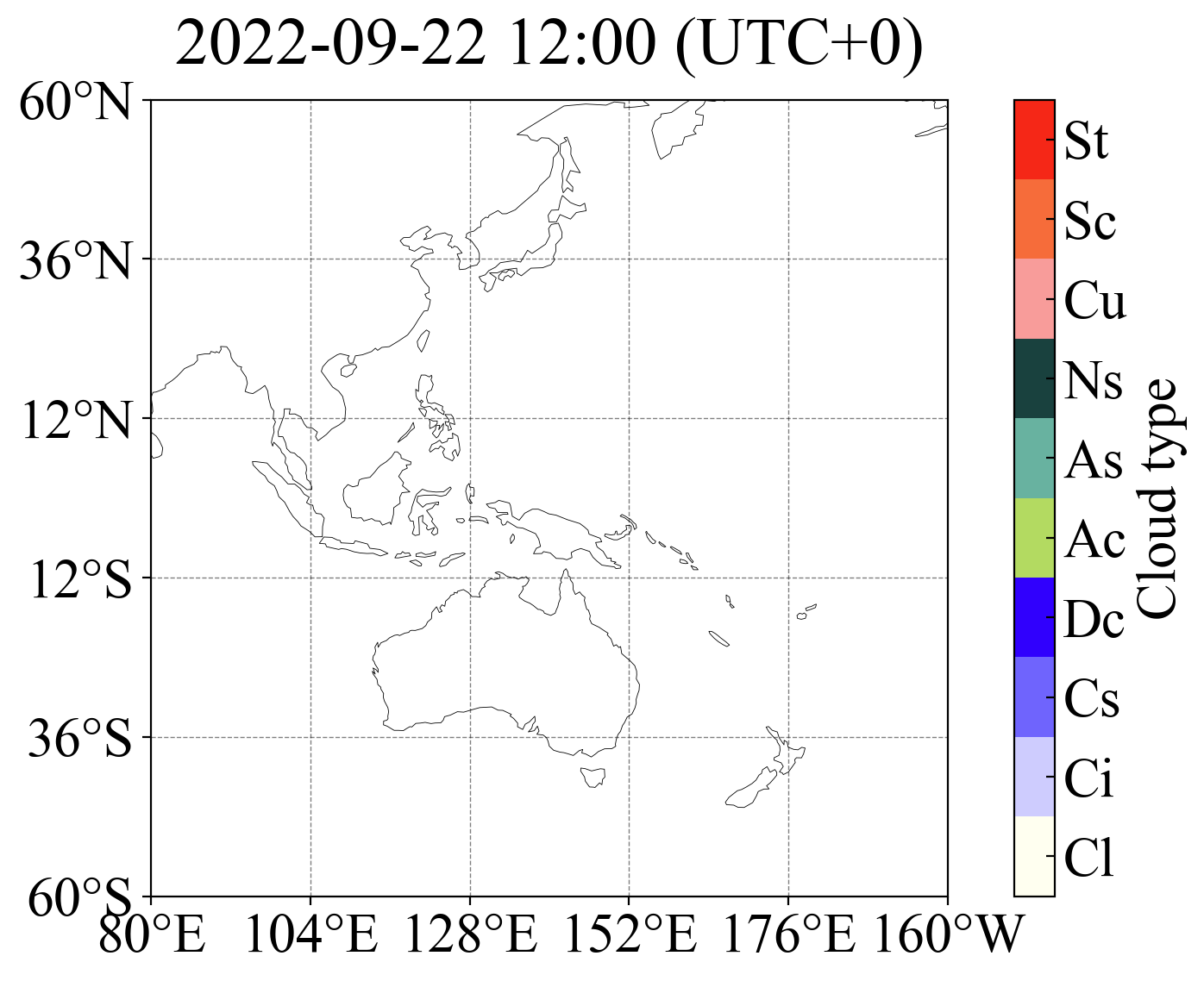}
            \hfill
            \begin{overpic}[abs,unit=1mm,width=0.32\textwidth,trim=30 0 0 -10,clip]{
                    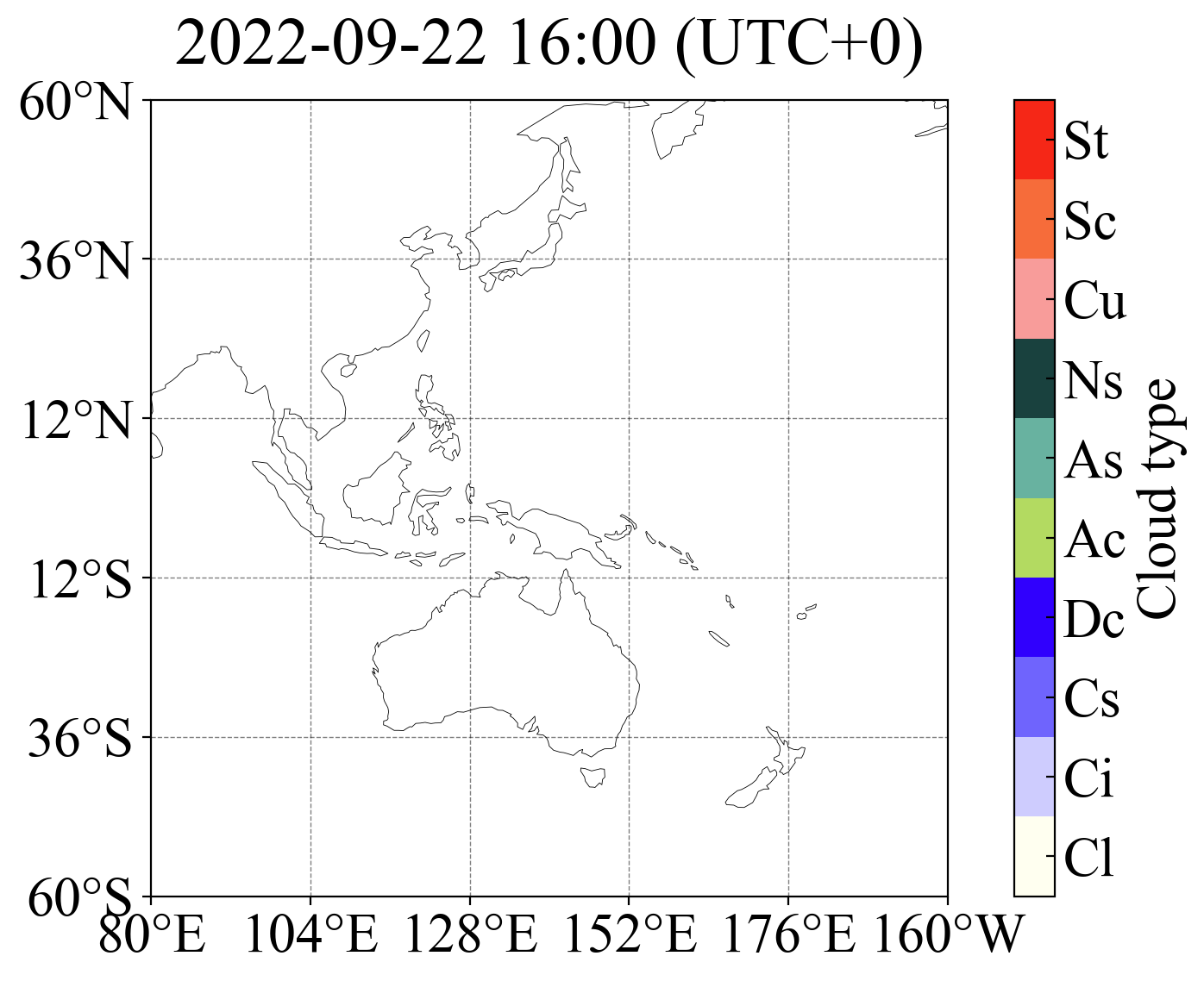
                }
            \end{overpic}
            \hfill
            \includegraphics[width=0.32\textwidth,trim=30 0 0 -10,clip]{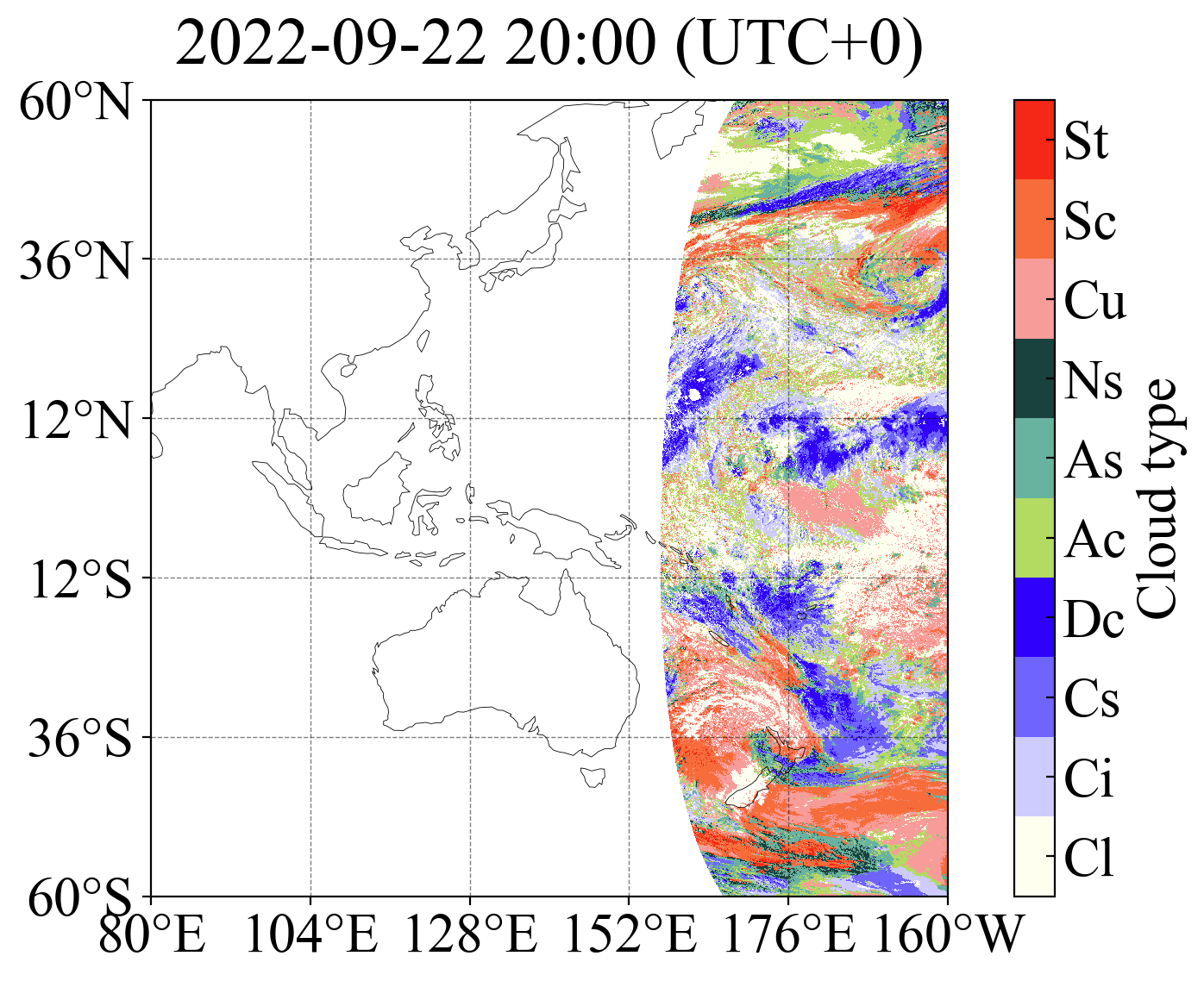}
        \end{minipage}
    }
    \caption{The overall distributions of cloud types using (a) CldNet-O and (b) reference every 4 h during 2022-09-22 are shown, and the black box is enlarged in Fig.~\ref{fig:models_stage_bb:cc}.}
    \label{fig:Cld_20220922_a}
\end{figure}

\begin{figure}[!htp]
    \vskip-0pt
    \centering
    \begin{minipage}[width=0.95\textwidth,trim=0 0 0 0,clip]{\textwidth}
        \centering
        \includegraphics[width=0.32\textwidth,trim=0 0 0 -10,clip]{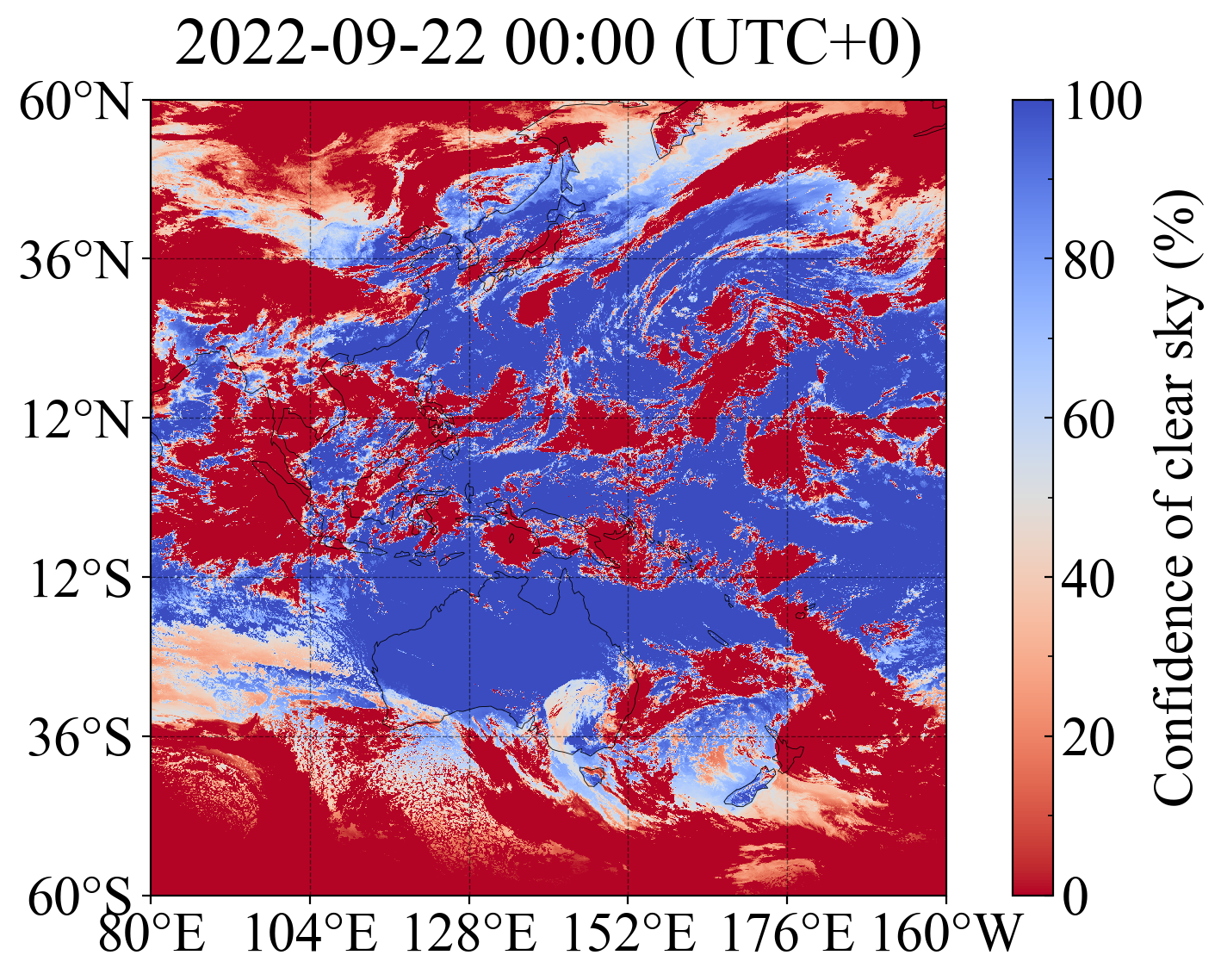}
        \hfill
        \includegraphics[width=0.32\textwidth,trim=0 0 0 -10,clip]{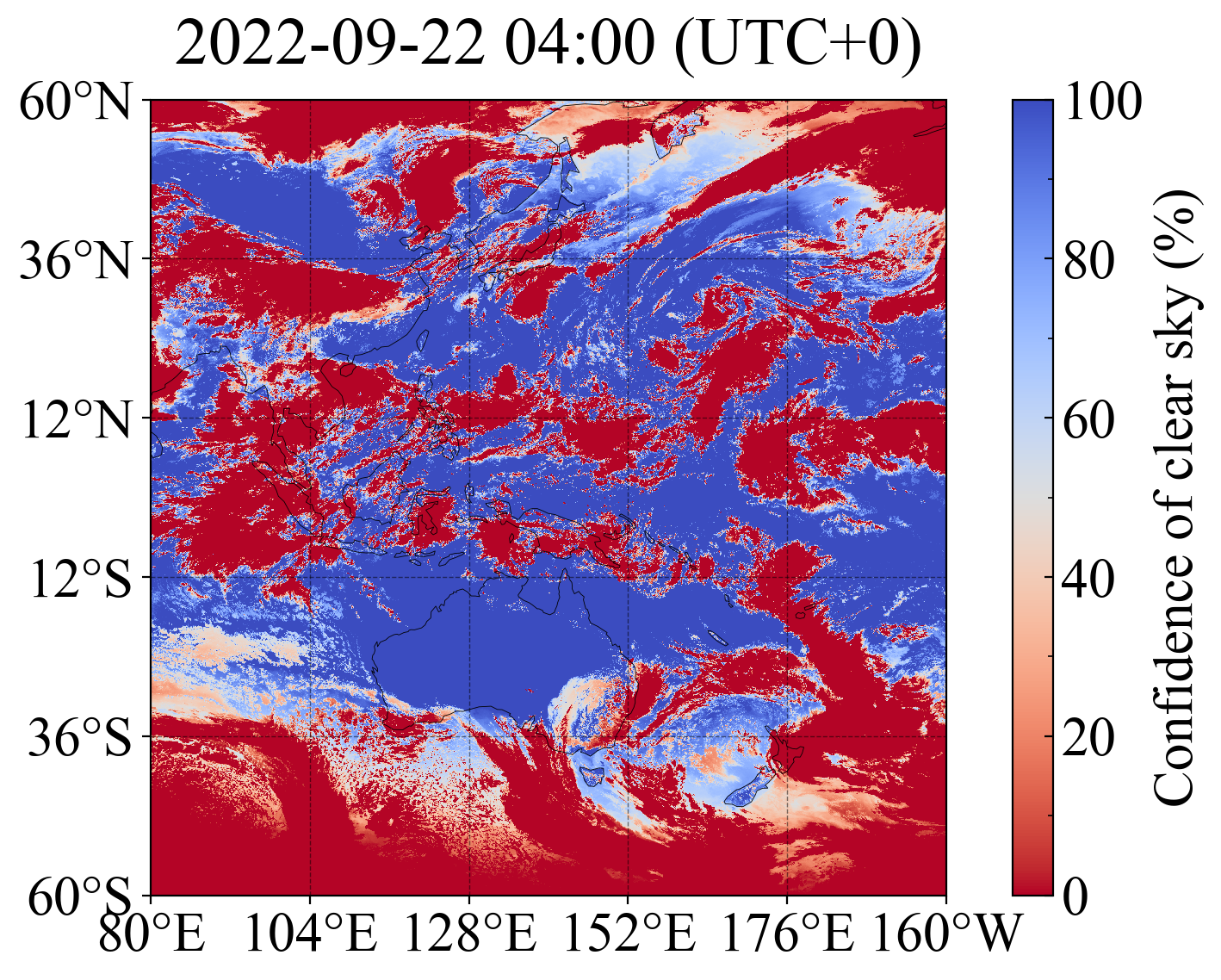}
        \hfill
        \includegraphics[width=0.32\textwidth,trim=0 0 0 -10,clip]{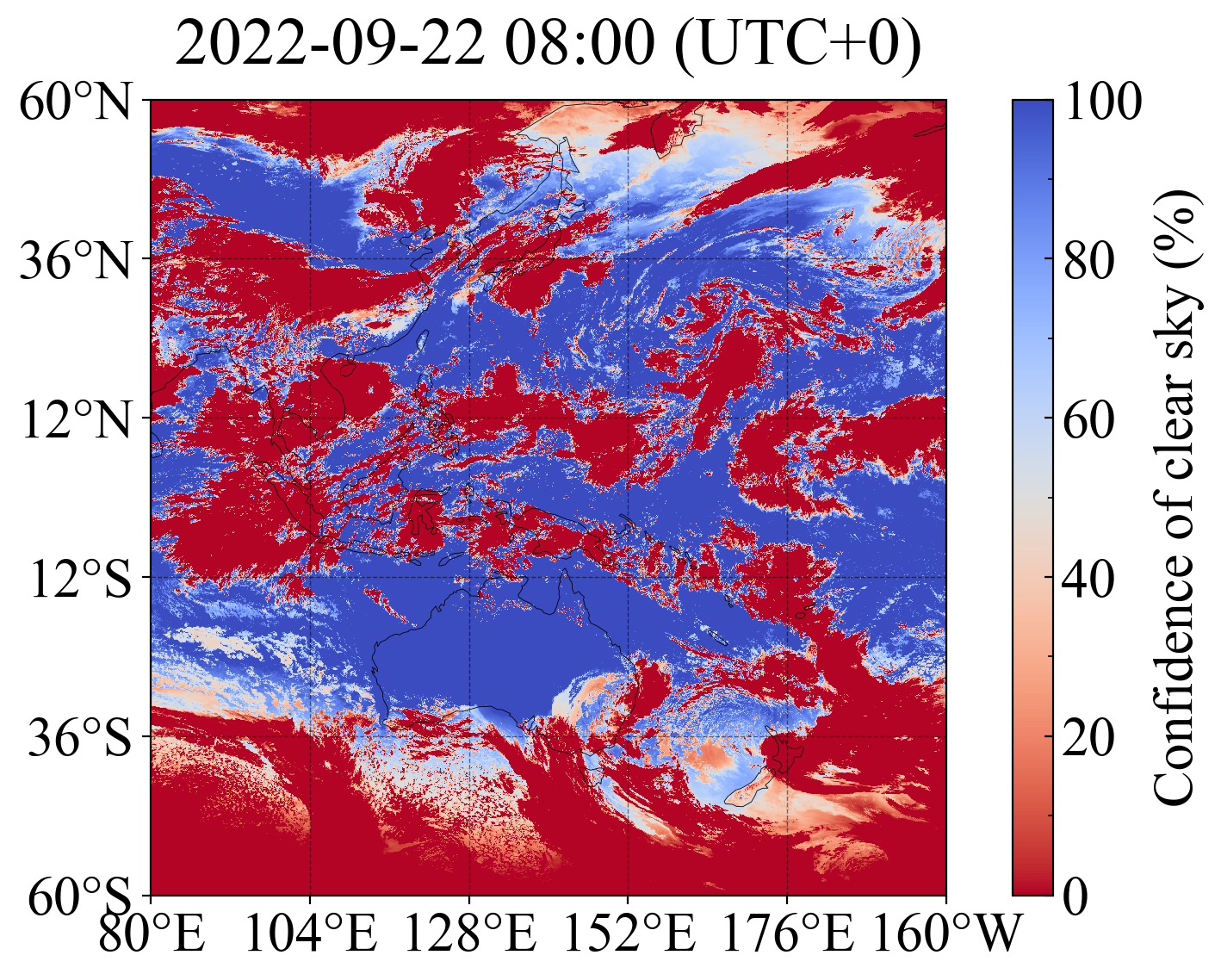}
        \\
        \includegraphics[width=0.32\textwidth,trim=0 0 0 -10,clip]{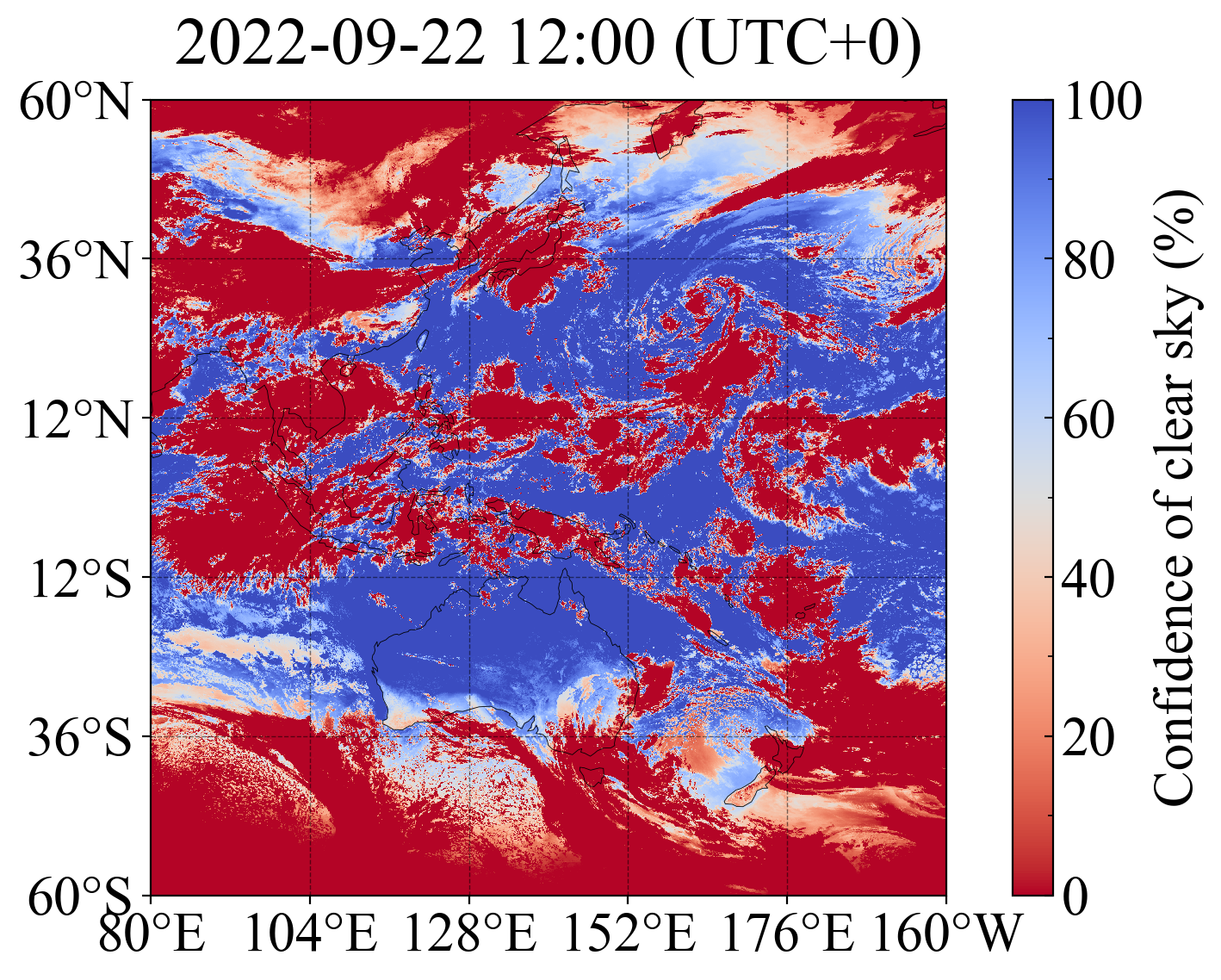}
        \hfill
        \begin{overpic}[abs,unit=1mm,width=0.32\textwidth,trim=0 0 0 -10,clip]{
                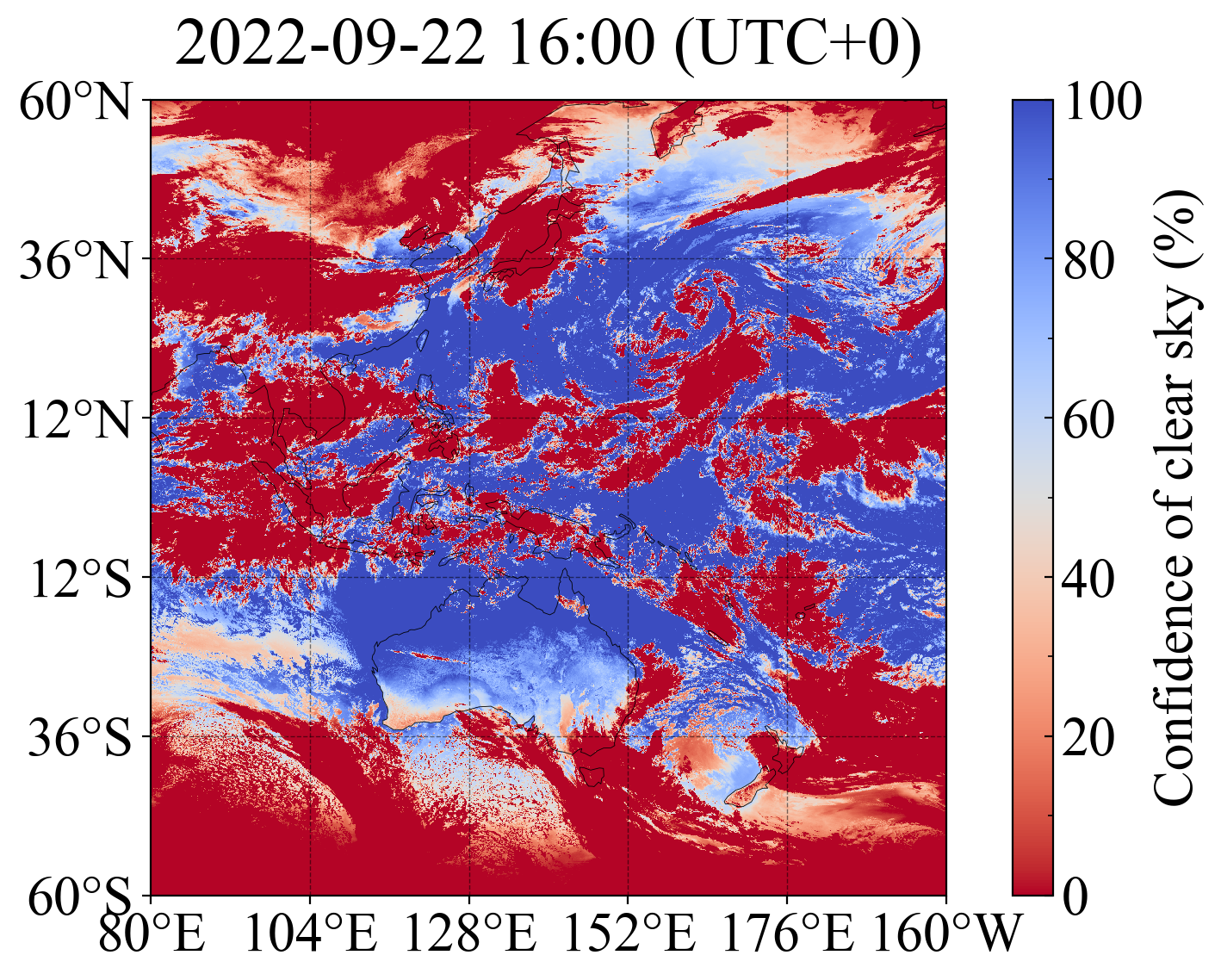
            }
            \put(16.5,14.5){\color{black}Fig.\ref{fig:models_stage_bb:ccs}}
            \put(0,0){
                \color{black}
                \linethickness{0.35mm}
                \polygon(15,12)(15,18.3)(27,18.3)(27,12)
            }
        \end{overpic}
        \hfill
        \includegraphics[width=0.32\textwidth,trim=0 0 0 -10,clip]{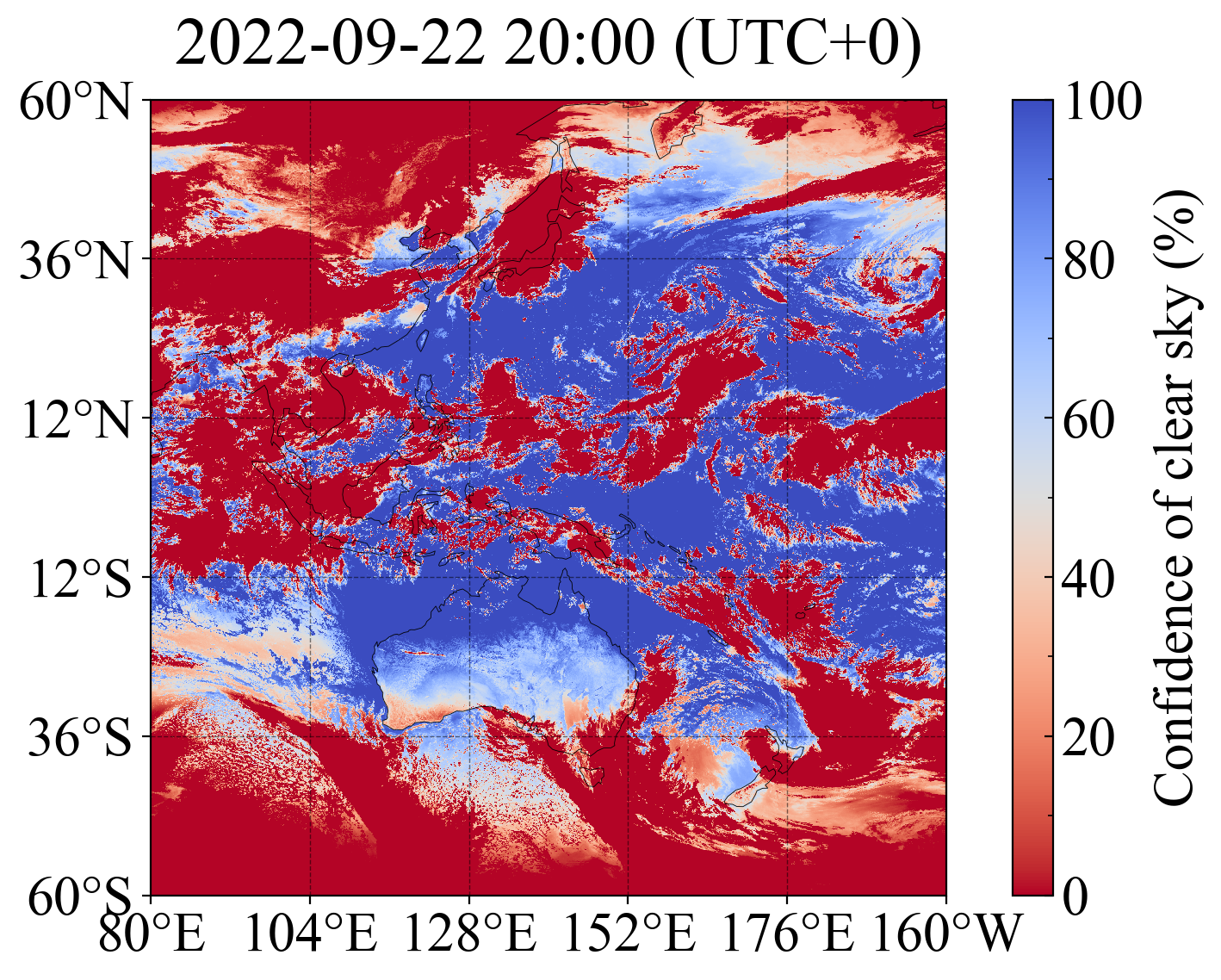}
    \end{minipage}
    \caption{The overall distributions of confidence of clear sky using the method proposed by \citet{Shang2016JD025659} every 4 h during 2022-09-22 are shown, and the black box is enlarged in Fig.~\ref{fig:models_stage_bb:ccs}.}
    \label{fig:Cld_20220922_clearsky_confidence}
\end{figure}

\subsection{All-day recognition of cloud types} 
The overall distributions of cloud types every 4 h for CldNet-O and reference during 2022-09-22 are shown in Fig.~\ref{fig:Cld_20220922_a}.
Compared to the reference in Fig.~\ref{fig:Cld_20220922_reference}, the cloud-type distributions in the nighttime area can be observed using CldNet-O in Fig.~\ref{fig:Cld_20220922_CldNet-O1}.
In order to evidence the predicted results in the nighttime area, a method~\citep{Shang2016JD025659} for the identification of clear and cloudy sky over land based on BT of the satellites Himawari-8/9 spectral channel B14 is adopted.
Here, confidence of clear sky for each pixel in the satellites Himawari-8/9 image is computed through the formula $\mathrm{(B14-270)/(288-270)\times100\%}$.
The overall distributions of confidence of clear sky every 4 h during 2022-09-22 are shown in Fig.~\ref{fig:Cld_20220922_clearsky_confidence}.

\begin{figure}[!htp]
    \centering
    \subcaptionbox{CldNet-O\vspace{0mm}\label{fig:models_stage_bb:cc}}{
        \begin{overpic}[abs,unit=1mm,width=0.45\textwidth,trim=0 0 0 0,clip]{
                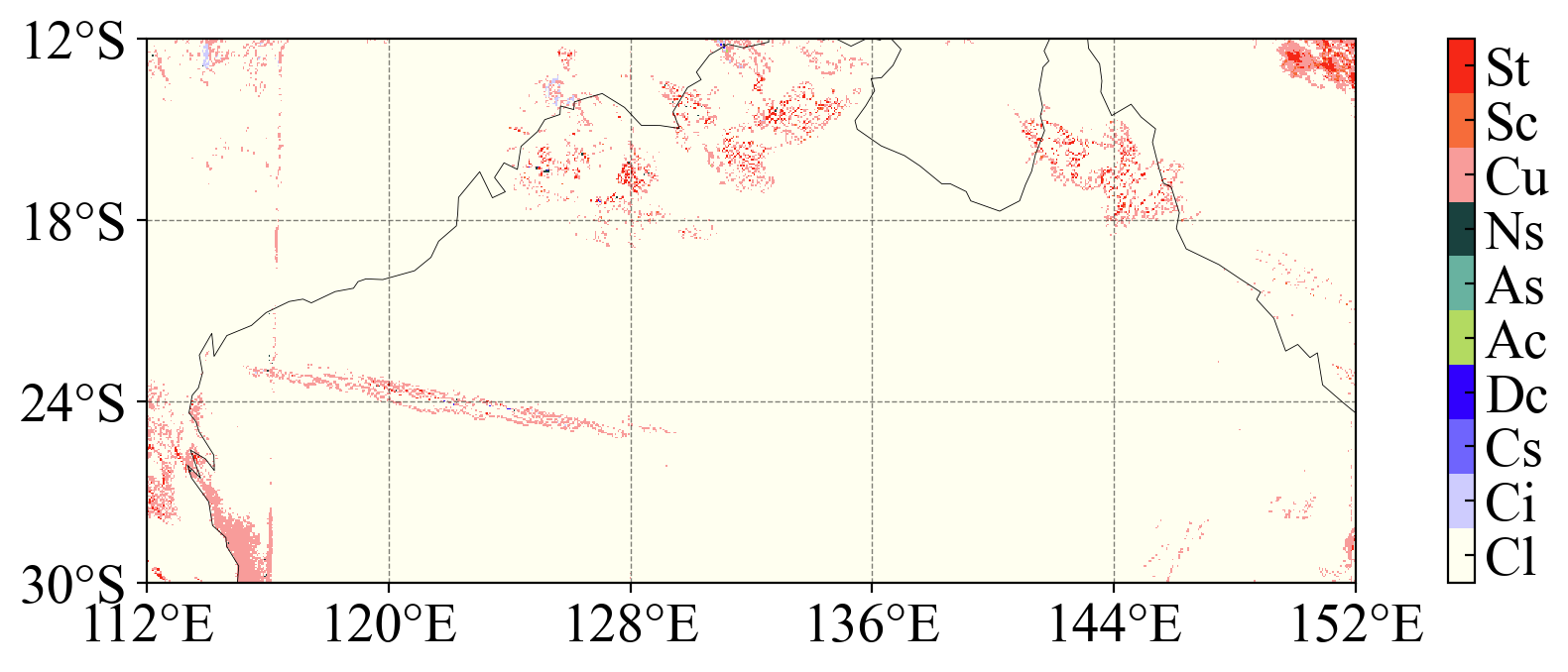
            }
            \put(32,19){\color{red}R01}
            \put(0,0){
                \color{red}
                \linethickness{0.35mm}
                \polygon(28,22.3)(28,28.3)(40.3,28.3)(40.3,22.3)
            }
            \put(17.3,16){\color{green}R02}
            \put(0,0){
                \color{green}
                \linethickness{0.35mm}
                \polygon(11,11)(11,15)(28.5,15)(28.5,11)
            }
            \put(49.5,26.8){\color{orange}R03}
            \put(0,0){
                \color{orange}
                \linethickness{0.35mm}
                \polygon(47.7,21)(47.7,26.3)(55.7,26.3)(55.7,21)
            }
        \end{overpic}
    }
    \subcaptionbox{The method proposed by \citet{Shang2016JD025659}\vspace{0mm}\label{fig:models_stage_bb:ccs}}{
        \begin{overpic}[abs,unit=1mm,width=0.483\textwidth,trim=0 0 0 -0.43,clip]{
                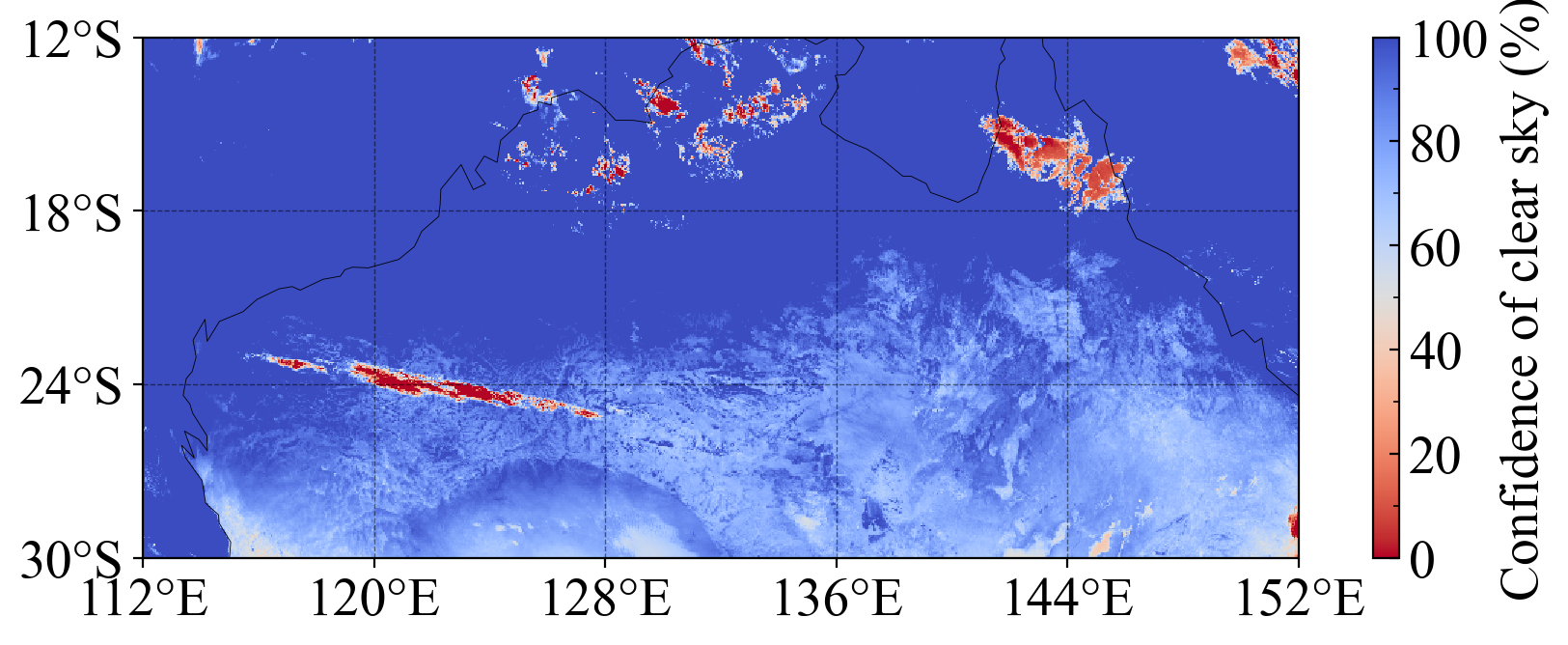
            }
            \put(33,19.5){\color{red}R01}
            \put(0,0){
                \color{red}
                \linethickness{0.35mm}
                \polygon(28.5,23)(28.5,29)(41.5,29)(41.5,23)
            }
            \put(18,16.5){\color{green}R02}
            \put(0,0){
                \color{green}
                \linethickness{0.35mm}
                \polygon(11.5,11.5)(11.5,15.5)(29.5,15.5)(29.5,11.5)
            }
            \put(51,27.5){\color{orange}R03}
            \put(0,0){
                \color{orange}
                \linethickness{0.35mm}
                \polygon(49.5,21.5)(49.5,27)(57.5,27)(57.5,21.5)
            }
        \end{overpic}
    }
    \caption{The overall cloud-type distributions predicted by the trained (a) CldNet-O and the overall distributions of confidence of clear sky using (b) the method proposed by \citet{Shang2016JD025659} at 2022-09-22 16:00 (UTC+0).}
    \label{fig:models_stage_bb}
\end{figure}
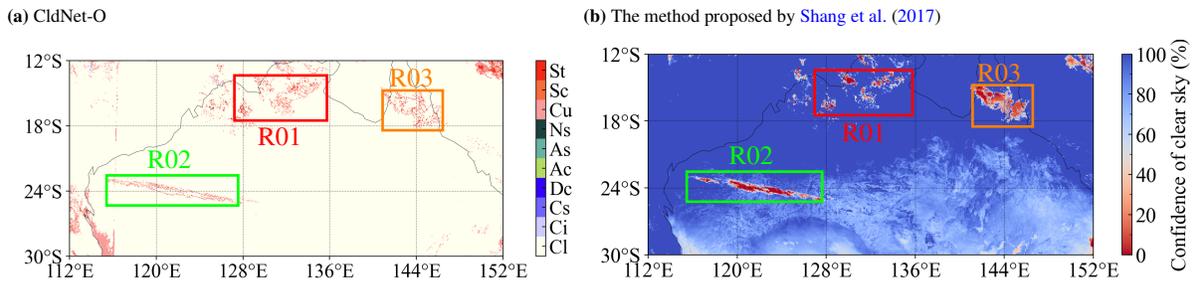

The cloud cover over Australia (the black box in Fig.~\ref{fig:Cld_20220922_a}) is enlarged and compared with the confidence of clear sky at 2022-09-22 16:00 (UTC+0) in Fig.~\ref{fig:models_stage_bb}.
The confidence of clear sky and cloud cover in the region R01 in Fig.~\ref{fig:models_stage_bb} exhibit a very similar distribution, which verifies the model's ability to capture cloud patterns.
The cloud cover in the region R02 in Fig.~\ref{fig:models_stage_bb:cc} shows a long strip distribution, which is supported by the confidence of clear sky in Fig.~\ref{fig:models_stage_bb:ccs}.
The cloud cover and confidence of clear sky in the region R03 are generally consistent.
Overall, the results indicate that CldNet-O can capture subtle features of cloud distribution.

\section{Discussion}
\label{section:Discussion}
\subsection{Generalization ability}

\begin{figure}[!htp]
    \vspace{5mm}
    \centering
    \subcaptionbox{CldNet-W\vspace{-2mm}\label{fig:TransferLearning:CldNet-W}}{%
        \begin{minipage}[b]{\textwidth}
            \centering
            \includegraphics[width=0.24\textwidth,trim=0 0 30 -10,clip]{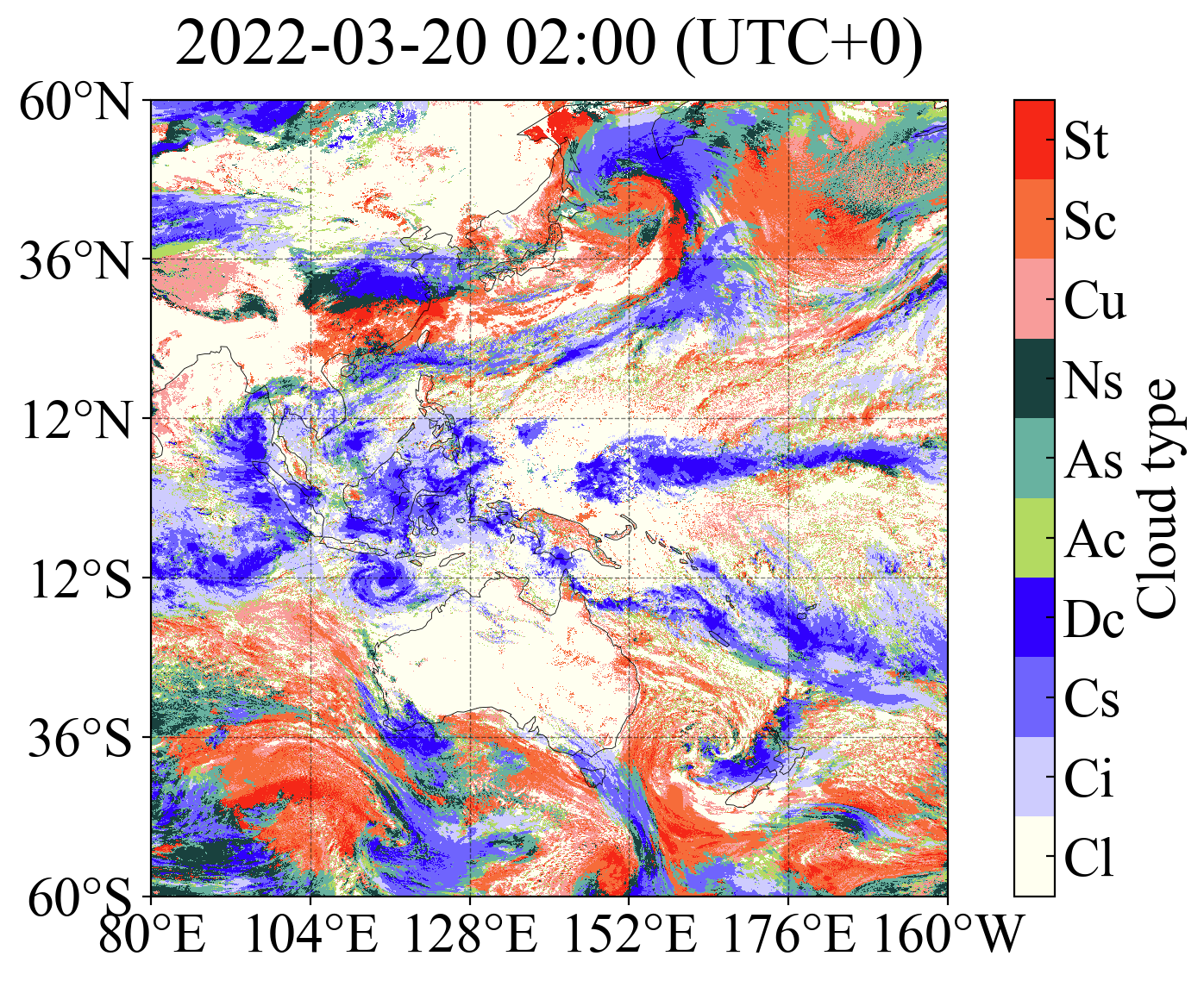}
            \includegraphics[width=0.24\textwidth,trim=0 0 30 -10,clip]{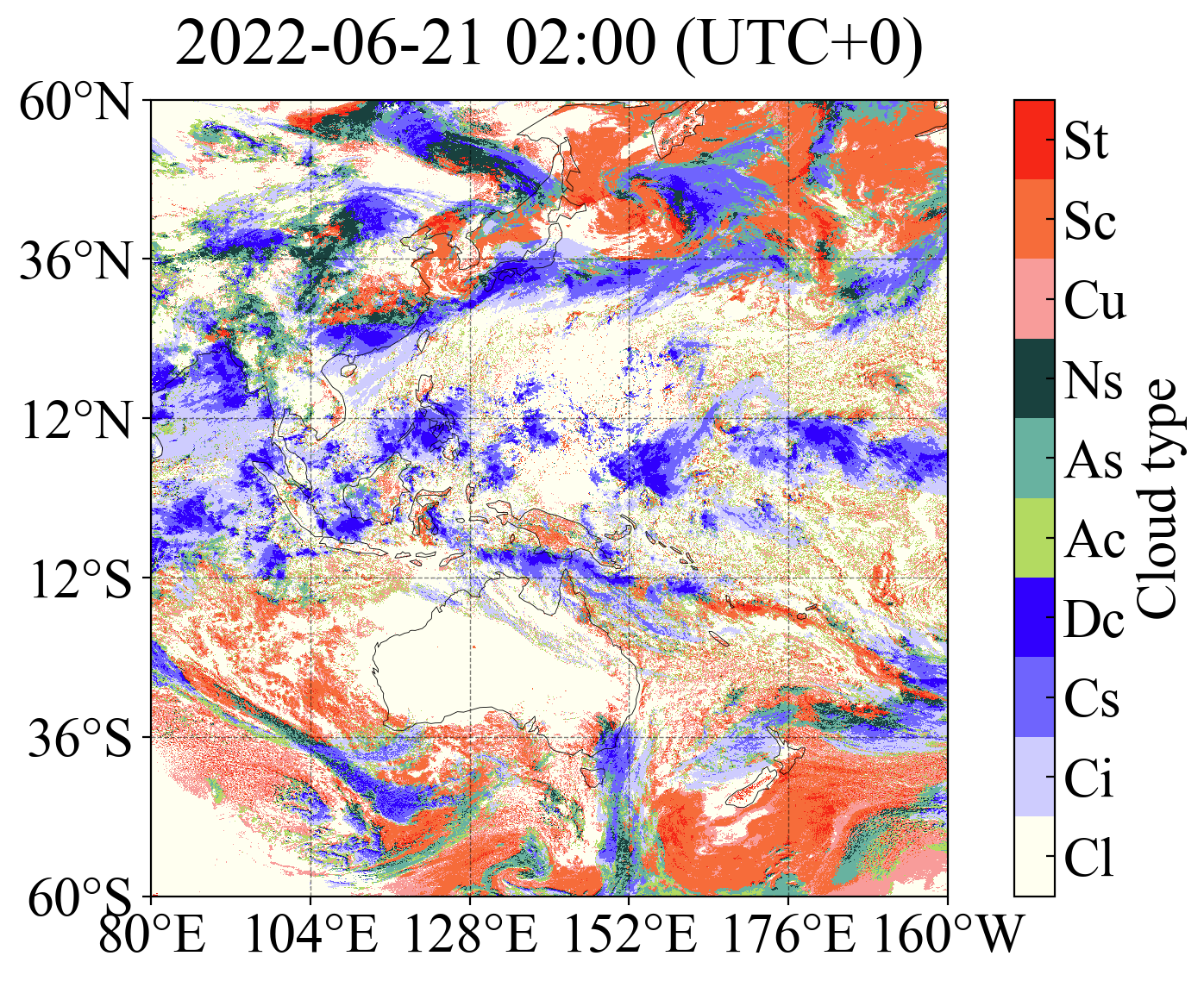}
            \includegraphics[width=0.24\textwidth,trim=0 0 30 -10,clip]{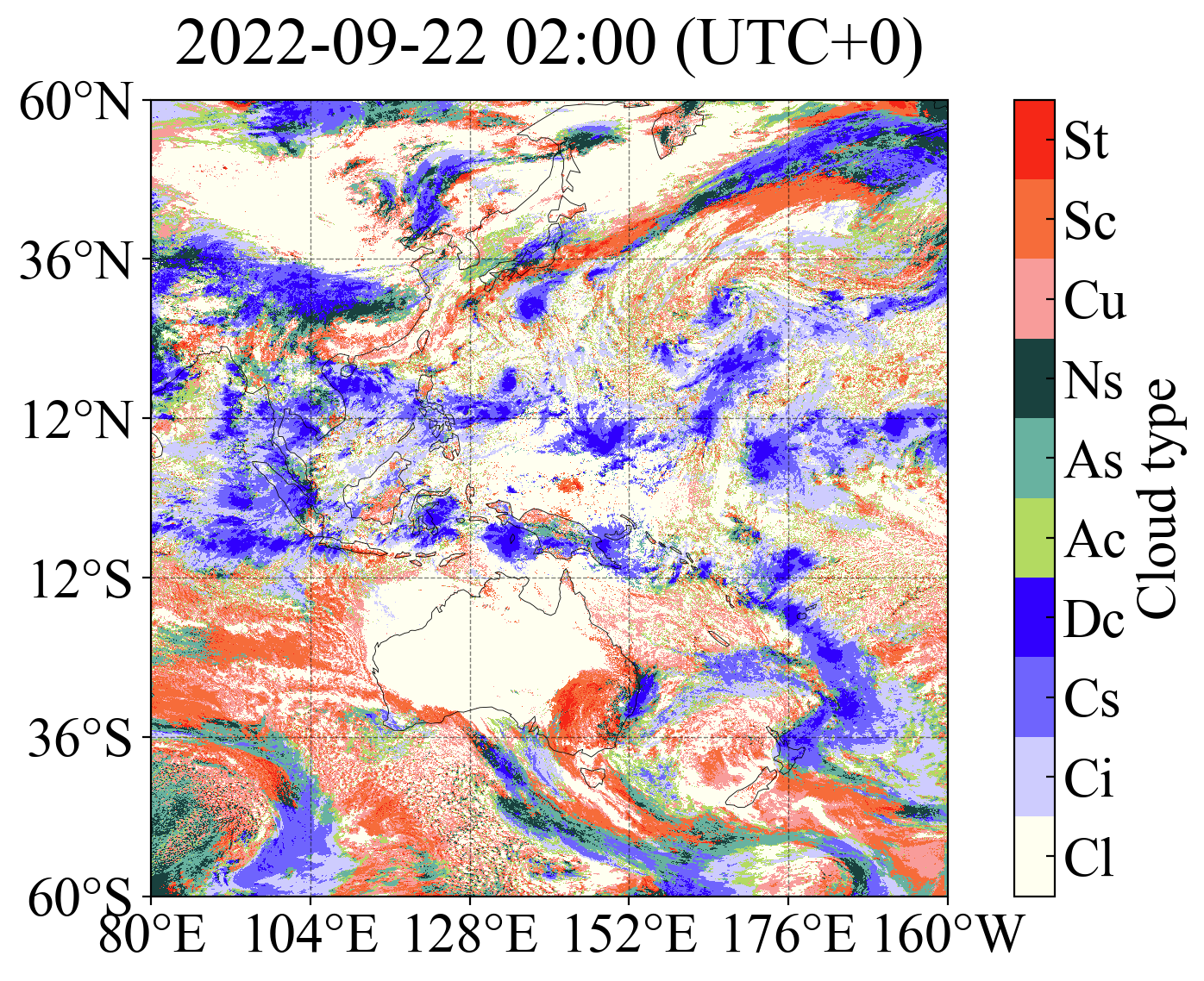}
            \includegraphics[width=0.24\textwidth,trim=0 0 30 -10,clip]{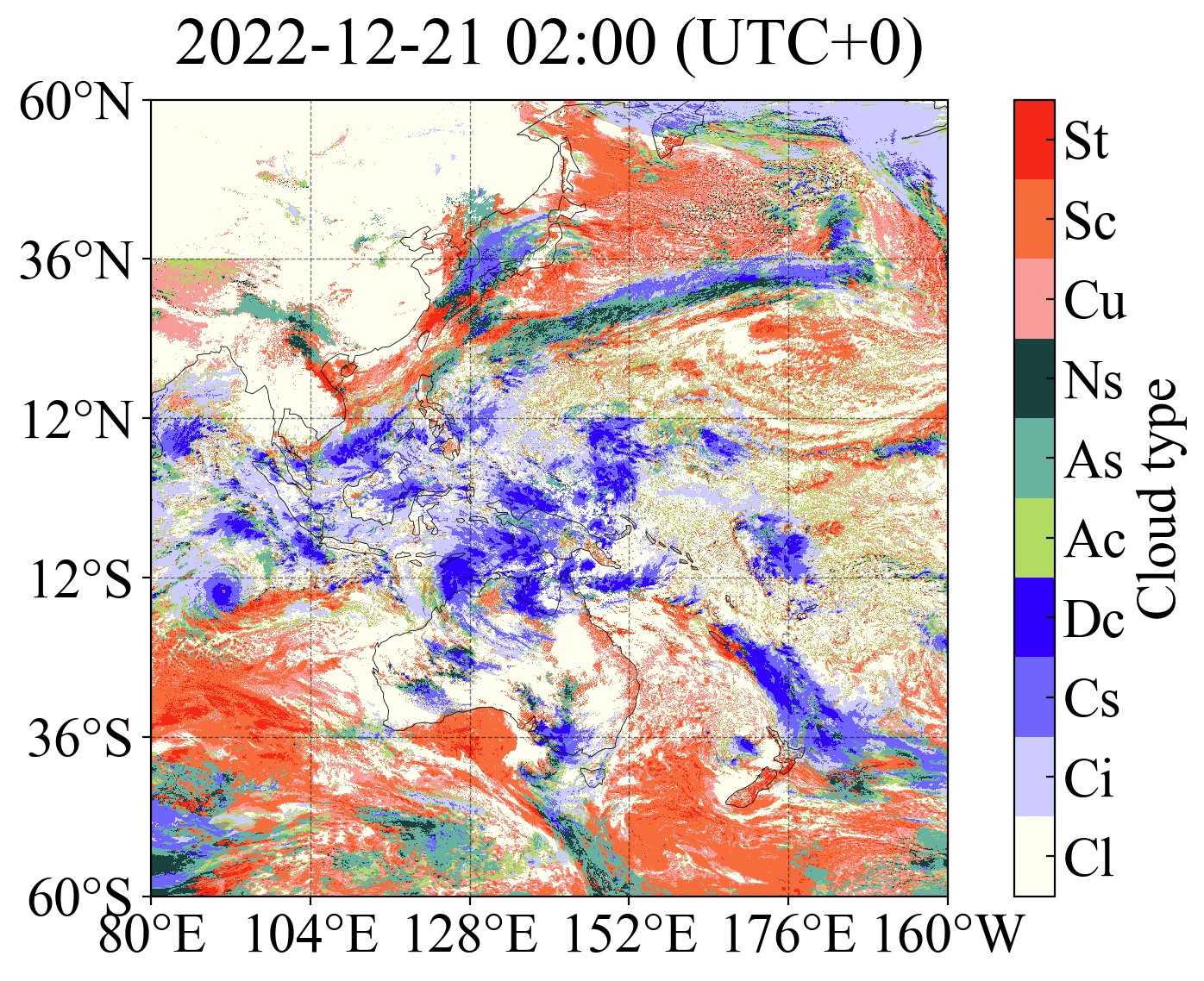}
        \end{minipage}
    }
    \subcaptionbox{CldNet-O\vspace{-2mm}\label{fig:TransferLearning:CldNet-O}}{%
        \begin{minipage}[b]{\textwidth}
            \centering
            \includegraphics[width=0.24\textwidth,trim=0 0 30 -10,clip]{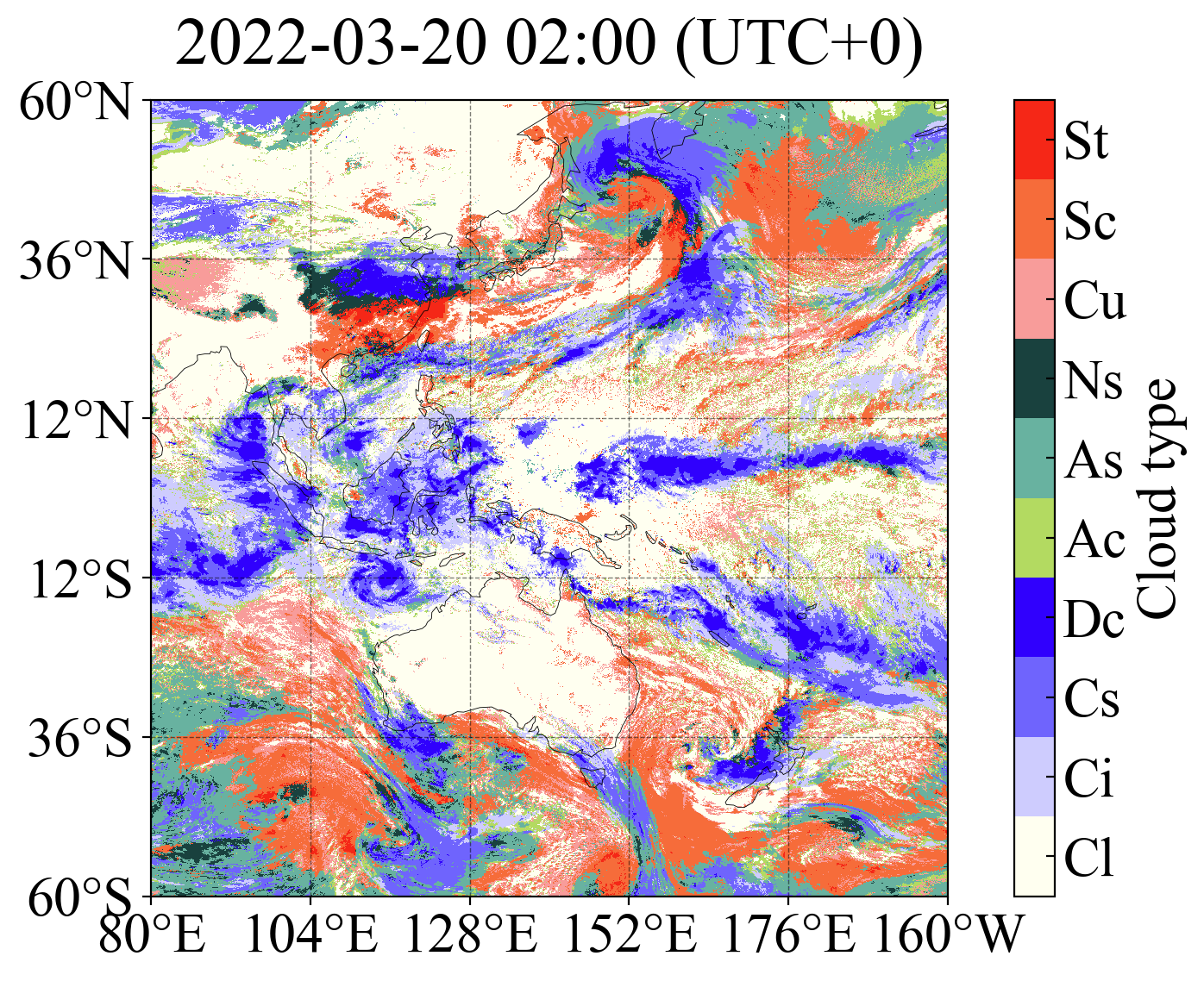}
            \includegraphics[width=0.24\textwidth,trim=0 0 30 -10,clip]{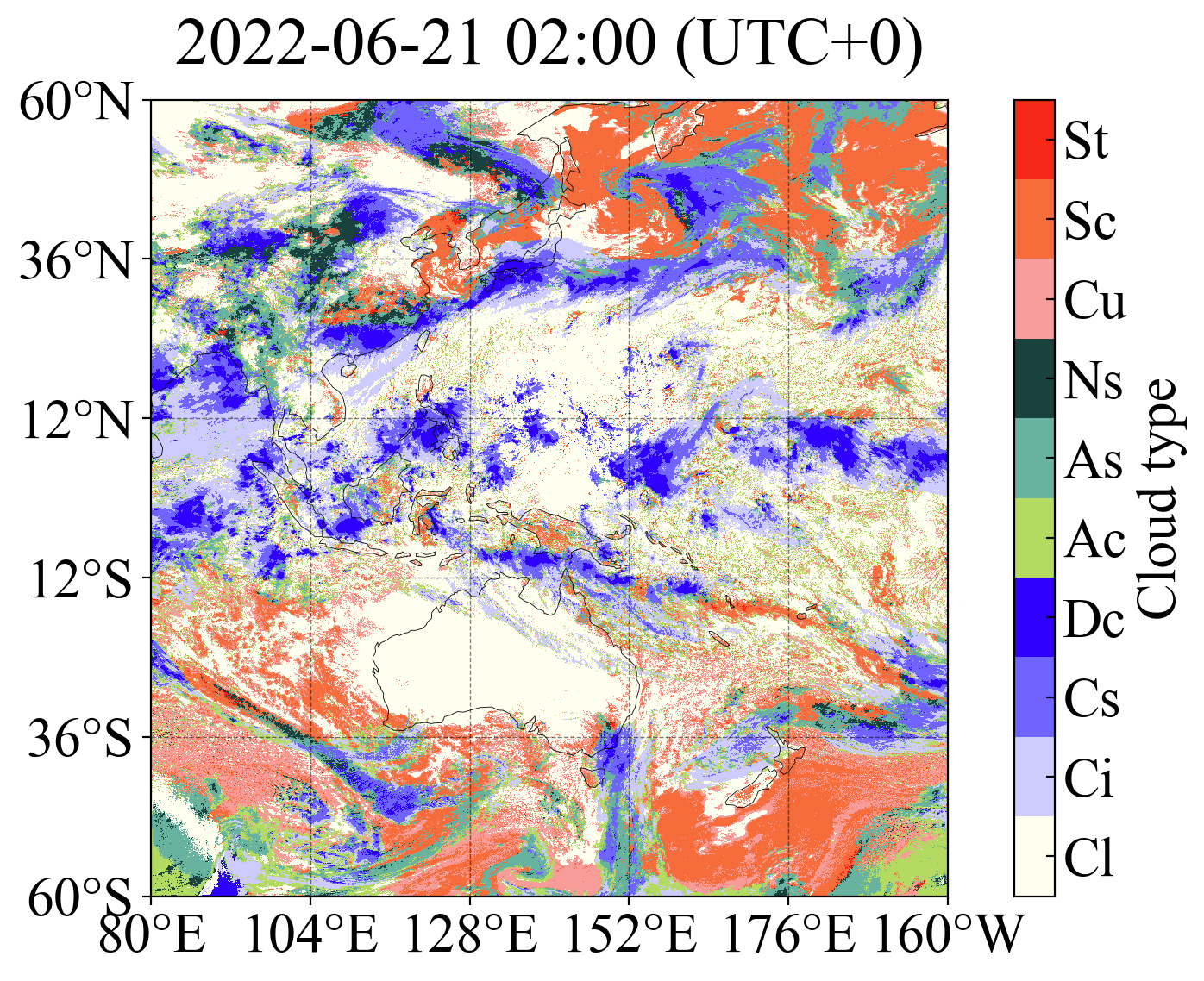}
            \includegraphics[width=0.24\textwidth,trim=0 0 30 -10,clip]{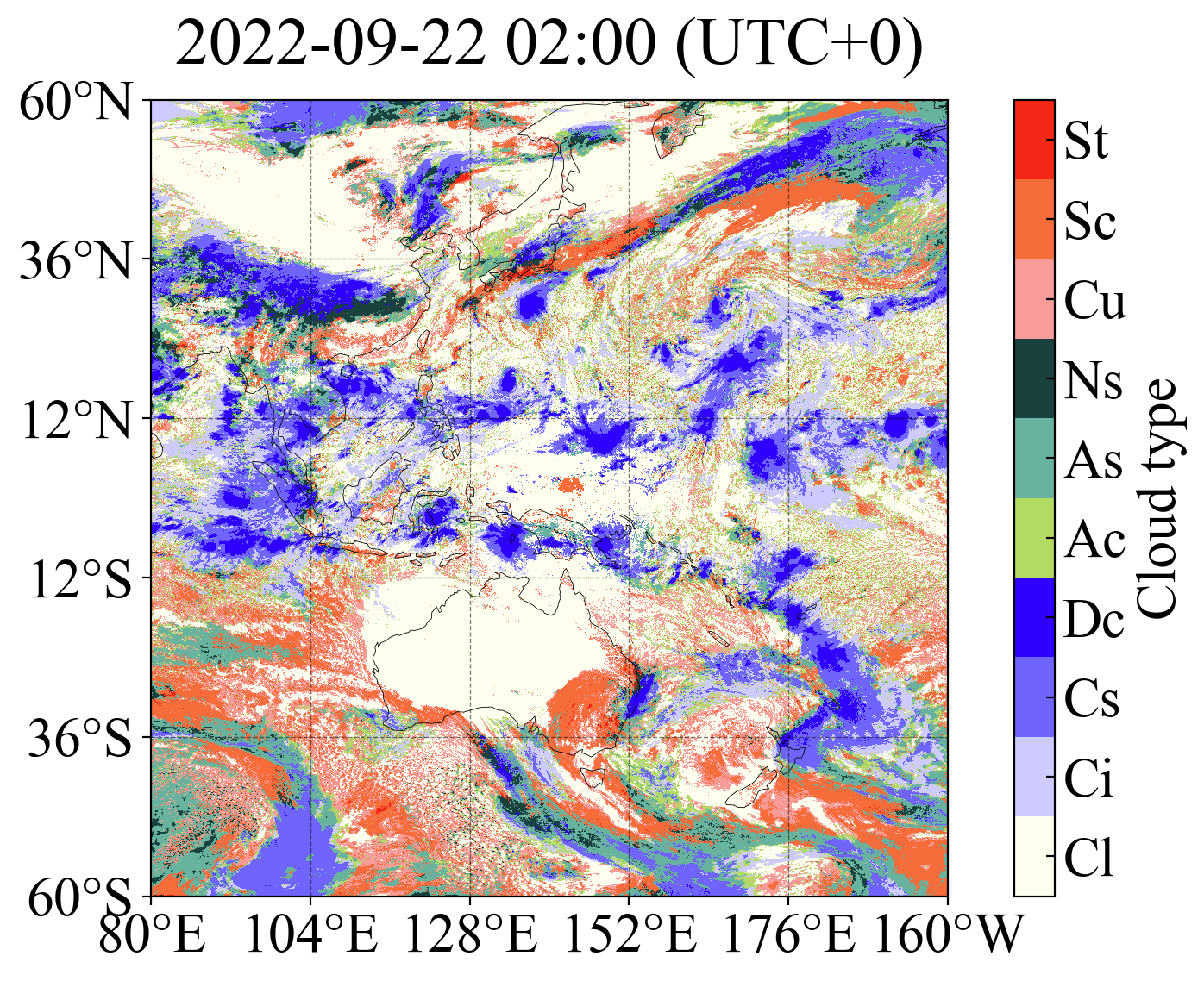}
            \includegraphics[width=0.24\textwidth,trim=0 0 30 -10,clip]{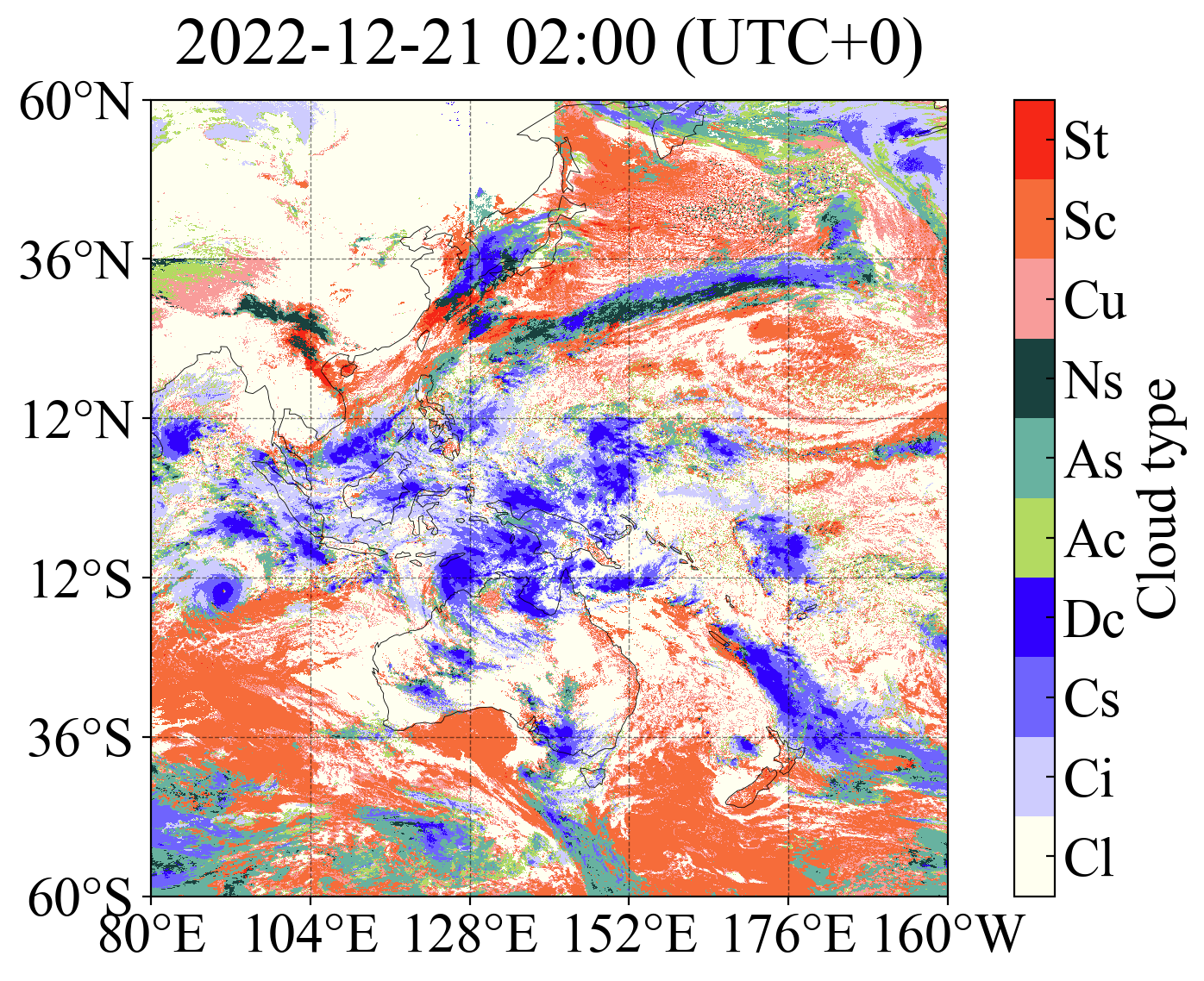}
        \end{minipage}
    }
    \subcaptionbox{Reference\vspace{-2mm}\label{fig:TransferLearning:Reference}}{%
        \begin{minipage}[b]{\textwidth}
            \centering
            \includegraphics[width=0.24\textwidth,trim=0 0 30 -10,clip]{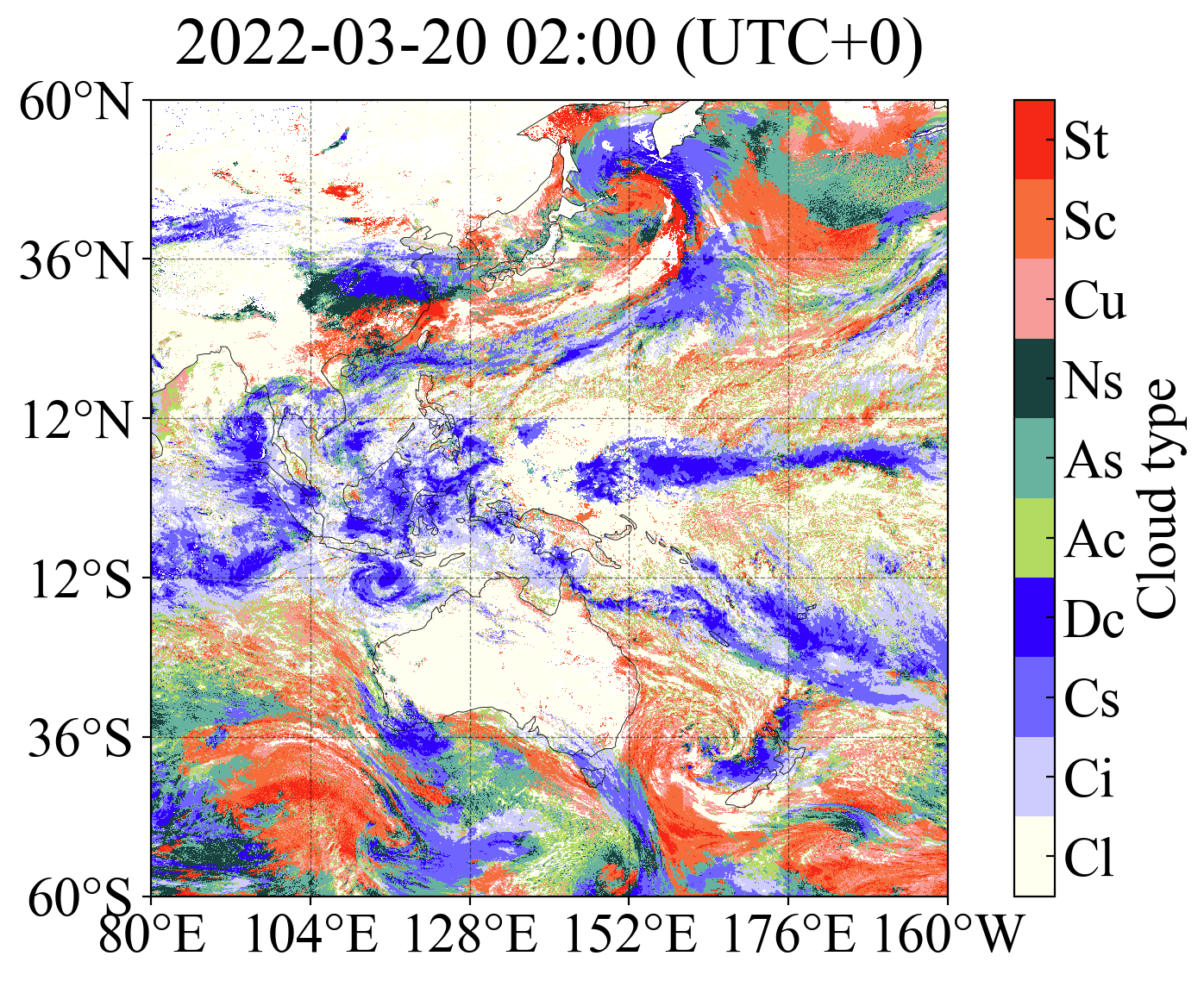}
            \includegraphics[width=0.24\textwidth,trim=0 0 30 -10,clip]{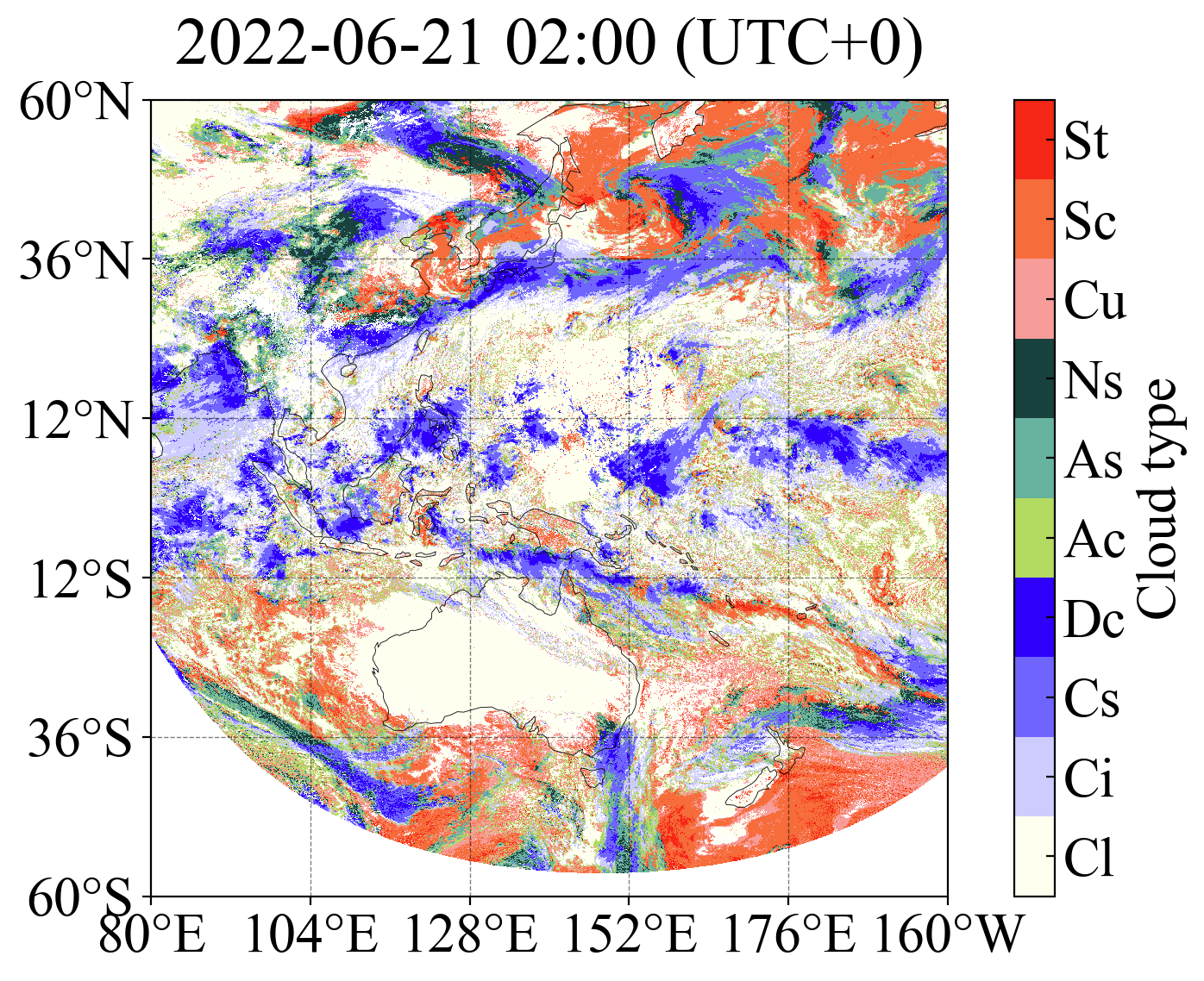}
            \includegraphics[width=0.24\textwidth,trim=0 0 30 -10,clip]{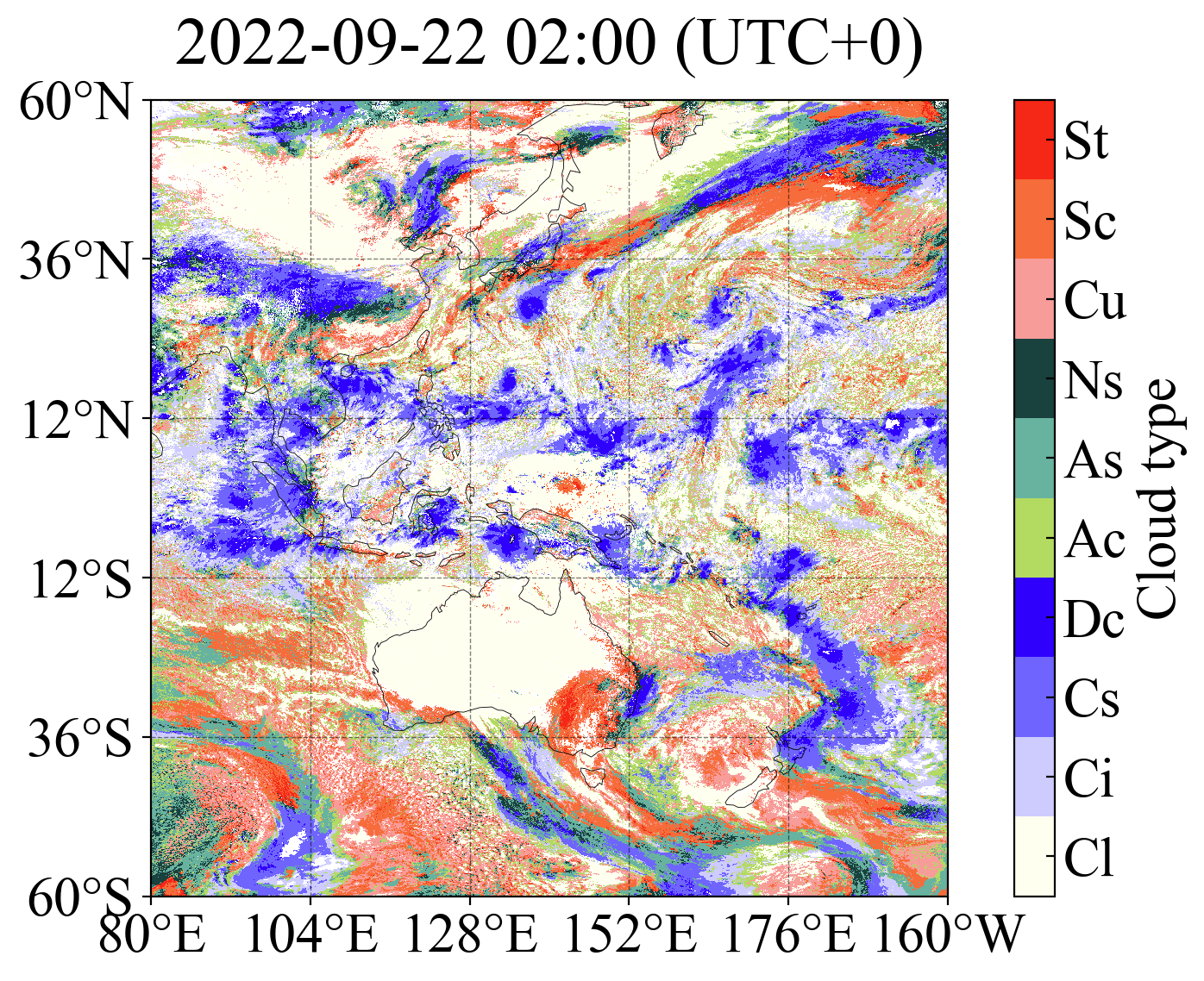}
            \includegraphics[width=0.24\textwidth,trim=0 0 30 -10,clip]{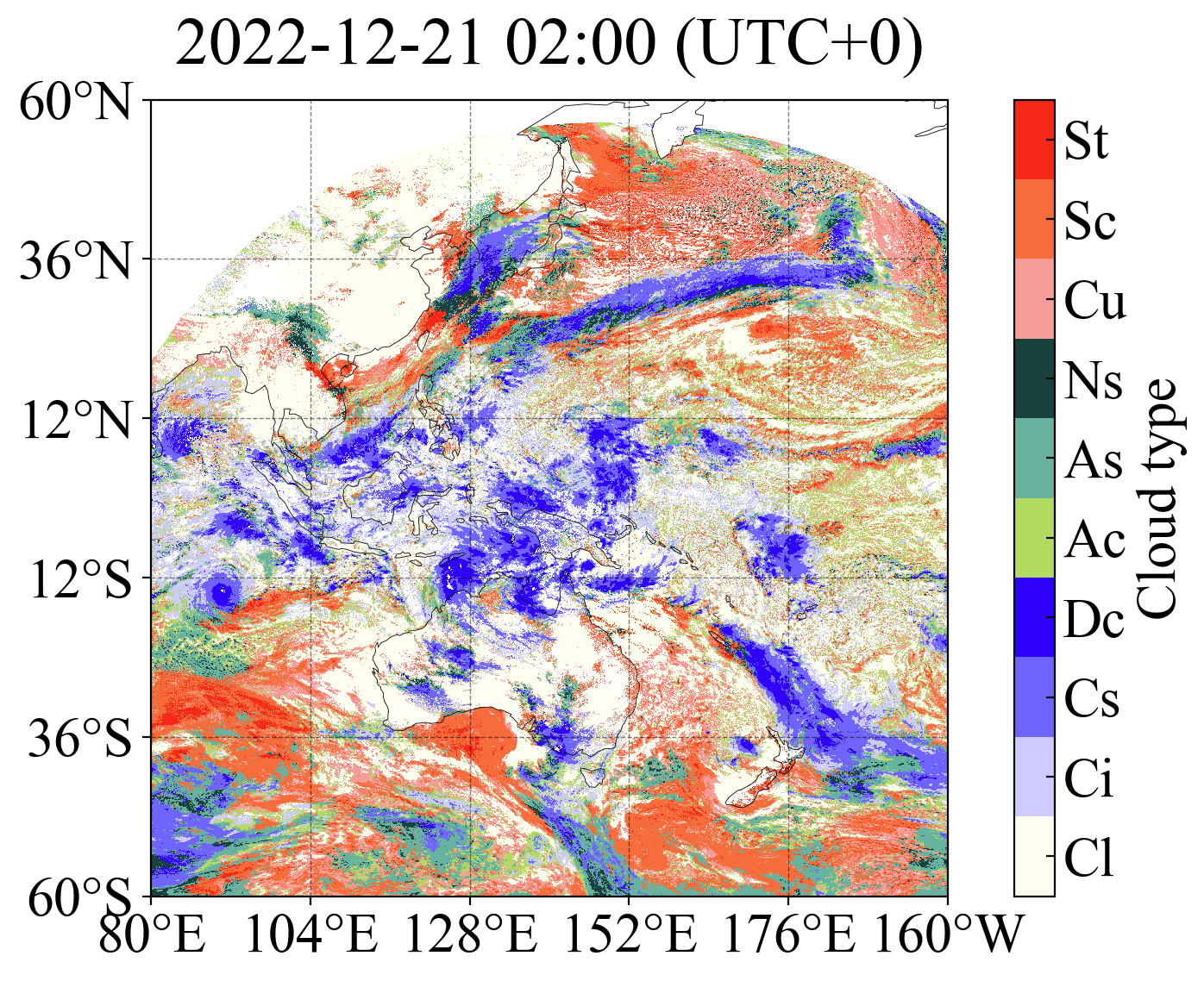}
        \end{minipage}
    }
    \caption{The cloud-type distributions with spatial resolution $0.02^{\circ}\times0.02^{\circ}$ predicted by the trained (a) CldNet-W and (b) CldNet-O, and the cloud-type distributions with spatial resolution $0.05^{\circ}\times0.05^{\circ}$ provided by (c) the JAXA's P-Tree system at 02:00 (UTC+0) for 2022-03-20, 2022-06-21, 2022-09-22, and 2022-12-21.}
    \label{fig:TransferLearning}
\end{figure}

In this section, we directly apply the previously trained model, i.e., the trained CldNet-W, and CldNet-O without any fine-tuning to satellite spectral data with spatial resolution $\mathrm{0.02^{\circ}\times0.02^{\circ}}$ from Himawari-8/9 satellite sensors to obtain the cloud-type distributions with the same spatial resolution $\mathrm{0.02^{\circ}\times0.02^{\circ}}$.
Four time points, including 2022-03-20 2:00 (UTC+0), 2022-06-21 2:00 (UTC+0), 2022-09-22 2:00 (UTC+0), and 2022-12-21 2:00 (UTC+0), are selected for cloud-type prediction.
The cloud-type distributions with spatial resolution $\mathrm{0.02^{\circ}\times0.02^{\circ}}$ predicted by the trained CldNet-W and CldNet-O are shown in Fig.~\ref{fig:TransferLearning:CldNet-W} and~\ref{fig:TransferLearning:CldNet-O}, respectively.
The JAXA's P-Tree system only provides the cloud-type product with spatial resolution $\mathrm{0.05^{\circ}\times0.05^{\circ}}$, which is used as a reference in Fig.~\ref{fig:TransferLearning:Reference}.
From the overall distribution of cloud types in Fig.~\ref{fig:TransferLearning}, the trained CldNet-W and CldNet-O have achieved good performance.

In order to quantitatively evaluate the generalization ability of our models CldNet-W and CldNet-O, two methods, Low2High and High2Low, are used to calculate cloud-type classification indicators due to the resolution difference between prediction and reference.
The process of Low2High is that the reference is first interpolated from $0.05^{\circ}\times0.05^{\circ}$ to $0.02^{\circ}\times0.02^{\circ}$ resolution, and then this is compared to the prediction to calculate those classification indicators.
The process of High2Low is that the prediction is first interpolated by neighboring interpolation from $0.02^{\circ}\times0.02^{\circ}$ to $0.05^{\circ}\times0.05^{\circ}$ resolution, and then this is compared to the reference to calculate those classification indicators.

The results of the cloud-type classification indicators are recorded in Table~\ref{table:table_model_TL}.
The indicators obtained by the method High2Low are better than those obtained by Low2High.
For accuracy (N/Y for cloud), the results of CldNet-W and CldNet-O are all above 88\%, which indicates that the model has excellent ability in the distinction between clear and cloudy skies.
Overall, the trained models CldNet-W and CldNet-O have achieved good results in directly applying high-resolution satellite spectral information to predict high-resolution cloud-type distributions in term of accuracy/$\mathrm{F_{1}\mbox{-}score_{micro}}$, $\mathrm{F_{1}\mbox{-}score_{macro}}$, $\mathrm{F_{1}\mbox{-}score_{weight}}$, and accuracy (N/Y for cloud).
This also demonstrates the good generalization ability of the model on higher resolution model input data.

\begin{table}[!htp]
    \centering
    \vspace{0cm}
    \caption{The cloud-type classification indicators obtained by applying the trained models CldNet-W and CldNet-O without any fine-tuning to satellite spectral information with spatial resolution $\mathrm{0.02^{\circ}\times0.02^{\circ}}$ from the Himawari-8/9 sensors, whose each column is the results of Low2High - High2Low.}
    \label{table:table_model_TL}
    \resizebox{0.96\linewidth}{!}{
        \begin{tabular}{cccccc}
            \hline
            Model    & Time                    & Accuracy/$\mathrm{F_{1}\mbox{-}score_{micro}}$ & $\mathrm{F_{1}\mbox{-}score_{macro}}$ & $\mathrm{F_{1}\mbox{-}score_{weight}}$ & Accuracy (N/Y for cloud) \\
            \hline
            CldNet-W & 2022-03-20 2:00 (UTC+0) & 0.71157 - 0.74954                              & 0.68238 - 0.72400                     & 0.70533 - 0.74367                      & 0.91217 - 0.92118        \\
            CldNet-W & 2022-06-21 2:00 (UTC+0) & 0.74295 - 0.79098                              & 0.70851 - 0.76545                     & 0.73599 - 0.78507                      & 0.90991 - 0.92289        \\
            CldNet-W & 2022-09-22 2:00 (UTC+0) & 0.72115 - 0.76174                              & 0.68462 - 0.73283                     & 0.71372 - 0.75503                      & 0.91207 - 0.92252        \\
            CldNet-W & 2022-12-21 2:00 (UTC+0) & 0.69669 - 0.74468                              & 0.66100 - 0.71047                     & 0.69160 - 0.73984                      & 0.91068 - 0.92309        \\
            CldNet-O & 2022-03-20 2:00 (UTC+0) & 0.66161 - 0.68174                              & 0.57814 - 0.59656                     & 0.64959 - 0.66994                      & 0.91331 - 0.92004        \\
            CldNet-O & 2022-06-21 2:00 (UTC+0) & 0.69331 - 0.71657                              & 0.58604 - 0.60743                     & 0.68059 - 0.70425                      & 0.90689 - 0.91522        \\
            CldNet-O & 2022-09-22 2:00 (UTC+0) & 0.66723 - 0.69030                              & 0.57313 - 0.59417                     & 0.65461 - 0.67831                      & 0.90940 - 0.91745        \\
            CldNet-O & 2022-12-21 2:00 (UTC+0) & 0.62770 - 0.64746                              & 0.53697 - 0.55409                     & 0.60352 - 0.62373                      & 0.87654 - 0.88287        \\
            \hline
        \end{tabular}
    }
\end{table}

\subsection{Limitation and future research}
One limitation of this study is that the satellite Himawari-8 was replaced by the satellite Himawari-9 in December 2022. Therefore, the training data is from the satellite Himawari-8, while the test data is partially from the satellite Himawari-9.
Although the sensors carried by Himawari-8 and Himawari-9 satellites are the Advanced Himawari Imager, the spectral data distribution of the two satellites may be slightly different.
The experimental results in Section~\ref{section:models_comparison} indicate that directly applying the model trained on satellite spectral data from the Himawari-8 sensor to satellite spectral data from the Himawari-9 sensor will reduce the accuracy of cloud-type recognition.
In order to ensure accuracy, it is necessary to train the spectral data obtained from each satellite separately to obtain the model parameters of the corresponding satellite spectral data.
Another limitation is that the labels used for model training are daytime cloud-type labels.
In order to enhance the credibility of the model, model parameters still need to be trained using other data containing nighttime cloud-type labels.

In future research, we will adopt appropriate measures to optimize the limitations mentioned above and strive to expand the application scope of this model from regional to global scale. 
In order to achieve global cloud-type coverage, the proposed model will be applied to multiple geostationary observation satellites, such as Meteosat, GEOS-W/E, FengYun, and Himawari.
The global distribution of cloud types is crucial in global climate change and environmental assessment research.

\section{Conclusions}
\label{section:5}
In this study, a knowledge-based data-driven (KBDD) framework for all-day identification of cloud types based on spectral information from Himawari-8/9 satellite sensors is designed, and the KBDD framework mainly consists of knowledge module, mask module, addition of auxiliary information, network candidate set, and mask loss.
Meanwhile, a novel simple and efficient network, named CldNet, is proposed in this study, which mainly consists of a DW-ASPP module and a DW-U module inspired by the ASPP module of DeepLabV3+ and UNet, respectively.

Our proposed model, CldNet, has achieved the accuracy of 80.89±2.18\% on the test dataset.
Compared with other commonly used segmentation networks, including SegNet (61.30±1.32\%), PSPNet (55.58±1.43\%), DeepLabV3+ (66.25±1.45\%), UNet (79.50±2.37\%), ResUnet (58.27±2.00\%) and UNetS (77.37±2.34\%), CldNet is state-of-the-art in cloud-type recognition.
Meanwhile, the addition of the auxiliary information, including SAZ, SAA, SOZ and SOA, improves the accuracy of CldNet by approximately 1.35\%.

By setting the input data involving VIS and NIR to zero in mask module, the trained CldNet-O is capable of achieving cloud-type prediction over nighttime areas.
More importantly, the trained models CldNet-W and CldNet-O without any fine-tuning are directly applied to satellite spectral data with spatial resolution $\mathrm{0.02^{\circ}\times0.02^{\circ}}$ from the Himawari-8/9 satellite sensors to obtain the cloud-type distributions with the same spatial resolution $\mathrm{0.02^{\circ}\times0.02^{\circ}}$, achieving accuracy of above 75\% and 65\%, respectively.
Furthermore, the cloud-type distributions with spatial resolution $\mathrm{0.02^{\circ}\times0.02^{\circ}}$ are similar to those with spatial resolution $\mathrm{0.05^{\circ}\times0.05^{\circ}}$ provided by the JAXA's P-Tree system.
This demonstrates that our framework has strong generalization ability for high-resolution model input data.

The KBDD framework using CldNet is a highly effective cloud-type identification system capable of providing a high-fidelity, all-day, spatiotemporal cloud-type database for many climate assessment fields.
In practice, CldNet-W and CldNet-O can be deployed for daytime and nighttime areas, respectively.
Meanwhile, the total parameters of CldNet-W/CldNet-O are only 0.46M, making it easy to deploy online on edge devices.
For long-term research, the KBDD framework with more cloud property prediction capabilities, and even embedding physical models, to improve accuracy will be explored globally in the future.

\section*{CRediT authorship contribution statement}
Longfeng Nie: formal analysis, data curation, methodology, software, visualization, writing - original draft. Yuntian Chen: formal analysis, writing - review and editing, funding acquisition. Mengge Du: writing - review and editing. Changqi Sun: writing - review and editing. Dongxiao Zhang: supervision, writing - review and editing, funding acquisition.

\section*{Declaration of Competing Interest}
The authors declare that they have no known competing financial interests or personal relationships that could have appeared to influence the work reported in this paper.

\section*{Data availability}
The GitHub page \href{https://github.com/rsai0/PMD/tree/main/CldNetV1_0_0}{https://github.com/rsai0/PMD/tree/main/CldNetV1\_0\_0} is available, which includes the specific network structure settings for all models.
Further updates will follow.

\section*{Acknowledgements}
This work was supported and partially funded by the National Center for Applied Mathematics Shenzhen (NCAMS), the National Natural Science Foundation of China (Grant No. 62106116), China Meteorological Administration Climate Change Special Program (CMA-CCSP) under Grant QBZ202316, the Major Key Project of PCL (Grant No. PCL2022A05), and High Performance Computing Center at Eastern Institute of Technology, Ningbo.

\begin{appendices}

    \section{}
    \setcounter{table}{0}
    \renewcommand{\thetable}{A.\arabic{table}}   
    \setcounter{figure}{0}
    \renewcommand{\thefigure}{A.\arabic{figure}}

    \begin{table}[H]
    \centering
    \caption{The classification accuracy of different models during training, validation, and test stages.}
    \label{table:model_acc_summary}
    \resizebox{0.6\linewidth}{!}{
        \begin{tabular}{ccccccc}
            \hline
            Model      & Stage      & Mean (\%) & Std (\%) & Min (\%) & Median (\%) & Max (\%) \\
            \hline
            CldNet     & Test       & 80.89     & 2.18     & 75.60    & 81.10       & 84.91    \\
            CldNet     & Training   & 81.76     & 1.64     & 77.37    & 81.64       & 85.05    \\
            CldNet     & Validation & 81.58     & 1.63     & 78.42    & 81.21       & 85.14    \\
            SegNet     & Test       & 61.30     & 1.32     & 57.71    & 61.40       & 64.77    \\
            SegNet     & Training   & 61.42     & 1.42     & 57.22    & 61.49       & 64.56    \\
            SegNet     & Validation & 61.74     & 1.41     & 58.25    & 61.87       & 64.57    \\
            PSPNet     & Test       & 55.58     & 1.43     & 52.17    & 55.64       & 58.95    \\
            PSPNet     & Training   & 54.95     & 2.06     & 48.99    & 55.17       & 60.10    \\
            PSPNet     & Validation & 55.92     & 1.70     & 51.87    & 56.37       & 59.32    \\
            DeepLabV3+ & Test       & 66.25     & 1.45     & 60.74    & 66.23       & 69.80    \\
            DeepLabV3+ & Training   & 66.93     & 1.23     & 63.30    & 66.92       & 69.65    \\
            DeepLabV3+ & Validation & 66.83     & 1.23     & 63.53    & 67.06       & 69.35    \\
            UNet       & Test       & 79.50     & 2.37     & 73.12    & 79.80       & 83.70    \\
            UNet       & Training   & 80.23     & 1.85     & 74.98    & 80.18       & 83.97    \\
            UNet       & Validation & 80.37     & 1.60     & 77.31    & 79.84       & 83.75    \\
            ResUnet    & Test       & 58.27     & 2.00     & 53.77    & 58.47       & 62.39    \\
            ResUnet    & Training   & 58.04     & 2.18     & 50.11    & 58.38       & 61.88    \\
            ResUnet    & Validation & 58.77     & 2.36     & 54.40    & 59.15       & 63.29    \\
            UNetS      & Test       & 77.37     & 2.34     & 71.53    & 77.70       & 81.67    \\
            UNetS      & Training   & 77.64     & 1.89     & 72.40    & 77.53       & 81.41    \\
            UNetS      & Validation & 78.13     & 1.57     & 75.26    & 77.98       & 81.73    \\
            UNetS-W    & Test       & 79.08     & 2.23     & 73.83    & 79.35       & 83.09    \\
            UNetS-W    & Training   & 79.51     & 1.92     & 73.86    & 79.34       & 83.24    \\
            UNetS-W    & Validation & 79.95     & 1.46     & 76.79    & 79.61       & 83.04    \\
            UNetS-O    & Test       & 69.99     & 1.75     & 64.73    & 70.15       & 73.51    \\
            UNetS-O    & Training   & 70.11     & 1.43     & 66.77    & 70.11       & 73.20    \\
            UNetS-O    & Validation & 70.55     & 1.33     & 67.89    & 70.63       & 73.42    \\
            CldNet-W   & Test       & 82.23     & 2.14     & 77.23    & 82.38       & 86.17    \\
            CldNet-W   & Training   & 82.86     & 1.71     & 77.87    & 82.71       & 86.36    \\
            CldNet-W   & Validation & 82.96     & 1.60     & 79.50    & 82.44       & 86.58    \\
            CldNet-O   & Test       & 73.21     & 2.02     & 67.11    & 73.53       & 76.83    \\
            CldNet-O   & Training   & 74.04     & 1.34     & 70.63    & 74.07       & 76.90    \\
            CldNet-O   & Validation & 73.98     & 1.35     & 71.21    & 74.07       & 76.66    \\
            \hline
        \end{tabular}
    }
\end{table}
    \begin{table}[H]
    \centering
    \vspace{-0.5cm}
    \caption{The classification evaluation indicators of various cloud types obtained by different models for 2022-09-23 03:00 (UTC+0).}
    \label{table:model_statA}
    \resizebox{0.95\linewidth}{!}{
        \begin{tabular}{cccccccccccc}
            \hline
            Model      & Indicator              & Cl      & Ci      & Cs      & Dc      & Ac      & As      & Ns      & Cu      & Sc      & St      \\
            \hline
            CldNet     & Precision              & 0.89926 & 0.81549 & 0.87483 & 0.89708 & 0.55967 & 0.73586 & 0.73301 & 0.72223 & 0.79858 & 0.72630 \\
            CldNet     & Recall                 & 0.91963 & 0.69801 & 0.90647 & 0.85264 & 0.56451 & 0.74197 & 0.68178 & 0.78788 & 0.82451 & 0.48151 \\
            CldNet     & $\mathrm{F_{1}}$-score & 0.90933 & 0.75219 & 0.89037 & 0.87430 & 0.56208 & 0.73890 & 0.70647 & 0.75363 & 0.81134 & 0.57910 \\
            SegNet     & Precision              & 0.75953 & 0.65885 & 0.68310 & 0.70178 & 0.37352 & 0.48296 & 0.44760 & 0.49174 & 0.51699 & 0.30031 \\
            SegNet     & Recall                 & 0.87888 & 0.52666 & 0.77155 & 0.57951 & 0.23953 & 0.50415 & 0.27200 & 0.52409 & 0.66293 & 0.03409 \\
            SegNet     & $\mathrm{F_{1}}$-score & 0.81486 & 0.58538 & 0.72464 & 0.63481 & 0.29188 & 0.49333 & 0.33837 & 0.50740 & 0.58093 & 0.06124 \\
            PSPNet     & Precision              & 0.77093 & 0.49427 & 0.55745 & 0.54970 & 0.30614 & 0.43938 & 0.49505 & 0.46337 & 0.49045 & 0.54712 \\
            PSPNet     & Recall                 & 0.77889 & 0.54580 & 0.73306 & 0.72357 & 0.25343 & 0.37933 & 0.32664 & 0.33915 & 0.58362 & 0.09041 \\
            PSPNet     & $\mathrm{F_{1}}$-score & 0.77489 & 0.51876 & 0.63331 & 0.62476 & 0.27730 & 0.40715 & 0.39359 & 0.39165 & 0.53299 & 0.15517 \\
            DeepLabV3+ & Precision              & 0.80689 & 0.70137 & 0.77049 & 0.80571 & 0.42516 & 0.60434 & 0.60918 & 0.52389 & 0.63459 & 0.57543 \\
            DeepLabV3+ & Recall                 & 0.86784 & 0.58361 & 0.81049 & 0.71387 & 0.37366 & 0.65713 & 0.44484 & 0.59872 & 0.65803 & 0.18535 \\
            DeepLabV3+ & $\mathrm{F_{1}}$-score & 0.83626 & 0.63709 & 0.78999 & 0.75701 & 0.39775 & 0.62963 & 0.51420 & 0.55881 & 0.64610 & 0.28039 \\
            UNet       & Precision              & 0.89000 & 0.82380 & 0.85197 & 0.88591 & 0.55346 & 0.69083 & 0.70279 & 0.71192 & 0.78000 & 0.66344 \\
            UNet       & Recall                 & 0.91971 & 0.66758 & 0.89256 & 0.78625 & 0.54091 & 0.73454 & 0.59482 & 0.78987 & 0.82110 & 0.47234 \\
            UNet       & $\mathrm{F_{1}}$-score & 0.90461 & 0.73751 & 0.87180 & 0.83311 & 0.54712 & 0.71201 & 0.64431 & 0.74887 & 0.80002 & 0.55181 \\
            ResUnet    & Precision              & 0.70701 & 0.60504 & 0.64891 & 0.71789 & 0.35650 & 0.42708 & 0.56189 & 0.48591 & 0.51580 & 0.08512 \\
            ResUnet    & Recall                 & 0.86831 & 0.55085 & 0.78377 & 0.51449 & 0.13640 & 0.44127 & 0.23545 & 0.52007 & 0.66036 & 0.00203 \\
            ResUnet    & $\mathrm{F_{1}}$-score & 0.77940 & 0.57668 & 0.71000 & 0.59940 & 0.19731 & 0.43406 & 0.33185 & 0.50241 & 0.57920 & 0.00396 \\
            UNetS      & Precision              & 0.87848 & 0.80943 & 0.83723 & 0.84560 & 0.52165 & 0.67380 & 0.66010 & 0.68090 & 0.74660 & 0.63135 \\
            UNetS      & Recall                 & 0.91656 & 0.65693 & 0.87540 & 0.78989 & 0.49330 & 0.69061 & 0.56905 & 0.76106 & 0.80982 & 0.37454 \\
            UNetS      & $\mathrm{F_{1}}$-score & 0.89711 & 0.72525 & 0.85589 & 0.81680 & 0.50708 & 0.68210 & 0.61120 & 0.71875 & 0.77693 & 0.47016 \\
            UNetS-W    & Precision              & 0.86887 & 0.82598 & 0.88479 & 0.87900 & 0.56838 & 0.74024 & 0.68192 & 0.71039 & 0.74202 & 0.54179 \\
            UNetS-W    & Recall                 & 0.91796 & 0.66130 & 0.91032 & 0.93918 & 0.48786 & 0.67082 & 0.76216 & 0.77122 & 0.84260 & 0.77500 \\
            UNetS-W    & $\mathrm{F_{1}}$-score & 0.89274 & 0.73452 & 0.89737 & 0.90810 & 0.52505 & 0.70382 & 0.71981 & 0.73955 & 0.78912 & 0.63774 \\
            UNetS-O    & Precision              & 0.86115 & 0.75843 & 0.73966 & 0.68304 & 0.49476 & 0.57628 & 0.57680 & 0.61483 & 0.63152 & 0.40416 \\
            UNetS-O    & Recall                 & 0.91666 & 0.57067 & 0.81494 & 0.61837 & 0.39027 & 0.67186 & 0.31917 & 0.69160 & 0.77872 & 0.03141 \\
            UNetS-O    & $\mathrm{F_{1}}$-score & 0.88804 & 0.65129 & 0.77548 & 0.64910 & 0.43635 & 0.62041 & 0.41094 & 0.65096 & 0.69744 & 0.05828 \\
            CldNet-W   & Precision              & 0.89334 & 0.83895 & 0.91096 & 0.90900 & 0.57737 & 0.77264 & 0.69976 & 0.74225 & 0.80456 & 0.64512 \\
            CldNet-W   & Recall                 & 0.92354 & 0.68315 & 0.92700 & 0.95484 & 0.56018 & 0.75058 & 0.82726 & 0.78538 & 0.84481 & 0.72012 \\
            CldNet-W   & $\mathrm{F_{1}}$-score & 0.90819 & 0.75307 & 0.91891 & 0.93135 & 0.56865 & 0.76145 & 0.75819 & 0.76321 & 0.82419 & 0.68056 \\
            CldNet-O   & Precision              & 0.88089 & 0.79055 & 0.78105 & 0.69935 & 0.54131 & 0.62346 & 0.53390 & 0.67256 & 0.68904 & 0.40466 \\
            CldNet-O   & Recall                 & 0.92033 & 0.61353 & 0.83083 & 0.75633 & 0.45802 & 0.69151 & 0.52409 & 0.71505 & 0.77811 & 0.22495 \\
            CldNet-O   & $\mathrm{F_{1}}$-score & 0.90018 & 0.69088 & 0.80517 & 0.72673 & 0.49620 & 0.65572 & 0.52895 & 0.69316 & 0.73087 & 0.28916 \\
            \hline
        \end{tabular}
    }
\end{table}
    \begin{table}[H]
    \centering
    \vspace{0mm}
    \caption{Statistics of the number of pixels for each cloud type obtained by CldNet at 2022-09-23 03:00 (UTC+0).}
    \label{table:CldNet_stat}
    \resizebox{0.95\linewidth}{!}{
        \begin{tabular}{ccccccccccccc}
            \hline
            \multicolumn{2}{c}{\multirow{2}{*}{Cloud type}} & \multicolumn{10}{c}{Prediction} & \multirow{2}{*}{Recall}                                                                                                                                                                           \\ \cline{3-12}
            \multicolumn{2}{c}{}                            & Cl                              & Ci                      & Cs              & Dc              & Ac              & As              & Ns              & Cu             & Sc              & St              &                          \\ \hline
            \multirow{10}{*}{Reference}                     & Cl                              & \textbf{1292447}        & 33454           & 381             & 83              & 56481           & 664             & 78             & 21029           & 743             & 34             & 0.91963 \\
                                                            & Ci                              & 59786                   & \textbf{482596} & 5715            & 0               & 134868          & 5302            & 0              & 2955            & 165             & 0              & 0.69801 \\
                                                            & Cs                              & 30                      & 22777           & \textbf{558228} & 14000           & 1726            & 18853           & 210            & 0               & 0               & 0              & 0.90647 \\
                                                            & Dc                              & 22                      & 0               & 26457           & \textbf{184882} & 0               & 2559            & 2916           & 0               & 0               & 0              & 0.85264 \\
                                                            & Ac                              & 49754                   & 47880           & 669             & 0               & \textbf{367854} & 21679           & 3              & 151048          & 12743           & 0              & 0.56451 \\
                                                            & As                              & 461                     & 4689            & 45733           & 1316            & 17942           & \textbf{383128} & 15806          & 3314            & 43423           & 555            & 0.74197 \\
                                                            & Ns                              & 417                     & 0               & 918             & 5811            & 11              & 23972           & \textbf{84935} & 2               & 4702            & 3810           & 0.68178 \\
                                                            & Cu                              & 32946                   & 388             & 0               & 0               & 74181           & 975             & 0              & \textbf{507022} & 28015           & 1              & 0.78788 \\
                                                            & Sc                              & 1255                    & 0               & 0               & 0               & 4209            & 60139           & 2701           & 16655           & \textbf{430180} & 6601           & 0.82451 \\
                                                            & St                              & 119                     & 0               & 0               & 0               & 0               & 3383            & 9222           & 0               & 18710           & \textbf{29192} & 0.48151 \\
            \multicolumn{2}{c}{Precision}                   & 0.89371                         & 0.81867                 & 0.87425         & 0.89895         & 0.55957         & 0.71936         & 0.75734         & 0.71921        & 0.80165         & 0.72324         &                          \\ \hline
        \end{tabular}
    }
\end{table}
    \begin{table}[H]
    \centering
    \vspace{-0.0cm}
    \caption{The overall classification evaluation indicators of different models for 2022-09-23 03:00 (UTC+0).}
    \label{table:model_statB}
    \resizebox{0.8\linewidth}{!}{
        \begin{tabular}{ccccc}
            \hline
            Model      & Accuracy/$\mathrm{F_{1}\mbox{-}score_{micro}}$ & $\mathrm{F_{1}\mbox{-}score_{macro}}$ & $\mathrm{F_{1}\mbox{-}score_{weight}}$ & Accuracy (N/Y for cloud) \\
            \hline
            CldNet     & 0.79305                                        & 0.75777                               & 0.79207                                & 0.95269                  \\
            SegNet     & 0.61227                                        & 0.50328                               & 0.59734                                & 0.89697                  \\
            PSPNet     & 0.55256                                        & 0.47096                               & 0.54198                                & 0.88326                  \\
            DeepLabV3+ & 0.67093                                        & 0.60472                               & 0.66603                                & 0.91233                  \\
            UNet       & 0.77928                                        & 0.73512                               & 0.77754                                & 0.94996                  \\
            ResUnet    & 0.59120                                        & 0.47143                               & 0.56555                                & 0.87320                  \\
            UNetS      & 0.75930                                        & 0.70613                               & 0.75655                                & 0.94577                  \\
            UNetS-W    & 0.78079                                        & 0.75478                               & 0.77710                                & 0.94310                  \\
            UNetS-O    & 0.69991                                        & 0.58383                               & 0.68996                                & 0.94037                  \\
            CldNet-W   & 0.80649                                        & 0.78678                               & 0.80498                                & 0.95183                  \\
            CldNet-O   & 0.73310                                        & 0.65170                               & 0.72853                                & 0.94735                  \\
            \hline
        \end{tabular}
    }
\end{table}

\end{appendices}

\bibliography{./refs/mybibfile_abbreviated}

\end{document}